\documentclass[DIV15, a4paper, 11pt]{article}

\usepackage[ruled, vlined, linesnumbered, commentsnumbered]{algorithm2e}
\usepackage[toc,page]{appendix}
\usepackage{authblk}
\usepackage{prettyref}
\usepackage{subfig}
\usepackage{graphicx}
\usepackage{booktabs}
\usepackage{colortbl}
\usepackage{amsthm,amsopn,amssymb,amsmath,amsfonts}
\usepackage{algorithmic}
\usepackage{multirow}
\usepackage{tabularx}
\usepackage{footnote}
\usepackage{threeparttable}
\usepackage{hhline}
\usepackage{cite}
\usepackage{scrpage2}
\usepackage{setspace}
\usepackage{color}
\usepackage{xcolor}  % Required for custom colors

\usepackage[margin=2.2cm]{geometry}	% set the page layout

\usepackage{tikz,pgfplots}
\definecolor{myblue}{RGB}{34,31,217}
\definecolor{mycyan}{gray}{.7}

\newtheorem{theorem}{Theorem}
\newtheorem{definition}{Definition}

\DeclareMathOperator*{\argmax}{argmax}
\DeclareMathOperator*{\argmin}{argmin}

\newcommand{\bb}[1]{\multicolumn{1}{>{\columncolor{mycyan}}c}{\textbf{{#1}}}}

\newcommand{\pref}{\prettyref}

\newrefformat{fig}{Fig.~\ref{#1}}
\newrefformat{tab}{Table~\ref{#1}}
\newrefformat{sec}{Section~\ref{#1}}
\newrefformat{alg}{Algorithm~\ref{#1}}
\newrefformat{property}{Property~\ref{#1}}
\newrefformat{theorem}{Theorem~\ref{#1}}
\newrefformat{def}{Definition~\ref{#1}}
\newrefformat{eq}{equation~(\ref{#1})}
\newrefformat{app}{Appendix~\ref{#1}}
\newrefformat{proposition}{Proposition~\ref{#1}}

\usepackage{lscape}

\begin{document}

\title{\textbf\LARGE\selectfont Dynamic Multi-Objectives Optimization with a Changing Number of Objectives\setcounter{footnote}{1}\footnote{This work has been accepted for publication to the IEEE Transactions on Evolutionary Computation. Copyright may be transferred without notice, after which this version may no longer be accessible.}}

\author[1]{\normalsize\fontfamily{lmss}\selectfont Renzhi Chen$^{\ddag}$}
\author[2]{\normalsize\fontfamily{lmss}\selectfont Ke Li\setcounter{footnote}{2}\footnote{The first two authors, sorted alphabetically, make equal contributions to this work.}}
\author[1]{\normalsize\fontfamily{lmss}\selectfont Xin Yao}
\affil[1]{\normalsize\fontfamily{lmss}\selectfont CERCIA, School of Computer Science, University of Birmingham}
\affil[2]{\normalsize\fontfamily{lmss}\selectfont College of Engineering, Mathematics and Physical Sciences, University of Exeter}
\affil[*]{\normalsize\fontfamily{lmss}\selectfont Email: k.li@exeter.ac.uk, \{rxc332, x.yao\}@cs.bham.ac.uk}

\date{}
\maketitle

{\normalsize\fontfamily{lmss}\selectfont \textbf{Abstract:}}
Existing studies on dynamic multi-objective optimization focus on problems with time-dependent objective functions, while the ones with a changing number of objectives have rarely been considered in the literature. Instead of changing the shape or position of the Pareto-optimal front/set when having time-dependent objective functions, increasing or decreasing the number of objectives usually leads to the expansion or contraction of the dimension of the Pareto-optimal front/set manifold. Unfortunately, most existing dynamic handling techniques can hardly be adapted to this type of dynamics. In this paper, we report our attempt toward tackling the dynamic multi-objective optimization problems with a changing number of objectives. We implement a dynamic two-archive evolutionary algorithm which maintains two co-evolving populations simultaneously. In particular, these two populations are complementary to each other: one concerns more about the convergence while the other concerns more about the diversity. The compositions of these two populations are adaptively reconstructed once the environment changes. In addition, these two populations interact with each other via a mating selection mechanism. Comprehensive experiments are conducted on various benchmark problems with a time-dependent number of objectives. Empirical results fully demonstrate the effectiveness of our proposed algorithm.

{\normalsize\fontfamily{lmss}\selectfont \textbf{Keywords:}} Multi-objective optimization, dynamic optimization, evolutionary algorithms, changing objectives, decomposition-based method.

% !Tex root = main.tex

\section{Introduction}
\label{sec:introduction}

Dynamic multi-objective optimization (DMO) is more challenging than the stationary scenarios, as it needs to deal with the trade-offs and the time-dependent objective functions or constraints simultaneously. An effective method for solving the DMO problems (DMOPs) needs to overcome certain difficulties raised by the dynamics, such as tracking the time-dependent Pareto-optimal front (PF) or Pareto-optimal set (PS), and providing diversified solutions that enable a timely adaptation to the changing environment. Due to its inherent adaptation property~\cite{GA}, evolutionary algorithm (EA) has been recognized as a powerful tool for DMO. According to the ways of handling dynamics, the existing techniques can be classified into the following three categories~\cite{dynamic}.
\begin{itemize}
    \item\textit{Diversity enhancement}: As the name suggests, its basic idea is to propel an already converged population jump out of the current optimum by increasing the population diversity. This is usually implemented by either responding to the dynamics merely after detecting a change of the environment or maintaining the population diversity throughout the whole optimization process. The prior one is known as \textit{diversity introduction} and is usually accompanied by a change detection technique. For example, \cite{DebNK06,Zheng07,LiuZJLM10} developed some hyper-mutation methods by which the solutions are expected to jump out of their current positions and thus to adapt to the changed environment. In~\cite{DebNK06,AragonEC05} and~\cite{AzevedoA11}, some additional solutions, either generated in a random manner or by some heuristics, are injected into the current population to participate the survival competition. On the other hand, the latter one, called \textit{diversity maintenance}, usually does not explicitly respond to the changing environment~\cite{AzzouzBS14,LiKCLZS12} whereas it mainly relies on its intrinsic diversity maintenance strategy to adapt to the dynamics. In~\cite{GohT09}, a multi-population strategy is developed to enhance the population diversity in a competitive and cooperative manner.
    \item\textit{Memory mechanism}: The major purpose of this sort of method is to utilize the historical information to accelerate the convergence progress. For example, \cite{WangL09a} suggested a hybrid memory scheme that re-initializes a new population either by re-evaluating the previous non-dominated solutions or by using a local search upon the elite solutions preserved in the archive. \cite{PengZZ14} exploited the historical knowledge and used it to build up an evolutionary environment model. Accordingly, this model is used to guide the adaptation to the changing environment.
    \item\textit{Prediction strategy}: This sort of method is usually coupled with a memory mechanism to re-initialize the population according to the anticipated optimal information. For example, \cite{HatzakisW06a} and~\cite{HatzakisW06b} employed an autoregressive model, built upon the historical information, to seed a new population within the neighborhood of the anticipated position of the new Pareto-optimal set (PS). By utilizing the regularity property~\cite{ZhangZJ08}, \cite{ZhouJZ2014,ZhouJZST06,PengZZL15} developed various models to capture the movement of the PS manifolds' centers under different dynamic environments. Instead of anticipating the position of the new PS when the environment changes, \cite{WuJL15} and \cite{KooGT10} suggested to predict the optimal moving direction of the population. More recently, \cite{MurugananthaTV15} proposed a hybrid scheme, which uses the Kalman Filter or random re-initialization, alternatively, according to their online performance, to respond to the changing environment.
\end{itemize}

It is worth noting that the current studies on DMO mainly focus on problems with time-dependent objective functions, which result in the change of the PF or PS over time. In addition to this type of dynamic, many real-life scenarios consider problems in which the number of objective functions may also change over time. For example, a software development life cycle usually contains four phases~\cite{Kruchten2000}, i.e., inception, elaboration, construction and transition. Each phase involves various tasks, some of which might be different from the other phases. In other words, different phases consider various number of objectives. In order to manage changes (changing requirements, technology, resources, products and so on), the software development process needs to switch among different phases back and forth, and thus changing the number of objectives. In project scheduling, we usually want to minimize the makespan and total salary cost simultaneously. However, given a tighter deadline must be met, the salary cost may not be important and would not be considered as an objective any longer under the new circumstance~\cite{ShenMBY16,ShenY15,XiaoOWL10}. In some other examples, if an application is running on a system with a wired power supply, there is no need to consider the energy consumption~\cite{ZhuoC05}. However, once the system is detached from the wired power supply and only relies on batteries, minimizing the energy consumption thus becomes a new objective. Furthermore, in a heterogeneous multi-core system, power consumption, performance, temperature, master frequency and reliability are usually considered altogether when designing a runtime scheduler~\cite{ChenLY14,SharifiCR10,LuoJ02}. However, these objectives can be ignored or re-considered according to some external instances, e.g., the power plugins, excessive temperature. Similar situations happen in the cloud computing resource scheduling, where even more varieties of objectives will be considered, e.g., load balance, quality of service~\cite{FardPF13,ZhanLGZCL15,ChenDZZ15}.

The study on how to handle the DMOP with a changing number of objectives is rather limited, although this concept was mentioned in the literature~\cite{GuanCM05,HuangSA11} and~\cite{AzzouzBS14}. To the best of our knowledge, \cite{GuanCM05} is the only one that discussed the effects of the objective increment and replacement. It proposed an inheritance strategy that re-evaluates the objective functions of the current non-dominated solutions to adapt to this type of dynamics. Generally speaking, the DMOP with a changing number of objectives has two major characteristics: 1) the PF or PS of the original problem is a subset of the one after increasing the number of objectives, and vice versa; and 2) instead of changing the position or the shape of the original PF or PS, increasing or decreasing the number of objectives usually results in the expansion and contraction of the dimension of the PF or PS manifold. Bearing these two characteristics in mind, this paper presents a dynamic two-archive EA to handle the DMOP with a changing number of objectives. More specifically, our proposed method simultaneously maintains two co-evolving populations, which have complementary effects on the search process. In particular, one population, denoted as the convergence archive (CA), mainly focuses on providing a competitive selection pressure towards the PF; while the other one, called the diversity archive (DA), maintains a set of solutions with a promising diversity. Furthermore, the compositions of the CA and the DA are adaptively reconstructed when the environment changes. In addition, the interaction between these two populations are implemented by a restricted mating selection mechanism, where the mating parents are selected from the CA and the DA, respectively, according to the population distribution. Note that the multi-population strategy is not a brand new technique in the evolutionary multi-objective optimization (EMO) literature. For example, \cite{PraditwongY2006} and \cite{WangJY15}, the ancestors of this paper, used two complementary populations to balance the convergence and diversity during the search process. \cite{LiKD2015} developed a dual-population paradigm to take advantages of the strengths of the Pareto- and decomposition-based techniques. Nevertheless, none of them have taken the dynamic environment into consideration.

In the remainder of this paper, we will at first provide some preliminaries of this paper in~\pref{sec:preliminaries}. Then, the technical details of our proposed algorithm are described step by step in~\pref{sec:proposal}. Comprehensive experiments are conducted and analyzed in \pref{sec:empirical}. At last, \pref{sec:conclusions} concludes this paper and provides some future directions.

\section{Preliminaries}
\label{sec:preliminaries}

In this section, we first provide some definitions useful for this paper. Then, we describe the challenges posed by the DMOPs with a changing number of objectives. Finally, we discuss the pitfalls of the current dynamic handling techniques.

\subsection{Basic Definitions}
\label{sec:definitions}

There are various types of dynamic characteristics that can result in different mathematical definitions. This paper focuses on the continuous DMOPs defined as follows:
\begin{equation}
\begin{array}{l l}
    \mathrm{minimize} \quad \mathbf{F}(\mathbf{x},t)=(f_{1}(\mathbf{x},t),\cdots,f_{m(t)}(\mathbf{x},t))^{T}\\
	\mathrm{subject\ to} \quad \mathbf{x}\in\Omega, t\in\Omega_t
\end{array}
\label{eq:DMOP}
\end{equation}
where $t$ is a discrete time defined as $t=\lfloor\frac{\tau}{\tau_t}\rfloor$, $\tau$ and $\tau_t$ represent the iteration counter and the frequency of change, respectively, and $\Omega_t\subseteq\mathbb{N}$ is the time space. $\Omega=\prod_{i=1}^n[a_i,b_i]\subseteq\mathbb{R}^n$ is the decision (variable) space, $\mathbf{x}=(x_1,\cdots,x_n)^T\in\Omega$ is a candidate solution. $\mathbf{F}:\Omega\rightarrow\mathbb{R}^{m(t)}$ constitutes of $m(t)$ real-valued objective functions and $\mathbb{R}^{m(t)}$ is the objective space at time step $t$, where $m(t)$ is a discrete function of $t$.

\begin{definition}
    At time step $t$, $\mathbf{x}^1$ is said to \textit{Pareto dominate} $\mathbf{x}^2$, denoted as $\mathbf{x}^1\preceq_t\mathbf{x}^2$, if and only if: $\forall i\in\{1,\cdots,m(t)\}$, $f_i(\mathbf{x}^1,t)\leq f_i(\mathbf{x}^2,t)$; and $\exists j\in\{1,\cdots,m(t)\}$, $f_j(\mathbf{x}^1,t)<f_j(\mathbf{x}^2,t)$.
\end{definition}

\begin{definition}
    At time step $t$, $\mathbf{x}^{\ast}\in\Omega$ is said to be \textit{Pareto-optimal}, if there is no other $\mathbf{x}\in\Omega$ such that $\mathbf{x}\preceq_t\mathbf{x}^{\ast}$.
\end{definition}

\begin{definition}
    At time step $t$, the set of all Pareto-optimal solutions is called the $t$-th Pareto-optimal set ($PS_t$). The corresponding set of Pareto-optimal objective vectors is called the $t$-th Pareto-optimal front ($PF_t$), i.e., $PF_t=\{\mathbf{F}(\mathbf{x},t)|\mathbf{x} \in PS_t\}$.
\end{definition}

\begin{theorem}
	When increasing the number of objectives, the PF and the PS at time step $t$ are a subset of those at time step $t+1$; on the contrary, when decreasing the number of objectives, the PF and the PS at time step $t$ are a superset of those at time step $t+1$.
\label{theorem:subset}
\end{theorem}

\pref{theorem:subset} supports the first characteristic given in~\pref{sec:introduction}, and its proof can be found in~\pref{sec:theorem}.

\begin{definition}
    At time step $t$, the $t$-th ideal objective vector is $\mathbf{z}^{\ast}(t)=(z^{\ast}_1(t),\dotsm,z^{\ast}_{m(t)}(t))^T$, where $z^{\ast}_i(t)=\min\limits_{\mathbf{x}\in\Omega}f_i(\mathbf{x},t),i\in\{1,\dotsm,m(t)\}$.
\end{definition}

\begin{definition}
    At time step $t$, the $t$-th nadir objective vector is $\mathbf{z}^{nad}(t)=(z^{nad}_1(t),\dotsm,z^{nad}_{m(t)}(t))^T$, where $z^{nad}_i(t)=\max\limits_{\mathbf{x}\in\Omega}f_i(\mathbf{x},t),i\in\{1,\dotsm,m(t)\}$.
\end{definition}

\begin{definition}
    Under some mild conditions, the regularity property, induced from the Karush-Kuhn-Tucker condition, means that the PF and PS of a continuous $m$-objective MOP is an $(m-1)$-dimensional piecewise continuous manifold in both the objective space and the decision space.
\label{definition:manifold}
\end{definition}
Note that an efficient EMO algorithm may employ this regularity property explicitly (e.g.,~\cite{ZhangZJ2008,SchutzeCMTD08,ZhangZSZGZ16}) or implicitly (e.g.,~\cite{ZhangL2007,NebroDLDA09,MaLQWLJYG16}) in its implementation. This paper considers continuous DMOPs with the regularity property, according to which the $PF_t$ and $PS_t$ of the DMOP at time step $t$ is a $(m(t)-1)$-dimensional piecewise continuous manifold in both $\mathbb{R}^{m(t)}$ and $\Omega$. In addition, according to~\pref{definition:manifold}, we can easily derive the second characteristic given in~\pref{sec:introduction}.

\subsection{Challenges of DMOP with a Changing Number of Objectives}
\label{sec:challenges}

Rather than shifting the position or adjusting the geometric shape of the PF or PS~\cite{FarinaDA04}, changing the number of objectives usually results in the expansion or contraction of the dimensionality of the PF or PS manifold. In order to understand the challenges posed by changing the number of objectives, this subsection uses DTLZ2~\cite{DTLZ} as the benchmark problem and multi-objective EA based on decomposition (MOEA/D)~\cite{LiZ09} as the baseline algorithm for investigations. Note that the DTLZ problems are scalable to any number of objectives and MOEA/D has been reported to show good performance on a wide range of problems~\cite{LiZ09,IshibuchiAN15,LiLTY15}. To have a better visual comparison, we only discuss the challenges in the two- and three-objective scenarios while the challenges will become more severe with the growth of dimensionality.

\subsubsection{When Increasing the Number of Objectives}
\label{sec:increasing}

\begin{figure}[htbp]
\centering

\pgfplotsset{every axis/.append style={
font = \Large,
grid = major,
thick,
line width = 1pt,
tick style = {line width = 0.8pt, font = \Large}}}

\subfloat[Approximation in the 2-D case]{
\resizebox{0.4\textwidth}{!}{
    \begin{tikzpicture}
        \begin{axis}[
            xmin   = 0, xmax = 1,
            ymin   = 0, ymax = 1,
            xlabel = {$f_1$},
            ylabel = {$f_1$}
        ]
            \addplot[black, mark = *] table[x index = 0, y index = 1] {data/I_2D.dat};
            \addplot[gray, samples y = 0] table[x index = 0, y index = 1] {data/pf-background/2d.dat};
  	    \end{axis}
    \end{tikzpicture}
}}
\subfloat[After increasing one objective]{
\resizebox{0.4\textwidth}{!}{
    \begin{tikzpicture}
        \begin{axis}[
            xmin   = 0, xmax = 1,
            ymin   = 0, ymax = 1,
            zmin   = 0, zmax = 1,
            xlabel = {$f_1$},
            ylabel = {$f_2$},
            zlabel = {$f_3$},
            xtick = {0, 0.5, 1},
            ytick = {0, 0.5, 1},
            view = {145}{30}
        ]
            \addplot3[only marks, thick, black, mark = *] table {data/I_3D.dat};
            \addplot3[gray, samples y = 0] table {data/pf-background/1.dat};
            \addplot3[gray, samples y = 0] table {data/pf-background/2.dat};
            \addplot3[gray, samples y = 0] table {data/pf-background/3.dat};
            \addplot3[gray, samples y = 0] table {data/pf-background/4.dat};
            \addplot3[gray, samples y = 0] table {data/pf-background/5.dat};
            \addplot3[gray, samples y = 0] table {data/pf-background/6.dat};
            \addplot3[gray, samples y = 0] table {data/pf-background/7.dat};
            \addplot3[gray, samples y = 0] table {data/pf-background/8.dat};
            \addplot3[gray, samples y = 0] table {data/pf-background/9.dat};
            \addplot3[gray, samples y = 0] table {data/pf-background/10.dat};
            \addplot3[gray, samples y = 0] table {data/pf-background/11.dat};
            \addplot3[gray, samples y = 0] table {data/pf-background/12.dat};
            \addplot3[gray, samples y = 0] table {data/pf-background/13.dat};
            \addplot3[gray, samples y = 0] table {data/pf-background/14.dat};
            \addplot3[gray, samples y = 0] table {data/pf-background/15.dat};
            \addplot3[gray, samples y = 0] table {data/pf-background/16.dat};
            \addplot3[gray, samples y = 0] table {data/pf-background/17.dat};
            \addplot3[gray, samples y = 0] table {data/pf-background/18.dat};
            \addplot3[gray, samples y = 0] table {data/pf-background/19.dat};
            \addplot3[gray, samples y = 0] table {data/pf-background/20.dat};
            \addplot3[gray, samples y = 0] table {data/pf-background/21.dat};
            \addplot3[gray, samples y = 0] table {data/pf-background/22.dat};
            \addplot3[gray, samples y = 0] table {data/pf-background/23.dat};
            \addplot3[gray, samples y = 0] table {data/pf-background/24.dat};
            \addplot3[gray, samples y = 0] table {data/pf-background/25.dat};
        \end{axis}
    \end{tikzpicture}
}}
\caption{Comparison of population distributions when increasing the number of objectives on DTLZ2.}
\label{fig:DimIncrease}
\end{figure}
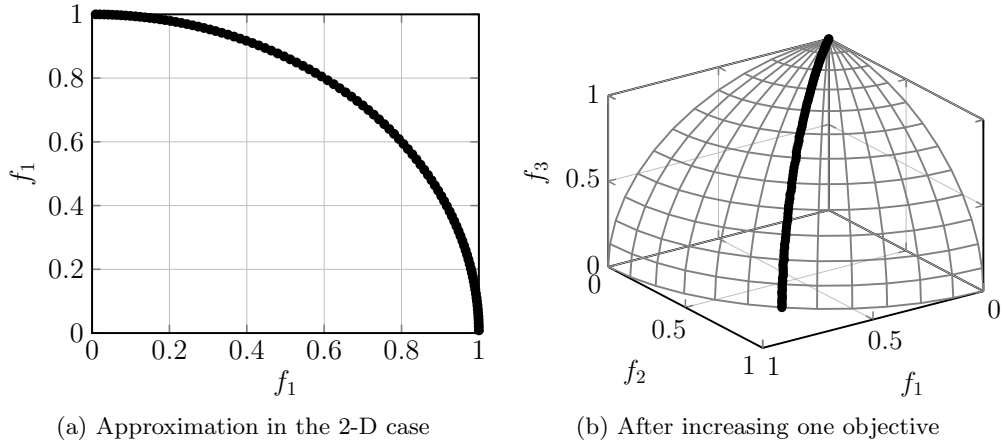

First, we run MOEA/D on a two-objective DTLZ2 instance for 200 generations\footnote{Here the differential evolution (DE)~\cite{Price2013} is used as the reproduction operator, where $CR=0.5$ and $F=0.5$. The population size is set to 100, and the other parameters are set the same as suggested in~\cite{LiZ09}.} and obtain a well approximation to the PF as shown in~\pref{fig:DimIncrease}(a). Then, we change the underlying problem to a three-objective instance. \pref{fig:DimIncrease}(b) shows the corresponding population distribution after re-evaluating the objective values. From this subfigure, we can see that although the population still converges well to the PF, the diversity is not satisfied any longer after increasing the number of objectives (as shown in~\pref{fig:DimIncrease}(b), the population crowds on a curve). Generally speaking, the most direct effect of increasing the number of objectives is the expansion of the PF or PS manifold. In particular, the population convergence might not be influenced after increasing the number of objectives; whereas the population diversity becomes severely insufficient to approximate the expanded PF or PS manifold. These characteristics pose significant challenges to the timely adaptation to the changing environment, i.e., how to propel the population jump out of the crowded areas.

\subsubsection{When Decreasing the Number of Objectives}
\label{sec:decreasing}

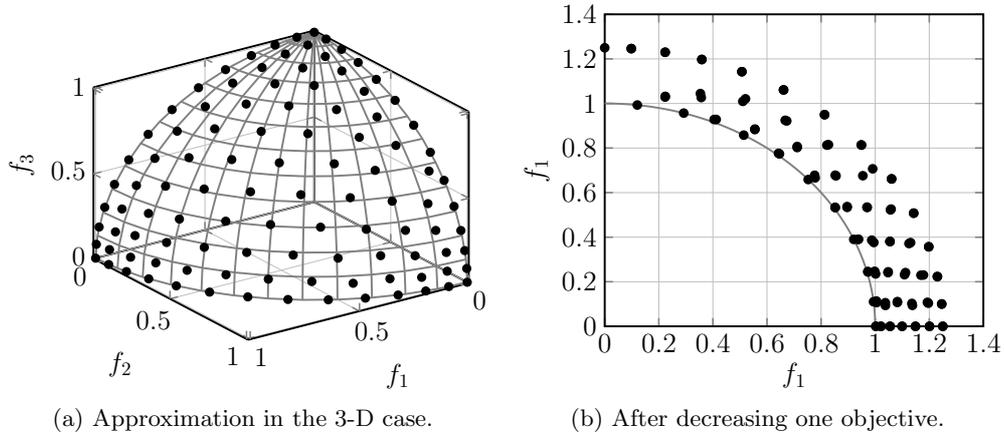
\begin{figure}[htbp]
\centering

\pgfplotsset{every axis/.append style={
font = \Large,
grid = major,
thick,
line width = 1pt,
tick style = {line width = 0.8pt, font = \Large}}}

\subfloat[Approximation in the 3-D case.]{
\resizebox{0.4\textwidth}{!}{
    \begin{tikzpicture}
        \begin{axis}[
            xmin   = 0, xmax = 1.01,
            ymin   = 0, ymax = 1.01,
            zmin   = 0, zmax = 1.01,
            xlabel = {$f_1$},
            ylabel = {$f_2$},
            zlabel = {$f_3$},
            xtick = {0, 0.5, 1},
            ytick = {0, 0.5, 1},
            view = {145}{30}
        ]
            \addplot3[only marks, thick, black, mark = *] table {data/D_3D.dat};
            \addplot3[gray, samples y = 0] table {data/pf-background/1.dat};
            \addplot3[gray, samples y = 0] table {data/pf-background/2.dat};
            \addplot3[gray, samples y = 0] table {data/pf-background/3.dat};
            \addplot3[gray, samples y = 0] table {data/pf-background/4.dat};
            \addplot3[gray, samples y = 0] table {data/pf-background/5.dat};
            \addplot3[gray, samples y = 0] table {data/pf-background/6.dat};
            \addplot3[gray, samples y = 0] table {data/pf-background/7.dat};
            \addplot3[gray, samples y = 0] table {data/pf-background/8.dat};
            \addplot3[gray, samples y = 0] table {data/pf-background/9.dat};
            \addplot3[gray, samples y = 0] table {data/pf-background/10.dat};
            \addplot3[gray, samples y = 0] table {data/pf-background/11.dat};
            \addplot3[gray, samples y = 0] table {data/pf-background/12.dat};
            \addplot3[gray, samples y = 0] table {data/pf-background/13.dat};
            \addplot3[gray, samples y = 0] table {data/pf-background/14.dat};
            \addplot3[gray, samples y = 0] table {data/pf-background/15.dat};
            \addplot3[gray, samples y = 0] table {data/pf-background/16.dat};
            \addplot3[gray, samples y = 0] table {data/pf-background/17.dat};
            \addplot3[gray, samples y = 0] table {data/pf-background/18.dat};
            \addplot3[gray, samples y = 0] table {data/pf-background/19.dat};
            \addplot3[gray, samples y = 0] table {data/pf-background/20.dat};
            \addplot3[gray, samples y = 0] table {data/pf-background/21.dat};
            \addplot3[gray, samples y = 0] table {data/pf-background/22.dat};
            \addplot3[gray, samples y = 0] table {data/pf-background/23.dat};
            \addplot3[gray, samples y = 0] table {data/pf-background/24.dat};
            \addplot3[gray, samples y = 0] table {data/pf-background/25.dat};
        \end{axis}
    \end{tikzpicture}
}}
\subfloat[After decreasing one objective.]{
\resizebox{0.4\textwidth}{!}{
    \begin{tikzpicture}
        \begin{axis}[
            xmin   = 0, xmax = 1.4,
            ymin   = 0, ymax = 1.4,
            xtick  = {0, 0.2, 0.4, 0.6, 0.8, 1.0, 1.2, 1.4},
            ytick  = {0, 0.2, 0.4, 0.6, 0.8, 1.0, 1.2, 1.4},
            xlabel = {$f_1$},
            ylabel = {$f_1$}
        ]
            \addplot[only marks, black, mark = *] table[x index = 0, y index = 1] {data/D_2D.dat};
            \addplot[gray, samples y = 0] table[x index = 0, y index = 1] {data/pf-background/2d.dat};
  	    \end{axis}
    \end{tikzpicture}
}}
\caption{Comparison of population distributions when decreasing the number of objectives on DTLZ2.}
\label{fig:DimDecrease}
\end{figure}

Here, we run MOEA/D on a three-objective DTLZ2 instance for 300 generations\footnote{All parameters are set the same as those in~\pref{sec:increasing}, except for the population size which is set to 105.} and obtain a reasonably good approximation to the PF as shown in~\pref{fig:DimDecrease}(a). Then, we change the underlying problem to a two-objective instance. \pref{fig:DimDecrease}(b) shows the corresponding population distribution after re-evaluating the objective values. In contrast to the scenario discussed in~\pref{sec:increasing}, the most direct effect of decreasing the number of objectives is the contraction of the PF or PS manifold. In addition, different from increasing the number of objectives, some solutions still stay on the PF after the objective reduction; while the others are moved away from the PF. Furthermore, although the spread of the population is not affected by decreasing an objective, the population diversity is not satisfied any longer (as shown in~\pref{fig:DimDecrease}(b), there are many duplicate or similar solutions in the reduced objective space). These characteristics make the algorithm design difficult on both pulling the drifted solutions back to the PF and propelling the solutions jump out from their duplicates.

\subsection{Weaknesses of Existing Dynamic Handling Techniques}
\label{sec:pitfalls}

As discussed in~\pref{sec:introduction}, the existing dynamic handling techniques do not take the changing number of objectives into consideration. They have the following weaknesses when encountering this sort of dynamics.
\begin{itemize}
    \item As introduced in~\pref{sec:introduction}, the diversity enhancement techniques use some randomly or heuristically injected candidates to adapt to the changing environment. However, when increasing the number of objectives, the injected candidates can hardly compete with the original population whose convergence has not be influenced. On the other hand, when decreasing the number of objectives, although the injected candidates are able to enhance the population diversity to a certain extent, they can hardly provide any selection pressure towards the PF.
    \item The memory mechanism has been used in~\cite{GuanCM05} to handle the DMOP with a changing number of objectives. In particular, they suggested an inheritance strategy that simply re-evaluates the objective values of the original population whenever the change of the environment is detected. Obviously, this strategy is too simple to handle problems with complicated properties.
    \item The prediction strategy, which takes advantages of the historical information, seems to be an antidote for dynamic optimization. But unfortunately, the existing prediction strategies are usually proposed to anticipate the movement of the PS when the environment changes. However, as discussed in~\pref{sec:challenges}, the change of the number of objectives usually leads to the expansion or contraction of the PS manifold, rather than the movement of its position. Thus, the prediction, based on the historical manifold model, can be heavily erroneous.
\end{itemize}

% !Tex root = main.tex

\section{Proposed Algorithm}
\label{sec:proposal}

\begin{figure*}[htbp]
\centering
\includegraphics[width=1.0\linewidth]{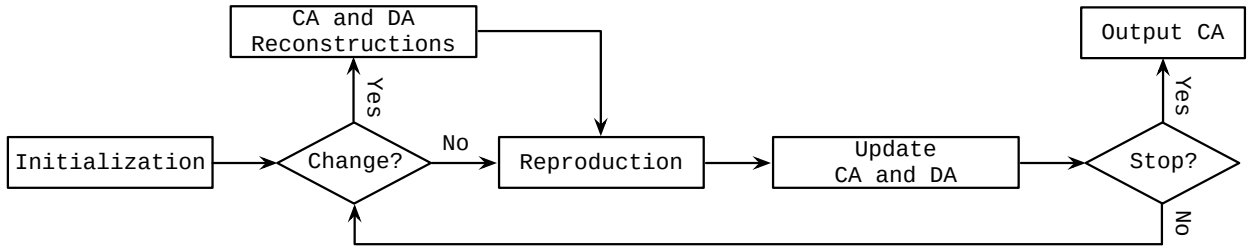}
\caption{Flow chart of DTAEA.}
\label{fig:flowchart}
\end{figure*}

In this section, we develop a dynamic two-archive EA (denoted as DTAEA), whose high-level flow chart is shown in~\pref{fig:flowchart}, for solving the DMOP with a changing number of objectives. Similar to its ancestors~\cite{PraditwongY06,LiLTY14} and~\cite{WangJY15}, DTAEA maintains two co-evolving populations simultaneously: one, called the convergence archive (CA), is used to provide a constantly competitive selection pressure towards the optima; the other, called the diversity archive (DA), is used to provide diversified solutions as much as possible. It is worth noting that the size of the CA and the DA is equal and fixed to a constant $N$ a priori. In each iteration, if a change of the environment is not occurred, mating parents are separately selected from the CA and the DA for offspring reproduction according to the mechanism described in~\pref{sec:reproduction}. Afterwards, the offspring is used to update the CA and the DA according to the mechanisms described in~\pref{sec:update}. On the other hand, the CA and the DA are reconstructed according to the mechanism described in~\pref{sec:reconstruction} whenever the underlying environment changes. At the end, all solutions of the CA are used to form the final population.

\subsection{Reconstruction Mechanisms}
\label{sec:reconstruction}

This step is the main crux to respond to the changing environment. Generally speaking, its basic idea is to reconstruct the CA and the DA whenever the environment changes. Since the challenges posed by increasing and decreasing the number of objectives are different, as discussed in~\pref{sec:challenges}, the reconstruction mechanisms are different accordingly.

\subsubsection{When Increasing the Number of Objectives}
\label{sec:reconstructionIns}

As discussed in~\pref{sec:increasing}, the convergence of the population is not affected when increasing the number of objectives; while the population diversity becomes severely inadequate to approximate the expanded PF or PS manifold. To avoid sacrificing any selection pressure towards the optima, we use all optimal solutions in the last CA before increasing the number of objectives to form the new CA. In the meanwhile, all members of the DA are replaced by the randomly generated solutions. In particular, to provide as much diversified solutions as possible, here we employ the canonical Latin hypercube sampling (LHS) method~\cite{McKayBC1979} to uniformly sample $N$ random solutions from the decision space. The pseudo-code of this reconstruction mechanism is given in~\pref{alg:updateInc}.

\begin{algorithm}[htbp]
\KwIn{The last $\mathrm{CA}$ before decreasing the number of objectives $\mathrm{P}$}
\KwOut{$\mathrm{CA}$, $\mathrm{DA}$}

$\mathrm{CA}\leftarrow\mathrm{P}$;\\
Use the LHS to sample $N$ random solutions to form the $\mathrm{DA}$;\\

\Return $\mathrm{CA}$ and $\mathrm{DA}$
\caption{Reconstruction Mechanism After Increasing the Number of Objectives}
\label{alg:updateInc}
\end{algorithm}

\subsubsection{When Decreasing the Number of Objectives}
\label{sec:reconstructionDec}

As discussed in~\pref{sec:decreasing}, the convergence of some, if not all, population members might be impaired when decreasing the number of objectives; while the population diversity is also adversely affected due the existence of many similar or duplicate solutions in the reduced objective space. To keep the selection pressure towards the optima as much as possible, all non-dominated solutions in the last CA before decreasing the number of objectives are used to form the new CA. If the new CA is not full, we perform the polynomial mutation~\cite{PolyMutation} on some selected competitive non-dominated solutions to generate some mutants to fill the gap in the new CA. In particular, the non-dominated solutions used for mutation are chosen by a binary tournament selection, where the non-dominated solution located in a less crowded area is preferred. Note that the density estimation method will be discussed in detail in~\pref{sec:density}. Since decreasing the number of objectives does not affect the spread of the population, instead of reconstructing the DA from scratch, we at first use all the dominated solutions in the last CA before decreasing the number of objectives to form the new DA. To provide enough diversified information to help the solutions jump out from their duplicates, we use the canonical LHS method to generate $N-|DA|$ random solutions to fill the gap in the new DA. The pseudo-code of this reconstruction mechanism is given in~\pref{alg:updateDec}.

\begin{algorithm}[htbp]
\KwIn{The last $\mathrm{CA}$ before decreasing the number of objectives $\mathrm{P}$}
\KwOut{$\mathrm{CA}$, $\mathrm{DA}$}

$\mathrm{CA}\leftarrow$ \texttt{NonDominationSelection}$(\mathrm{P})$\tcp*{choose all non-dominated solutions}
$\mathrm{DA}\leftarrow\mathrm{P}\setminus\mathrm{CA}$;\\
\While{$|\mathrm{CA}|<N$}{
    $\mathbf{x}^c\leftarrow$ \texttt{BirnaryTournamentSelection}$(\mathrm{CA})$;\\
    $\mathbf{x}^m\leftarrow$ \texttt{PolynomialMutation}$(\mathrm{x}^c)$;\\
    $\mathrm{CA}\leftarrow\mathrm{CA}\bigcup\{\mathbf{x}^m\}$;\\
}
Use the LHS to sample $N-|\mathrm{DA}|$ random solutions to fill the $\mathrm{DA}$;\\
\Return $\mathrm{CA}$ and $\mathrm{DA}$
\caption{Reconstruction Mechanism After Decreasing the Number of Objectives}
\label{alg:updateDec}
\end{algorithm}

\begin{algorithm}[htbp]
\KwIn{Solution set $\mathrm{S}$}
\KwOut{Selected solution $\mathbf{x}^c$}

Randomly select two solutions $\mathbf{x}^1$ and $\mathbf{x}^2$ from $\mathrm{S}$;\\
\uIf{$\mathtt{Density}(\mathbf{x}^1)<\mathtt{Density}(\mathbf{x}^2)$}{
    $\mathbf{x}^c\leftarrow\mathbf{x}^1$;\\
}\uElseIf{$\mathtt{Density}(\mathbf{x}^1)>\mathtt{Density}(\mathbf{x}^2)$}{
    $\mathbf{x}^c\leftarrow\mathbf{x}^2$;\\
}\Else{
    $\mathbf{x}^c\leftarrow$ Randomly pick one between $\mathbf{x}^1$ and $\mathbf{x}^2$;\\
}

\Return $\mathbf{x}^c$

\caption{Binary Tournament Selection}
\label{alg:binaryTournamentSelection}
\end{algorithm}

\subsection{Update Mechanisms}
\label{sec:update}

As described at the outset of this section, the CA and the DA have complementary effects on the search process. Accordingly, their update mechanisms are different and will be described in~\pref{sec:updateCA} and~\pref{sec:updateDA}, separately. Before that, \pref{sec:density} will at first introduces the density estimation method used in the update mechanisms.

\subsubsection{Density Estimation Method}
\label{sec:density}

To facilitate the density estimation, we at first specify $N(t)$ uniformly distributed weight vectors, i.e., $\mathrm{W}(t)=\{\mathbf{w}^1(t),\cdots,\mathbf{w}^{N(t)}(t)\}$, in $\mathbb{R}^{m(t)}$. In particular, we employ the weight vector generation method developed in~\cite{LiDZK15} for this purpose, since it is scalable to the many-objective scenarios. Note that the number of weight vectors at time step $t$ might not be equal to the size of the CA and the DA, i.e., $N(t)\leq N$. In this case, a weight vector might be associated with more than one solution if $N(t)<N$. Accordingly, these weight vectors divide $\mathbb{R}^{m(t)}$ into $N(t)$ subspaces, i.e., $\Delta^1(t),\cdots,\Delta^{N(t)}(t)$. In particular, a subspace $\Delta^i(t)$, where $i\in\{1,\cdots,N(t)\}$ is defined as:
\begin{equation}
\Delta^i(t)=\{\mathbf{F}(\mathbf{x},t)\in\mathbb{R}^{m(t)}|\langle\mathbf{F}(\mathbf{x},t),\mathbf{w}^i(t)\rangle\leq\langle\mathbf{F}(\mathbf{x},t),\mathbf{w}^j(t)\rangle\}
\end{equation}
where $j\in\{1,\cdots,N(t)\}$ and $\langle\mathbf{F}(\mathbf{x},t),\mathbf{w}(t)\rangle$ is the perpendicular distance between $\mathbf{F}(\mathbf{x},t)$ and the reference line formed by the origin and $\mathbf{w}(t)$. After the setup of subspaces, each solution of the underlying population is associated with an unique subspace according to its position in $\mathbb{R}^{m(t)}$. Specifically, for a solution $\mathbf{x}$, the index of its associated subspace is determined as:
\begin{equation}
k=\argmin\limits_{i\in\{1,\cdots,N(t)\}}\langle\mathbf{\overline{F}}(\mathbf{x},t),\mathbf{w}^i(t)\rangle
\end{equation}
where $\mathbf{\overline{F}}(\mathbf{x},t)$ is the normalized objective vector of $\mathbf{x}$, and its $i$-th objective function is calculated as:
\begin{equation}
\overline{f}_i(\mathbf{x},t)=\frac{f_i(\mathbf{x},t)-z^{\ast}_i(t)}{z^{nad}_i(t)-z^{\ast}_i(t)}
\end{equation}
where $i\in\{1,\dotsm,m(t)\}$. Based on the association relationship between solutions and subspaces, the density of a subspace is evaluated as the number of its associated solutions. \pref{fig:updateCA} gives a simple example to illustrate this density estimation method. In particular, five weight vectors divide the underlying objective space into five subspaces, where the corresponding density of each subspace is 2, 3, 1, 1 and 3, respectively.

%\begin{figure}[htbp]
%\centering
%\includegraphics[width=0.5\linewidth]{figs/density.eps}
%\caption{An illustration of the density estimation method.}
%\label{fig:density}
%\end{figure}

\subsubsection{Update Mechanism of the CA}
\label{sec:updateCA}

\begin{figure}[htbp]
\centering
\includegraphics[width=0.5\linewidth]{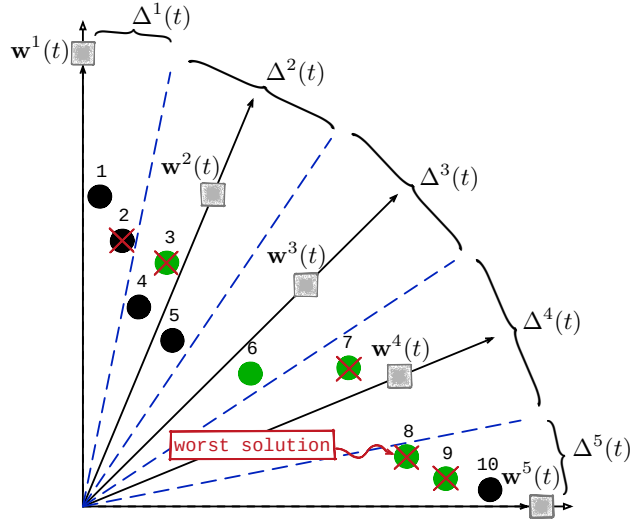}
\caption{An example of the CA's update mechanism. Note that $\mathbf{x}^i$ is denoted as its index $i$ for short.}
\label{fig:updateCA}
\end{figure}

The effect of the CA is to provide a constantly competitive selection pressure towards the optima. To achieve this, we at first combine the CA and the offspring population into a mixed population $\mathrm{R}$. Afterwards, we employ the fast non-dominated sorting method developed in~\cite{DebAPM2002} to divide $\mathrm{R}$ into several non-domination levels, i.e., $F_1$, $F_2$ and so on. Then, starting from $F_1$, each non-domination level is selected one at a time to construct the new CA. This procedure continues until the size of the new CA equals to or for the first time exceeds its predefined threshold. Let us denote the last included non-domination level as $F_l$, while all solutions from $F_{l+1}$ onwards are not taken into consideration any longer. If the size of the new CA equals its predefined threshold, this update procedure terminates. Otherwise, we use the density estimation method introduced in~\pref{sec:density} to evaluate the density information of the current CA. Then, we eliminate one worst solution from the most crowded subspace at a time until the CA is trimmed to its predefined threshold. In particular, for a subspace $\Delta^i(t)$, the worst solution $\mathbf{x}^w$ is defined as:
\begin{equation}
\mathbf{x}^w=\argmax\limits_{\mathbf{x}\in\Delta^i(t)}\{g^{tch}(\mathbf{x}|\mathbf{w}^i(t),\mathbf{z}^{\ast}(t))\}
\end{equation}
where
\begin{equation}
g^{tch}(\mathbf{x}|\mathbf{w}^i(t),\mathbf{z}^{\ast}(t))=\max\limits_{1\leq j\leq m(t)}\{|f_j(\mathbf{x},t)-z^{\ast}_j(t)|/w_j^i(t)\}.
\end{equation}
The pseudo-code of the update mechanism of the CA is given in~\pref{alg:updateCA}. \pref{fig:updateCA} gives an example to illustrate this update mechanism. Assume that the black and green circles indicate the solutions of the original CA and the offspring population respectively. Since the first non-domination level already contains 8 solutions, $\mathbf{x}^3$ and $\mathbf{x}^7$, which belong to the second non-domination level, are not considered any longer. In the first iteration, the worst solution $\mathbf{x}^8$ in the most crowded subspace, i.e., $\Delta^5(t)$, is eliminated. Analogously, $\mathbf{x}^2$ and $\mathbf{x}^9$ will be eliminated in the latter iterations.

\begin{algorithm}[htbp]
\KwIn{$\mathrm{CA}$, offspring population $\mathrm{Q}$, weight vector set $\mathrm{W}$}
\KwOut{Updated $\mathrm{CA}$}

$\mathrm{S}\leftarrow\emptyset$, $i\leftarrow 1$;\\
$\mathrm{R}\leftarrow\mathrm{CA}\bigcup\mathrm{Q}$;\\
$\{F_1,F_2,\cdots\}\leftarrow\texttt{NonDominatedSorting}(\mathrm{R})$;\\

\While{$|\mathrm{S}|\leq N$}{
    $\mathrm{S}\leftarrow\mathrm{S}\bigcup F_i$, $i\leftarrow i+1$;\\
}

\uIf{$|\mathrm{S}|=N$}{
    $\mathrm{CA}\leftarrow\mathrm{S}$;
}\Else{
    \ForEach{$\mathbf{x}\in\mathrm{S}$}{
        $\overline{\mathbf{F}}_k(\mathbf{x},t)=\frac{\mathbf{F}(\mathbf{x},t)-\mathbf{z}^{\ast}(t)}{\mathbf{z}^{nad}(t)-\mathbf{z}^{\ast}(t)}$;
    }
    $\{\Delta^1(t),\cdots,\Delta^{|\mathrm{W}|}(t)\}\leftarrow$ \texttt{Association}$(\mathrm{S},\mathrm{W})$;\\
    \While{$|\mathrm{S}|>N$}{
        Find the most crowded subspace $\Delta^i(t)$;\\
        $\mathbf{x}^w\leftarrow\argmax\limits_{\mathbf{x}\in\Delta^i(t)}\{g^{tch}(\mathbf{x}|\mathbf{w}^i(t),\mathbf{z}^{\ast}(t))\}$;\\
        $\mathrm{S}\leftarrow\mathrm{S}\setminus\{\mathbf{x}^w\}$;
    }
    $\mathrm{CA}\leftarrow\mathrm{S}$;
}

\Return $\mathrm{CA}$

\caption{Update Mechanism of the CA}
\label{alg:updateCA}
\end{algorithm}

\begin{algorithm}[h]
\KwIn{Solution set $\mathrm{S}$, weight vector set $\mathrm{W}$}
\KwOut{Subspaces $\Delta^1(t),\cdots,\Delta^{|\mathrm{W}|}(t)$}

\ForEach{$\mathbf{x}\in\mathrm{S}$}{
    \ForEach{$\mathbf{w}\in\mathrm{W}$}{
        Compute $d^{\perp}(\mathbf{x},\mathbf{w})=\mathbf{x}-\mathbf{w}^T\mathbf{x}/\|\mathbf{w}\|$;
    }
    $k\leftarrow\argmin\limits_{\mathbf{w}\in W}d^{\perp}(\mathbf{x},\mathbf{w})$;\\
    $\Delta^k(t)\leftarrow\Delta^k(t)\bigcup\{\mathbf{x}\}$;
}

\Return $\Delta^1(t),\cdots,\Delta^{|\mathrm{W}|}(t)$

\caption{Association Operation}
\label{alg:association}
\end{algorithm}

\subsubsection{Update Mechanism of the DA}
\label{sec:updateDA}

\begin{figure}[h]
\centering
\includegraphics[width=0.5\linewidth]{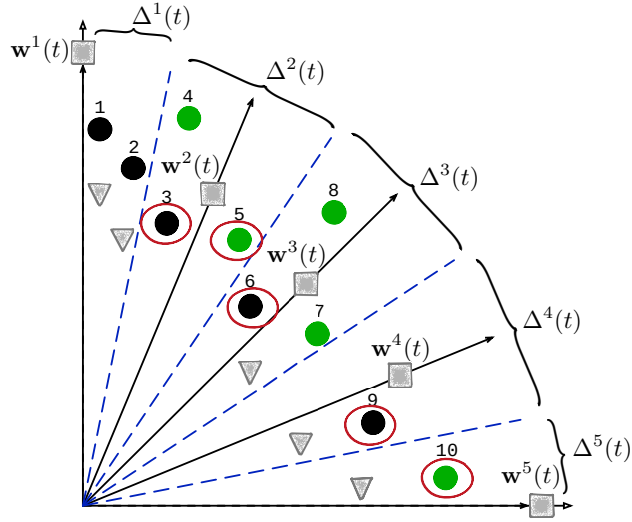}
\caption{An example of the DA's update mechanism. Note that $\mathbf{x}^i$ is denoted as its index $i$ for short.}
\label{fig:updateDA}
\end{figure}

Comparing to the CA, the DA has a complementary effect which aims at providing diversified solutions as much as possible, especially in the areas under exploited by the CA. Similar to the update mechanism of the CA, we at first combine the DA and the offspring population into a mixed population $\mathrm{R}$; in the meanwhile, we also take the up to date CA as the reference set. Afterwards, we employ the density estimation method introduced in~\pref{sec:density} to build up the association relationship between solutions in $\mathrm{R}$ and the subspaces. Then, based on the association relationship, we iteratively investigate each subspace for which we try to keep $\mathsf{itr}$ $(1\leq\mathsf{itr}\leq N)$ solutions at the $\mathsf{itr}$-th iteration. In particular, at the $\mathsf{itr}$-th iteration, if the CA already has $\mathsf{itr}$ solutions or there is no solution in $\mathrm{R}$ associated with the currently investigating subspace, we stop considering this subspace during this iteration and move to the next subspace directly. Otherwise, the best non-dominated solution in $\mathrm{R}$ associated with this currently investigating subspace is chosen to be included into the newly formed DA. In particular, the best solution $\mathbf{x}^b$ of the currently investigating subspace $\Delta^c(t)$ is defined as:
\begin{equation}
\mathbf{x}^b=\argmin\limits_{\mathbf{x}\in\mathrm{O}}\{g^{tch}(\mathbf{x}|\mathbf{w}^c(t),\mathbf{z}^{\ast}(t))\}
\end{equation}
where $\mathrm{O}$ constitutes of the non-dominated solutions in $\mathrm{R}$ associated with $\Delta^c(t)$. This iterative investigation continues until the DA is filled to its predefined threshold. The pseudo-code of the update mechanism of the DA is given in~\pref{alg:updateDA}. \pref{fig:updateDA} gives an example to illustrate this update mechanism. Assume that the gray triangles represent the solutions of the up to date CA while the black and green circles indicate the solutions of the original DA and the offspring population. At the first iteration, the update mechanism of the DA starts from $\Delta^1(t)$. Since the CA already has two solutions in this subspace, we move to investigate $\Delta^2(t)$. As the CA does not has any solution in this subspace, the best non-dominated solution in this subspace, i.e., $\mathbf{x}^3$ is included into the newly formed DA. Thereafter, since $\Delta^3(t)$ to $\Delta^5(t)$ already have a solution in the CA, they are not considered during this iteration. At the second iteration, since the CA already has two solutions in $\Delta^1(t)$, it is still not considered during this iteration. As for $\Delta^2(t)$, since the CA does not has any solution in this subspace, the remaining best solution $\mathbf{x}^5$ is selected this time. Then, since $\Delta^3(t)$ to $\Delta^5(t)$ only have one solution in the CA, we can choose the best solution from $\mathrm{R}$ to be included into the newly formed DA this time. At the end of the second iteration, the DA is filled and the update procedure terminates.

\begin{algorithm}[htbp]
\KwIn{$\mathrm{CA}$, $\mathrm{DA}$, offspring population $\mathrm{Q}$, weight vector set $\mathrm{W}$}
\KwOut{Updated $\mathrm{DA}$}

$\mathrm{S}\leftarrow\emptyset$, $i\leftarrow 1$;\\
$\mathrm{R}\leftarrow\mathrm{DA}\bigcup\mathrm{Q}$;\\

$\{\Delta^1(t),\cdots,\Delta^{|\mathrm{W}|}(t)\}\leftarrow$ \texttt{Association}$(\mathrm{R},\mathrm{W})$;\\

$\mathsf{itr}\leftarrow 1$;\\
\While{$|\mathrm{S}|\leq N$}{
    \For{$i\leftarrow 1$ \KwTo $|\mathrm{W}|$}{
        \If{$\Delta^i(t)\neq\emptyset$ $\wedge$ $\mathrm{CA}$ has less than $\mathsf{itr}$ solutions in $\Delta^i(t)$}{
            $\mathrm{O}\leftarrow$ \texttt{NonDominationSelection}($\Delta^i(t)$);\\
            $\mathbf{x}^b\leftarrow\argmin\limits_{\mathbf{x}\in\mathrm{O}}\{g^{tch}(\mathbf{x}|\mathbf{w}^c(t),\mathbf{z}^{\ast}(t))\}$;\\
            $\Delta^i(t)\leftarrow\Delta^i(t)\setminus\{\mathbf{x}^b\}$;\\
            $\mathrm{S}\leftarrow\mathrm{S}\bigcup\{\mathbf{x}^b\}$;
        }
    }
    $\mathsf{itr}\leftarrow \mathsf{itr}+1$;
}
$\mathrm{DA}\leftarrow\mathrm{S}$;

\Return $\mathrm{DA}$

\caption{Update Mechanism of the DA}
\label{alg:updateDA}
\end{algorithm}

\subsection{Offspring Reproduction}
\label{sec:reproduction}

The interaction between two co-evolving populations is a vital step in the DTAEA. To take advantage of the complementary effects between the CA and the DA, here the interaction is implemented as a restricted mating selection mechanism that chooses the mating parents from these two co-evolving populations respectively according to the population distribution. It is worth noting that, comparing to the CA, the DA plays as an auxiliary which provides diversified information as much as possible and propels the adaptation to the changing environments. In addition, the CA and the DA become gradually asymptotic with the progress of the evolution. This results in a weakened complementary effects between these two co-evolving populations. Bearing these considerations in mind, the pseudo code of the restricted mating selection mechanism is given in~\pref{alg:mating}. More specifically, the first mating parent is randomly chosen from the CA. If the diversity of the CA is not promising, the second mating parent is randomly chosen from the DA; otherwise, it still comes from the CA. In particular, the diversity of the CA is measured by the occupation rate (denoted as $I^o_{\mathrm{CA}}$) which is the percentage of the subspaces having associated solutions. It is intuitive to understand that a high $I^o_{\mathrm{CA}}$ indicates a well diversified distribution in the CA. Accordingly, we should give a larger chance to select the second mating parent from the CA. Based on the selected mating parents, the offspring solution can be generated by the appropriate crossover and mutation operations as desired. In particular, we use the classic simulated binary crossover~\cite{DebA94} and the polynomial mutation in this paper.

\begin{algorithm}[h]

\KwIn{$\mathrm{CA}$, $\mathrm{DA}$}
\KwOut{Mating parents $\mathbf{p}_1$, $\mathbf{p}_2$}

$\mathbf{p}_1\leftarrow$ \texttt{RandomSelection}$(\mathrm{CA})$;\\
\uIf{$\mathsf{rnd}<I^o_{\mathrm{CA}}$}{
    $\mathbf{p}_2\leftarrow$ \texttt{RandomSelection}$(\mathrm{CA})$;\\
}\Else{
    $\mathbf{p}_2\leftarrow$ \texttt{RandomSelection}$(\mathrm{DA})$;\\
}

\Return $\mathbf{p}_1$, $\mathbf{p}_2$

\caption{Restricted Mating Selection Mechanism}
\label{alg:mating}
\end{algorithm}

\subsection{Time Complexity Analysis}
\label{sec:complexity}

This subsection discusses the complexity of DTAEA in one generation. The CA and the DA will be reconstructed once the environment changes, i.e., when the number of objectives is increased or decreased. In the former case, i.e., when increasing the number of objectives, the reconstruction of the CA does not incur any further computation while the reconstruction of the DA costs $\mathcal{O}(N)$ function evaluations. As for the latter case, i.e., when decreasing the number of objectives, the identification of the non-dominated solutions costs $\mathcal{O}(N^2)$ comparisons. Then, filling the gap of the CA and the DA may respectively take $\mathcal{O}(N)$ function evaluations. As for the update mechanism of the CA, the most time consuming operations are the non-dominated sorting (line 3 of~\pref{alg:updateCA}) and the association operation (line 11 of~\pref{alg:updateCA}). In particular, the prior one costs $\mathcal{O}(N^2)$ comparisons while the latter one takes $\mathcal{O}(N|W|)$, where $|W|$ returns the cardinality of a $W$. As for the update mechanism of the DA, the association operation in line 3 of~\pref{alg:updateDA} costs $\mathcal{O}(N|W|)$ comparisons. During the main while loop, the update procedure performs one by one for the subspaces. In particular, if $\Delta^i(t)\neq\emptyset$, where $i\in\{1,\cdots,|W|\}$, the most time consuming part of the for loop from line 6 to line 11 is the non-dominated sorting in line 8 of~\pref{alg:updateDA}, which costs $\mathcal{O}(|\Delta^i(t)|^2)$ comparisons. Since $\sum_{i=1}^{|W|}\Delta^i(t)=2N$, this for loop costs $\mathcal{O}(N^2)$ comparisons in total. In summary, the complexity of DTAEA in one generation is $\mathcal{O}(N^2)$.

\subsection{Further Discussions}
\label{sec:discussions}

As DTAEA is developed upon the decomposition-based framework, this subsection discusses the connections with some state-of-the-art decomposition-based EMO algorithms, MOEA/D-M2M~\cite{LiuGZ14}, NSGA-III~\cite{DebJ14}, MOEA/D-GR~\cite{WangZZGJ16} and MOEA/D-STM~\cite{LiZKLW14,LiKZD15,WuLKZZ17}. Similar to MOEA/D-M2M, DTAEA uses the weight vectors to divide the objective space into several subspaces. However, the major purpose of subspace division in MOEA/D-M2M is to enforce an equal distribution of the computational resources to each subspace; while the subspace division is used to facilitate the density estimation in DTAEA. Similar to NSGA-III, MOEA/D-GR and MOEA/D-STM, each weight vector in DTAEA is associated with its related solutions to facilitate the environmental selection. It uses different strategies to maintain its two co-evolving archives. In particular, according to the association relationship, the CA trims solutions from the most crowded subspaces; while the DA gives higher survival rates to solutions resided in the subspaces under exploited by the CA. In contrast, NSGA-III merely uses the association relationship to guide the truncation through the most crowded subspaces. In MOEA/D-GR and MOEA/D-STM, the next parent population is composed of the elite solutions from each of the non-empty subspace, which has at least one associated solution.

% !Tex root = main.tex

\section{Experimental Settings}
\label{sec:settings}

This section introduces the benchmark problems, performance metrics and the state-of-the-art EMO algorithms used in the experimental studies of this paper.

\subsection{Benchmark Problems}
\label{sec:benchmark}

Benchmark problems play important roles in assessing and analysing the performance of an algorithm, and thus guiding its further developments. Although some dynamic multi-objective benchmark problems have been proposed in the literature, e.g.,~\cite{FarinaDA04,HelbigE14} and~\cite{JiangY16}, most, if not all, of them merely consider the dynamically changing shape or position of the PF or PS. In~\cite{HuangSA11}, a dynamic multi-objective test instance with a changing number of objectives was developed, for the first time, by modifying a particular test instance from the classic DTLZ benchmark suite~\cite{DTLZ}. In this paper, we develop a series of dynamic multi-objective benchmark problems as shown in~\pref{tab:benchmark}. Note that, in addition to the changing number of objectives, F5 and F6 are also accompanied by a time-dependent change of the shape or position of the PF or PS.

\begin{table*}[t]
\centering
\footnotesize
\caption{Mathematical Definitions of Dynamic Multi-Objective Benchmark Problems}
\begin{tabular}{@{}c@{ }|c|c}
\toprule
Problem Instance             & Definition & Domain \\\midrule
\multirow{4}{*}{F1} & $f_1 = (1+g)0.5\prod_{i=1}^{m(t)-1}x_i$ & \multirow{4}{*}{$[0,1]$} \\ & $f_{j=2:m(t)-1}=(1+g)0.5(\prod_{i=1}^{m(t)-j}x_i)(1-x_{m(t)-j+1})$ & \\ & $f_{m(t)}=(1+g)0.5(1-x_1)$ & \\ & $g=100[n-m(t)+1+\sum_{i=m(t)}^n((x_i-0.5)^2-\cos(20\pi(x_i-0.5)))]$ & \\\midrule
\multirow{4}{*}{F2} & $f_1 = (1+g)0.5\prod_{i=1}^{m(t)-1}\cos(x_i\pi/2)$ & \multirow{4}{*}{$[0,1]$} \\ & $f_{j=2:m(t)-1}=(1+g)0.5(\prod_{i=1}^{m(t)-j}\cos(x_i\pi/2))(\sin(x_{m(t)-j+1}\pi/2))$ & \\ & $f_{m(t)}=(1+g)\sin(x_1\pi/2)$ & \\ & $g=\sum_{i=m(t)}^n(x_i-0.5)^2$ & \\\midrule
F3 & as F2, except $g$ is replaced by the one from F1 & $[0,1]$ \\\midrule
F4 & as F2, except $x_i$ is replaced by $x_i^{\alpha}$, where $i\in\{1,\cdots,m(t)-1\},\alpha>0$ & $[0,1]$ \\\midrule
\multirow{2}{*}{F5} & as F2, except $g=\sum_{i=m(t)}^n(x_i-G(\bar{t}))^2$ & \multirow{2}{*}{$[0,1]$} \\ & where $G(\bar{t})=|\sin(0.5\pi\bar{t})|$, $\bar{t}=\frac{1}{n_{\bar{t}}}\lfloor\frac{\tau}{\tau_{\bar{t}}}\rfloor$ & \\\midrule
    \multirow{3}{*}{F6} & as F2, except $g=G(\bar{t})+\sum_{i=m(t)}^n(x_i-G(\bar{t}))^2$ & \multirow{3}{*}{$[0,1]$} \\ & where $G(\bar{t})=|\sin(0.5\pi\bar{t})|$, $\bar{t}=\frac{1}{n_{\bar{t}}}\lfloor\frac{\tau}{\tau_{\bar{t}}}\rfloor$ & \\ & and $x_i$ is replaced by $x_i^{F(\bar{t})}$, where $i\in\{1,\cdots,m(t)-1\}$ and $F(\bar{t})=1+100\sin^4(0.5\pi\bar{t})$ & \\
\bottomrule
\end{tabular}
\label{tab:benchmark}
\end{table*}

\subsection{Performance Metrics}
\label{sec:metrics}

In the field of DMO, there is still no standard metric for quantitatively evaluating the performance of algorithms. In this paper, we employ the following two metrics, adapted from two popular metrics used in the literature, for the performance assessment.
%Most commonly used metrics in this field are adapted from the stationary multi-objective optimization. In this paper, we employ the following two popular performance metrics, both of which are able to evaluate the convergence and diversity simultaneously.

\begin{itemize}
	\item\textit{Mean Inverted Generational Distance (MIGD)}~\cite{ZhouJZ2014}: Let $P^{\ast}_t$ is a set of points uniformly sampled along the $PF_t$, $S_t$ is the set of solutions obtained by an EMO algorithm for approximating the $PF_t$, and $T$ is a set of discrete time steps in a run,
\begin{equation}
\textsc{MIGD}=\frac{1}{|T|}\sum_{t\in T}\textsc{IGD}(P^{\ast}_t,S_t).
\end{equation}
In particular, as defined in~\cite{BosmanT03},
\begin{equation}
\textsc{IGD}(P^{\ast}_t,P_t)=\frac{\sum_{\mathbf{x}\in P}dist(\mathbf{x},P^{\ast}_t)}{|S|},
\end{equation}
where $dist(\mathbf{x},P^{\ast}_t)$ is the Euclidean distance between a point $\mathbf{x}\in S_t$ and its nearest neighbor in $P^{\ast}_t$. Note that the calculation of IGD requires the prior knowledge of the PF. In our experiments, we use the method suggested in~\cite{LiDZK15} to sample 10,000 uniformly distributed points on the corresponding $PF_t$ at the $t$-th time step where $t\in T$.
    \item\textit{Mean Hypervolume (MHV)}: Let $\mathbf{z}^{w}_t=(z_{1}^{w},\ldots,z_{m(t)}^{w})^T$ be a worst point in the objective space that is dominated by all Pareto-optimal objective vectors at the $t$-th time step,
    \begin{equation}
    \textsc{MHV}=\frac{1}{|T|}\sum_{t\in T}\textsc{HV}(S_t).
    \label{eq:MHV}
    \end{equation}
In particular, as defined in~\cite{SPEA}, HV measures the size of the objective space dominated by solutions in $S_t$ and bounded by $\mathbf{z}^{w}_t$.
    \begin{equation}
    \textsc{HV}(S_t)=\textsc{VOL}(\bigcup_{\mathbf{x}\in S_t}[f_{1}(\mathbf{x}),z^{w}_{1}]\times\ldots[f_{m(t)}(\mathbf{x}),z^{w}_{m(t)}])
    \label{eq:HV}
    \end{equation}
   	where VOL$(\cdot)$ indicates the Lebesgue measure. Note that solutions, dominated by the worst point, are discarded for HV calculation. In our experiments, we set $\mathbf{z}^w_t=(\underbrace{2,\cdots,2}_{m(t)})^T$. For a better presentation, the HV values used in~\pref{eq:MHV} are normalized to $[0,1]$ by dividing $z_t=\prod_{i=1}^mz_i^{w}$.
\end{itemize}

\subsection{EMO Algorithms Used in the Experimental Studies}
\label{sec:algorithms}

In our empirical studies, four state-of-the-art EMO algorithms are used for comparisons: the dynamic version of the elitist non-dominated sorting genetic algorithm (DNSGA-II)~\cite{DebNK06} and MOEA/D with Kalman Filter prediction (MOEA/D-KF)~\cite{MurugananthaTV15}; and their corresponding stationary baseline NSGA-II~\cite{DebAPM2002} and MOEA/D~\cite{ZhangL2007}. They were chosen because of their popularity and good performance in both dynamic and static environments. Comparisons with the baseline algorithms are important. Because we want to check whether the dynamic algorithms outperform their static counterparts or not when handling the DMOP with a changing number of objectives. The following paragraphs provide some brief descriptions of these compared algorithms.
\begin{itemize}
    \item\textit{DNSGA-II}: To make the classic NSGA-II suitable for handling dynamic optimization problems, \cite{DebNK06} suggested to replace some population members with either randomly generated solutions or mutated solutions upon existing ones once a change occurs. As reported in~\cite{DebNK06}, the prior one performs better on DMOPs with severely changing environments while the latter one may work well on DMOPs with moderate changes. In our experiments, we adopt the prior DNSGA-II version in view of its slightly better performance reported in~\cite{DebNK06}.
    \item\textit{MOEA/D-KF}: This is a recently proposed prediction-based strategy that employs a linear discrete time Kalman Filter to model the movements of the PS in the dynamic environment. Thereafter, this model is used to predict the new location of the PS when a change occurs. Empirical results in~\cite{MurugananthaTV15} has shown that MOEA/D-KF is very competitive for the dynamic optimization and it outperforms the other state-of-the-art predictive strategies, e.g.,~\cite{ZhouJZ2014} and~\cite{WuJL15}.
    \item\textit{NSGA-II}: It at first uses non-dominated sorting to divide the population into several non-domination levels. Solutions in the first several levels have a high priority to be selected as the next parents. The exceeded solutions are trimmed according to the density information.
    \item\textit{MOEA/D}: This is a representative of the decomposition-based EMO methods. Its basic idea is to decompose the original MOP into several subproblems, either single-objective scalar functions or simplified MOPs. Thereafter, it employs some population-based techniques to solve these subproblems in a collaborative manner.
    %\item\textit{NSGA-III}: This a recently developed NSGA-II variant to handle problems with more than three objectives. The only difference lies in the density estimation method which uses several pre-defined reference points to specify niches in the objective space. Solutions in the less crowded niche has a higher priority to survive to the next generation.
\end{itemize}

Each algorithm is independently run 31 times on each problem instance. 300 generations are given to each algorithm before the first change. In other words, the first change occurs after the first 300 generations. To have a statistically sound conclusion, we use the Wilcoxon rank sum test at the 5\% significance level in the comparisons.

% !Tex root = main.tex

\section{Experimental Results}
\label{sec:empirical}

\subsection{Results on F1 to F4}
\label{sec:empirical-F14}

Let us at first consider F1 to F4 in which only the number of objectives changes with time. In particular, we define the time varying number of objectives $m(t)$ as follows:

\begin{equation}
m(t)=
    \begin{cases}
        3, & \quad t=1\\
        m(t-1)+1, & \quad t\in[2,5]\\
        m(t-1)-1, & \quad t\in[6,10]
    \end{cases}
\label{eq:mt}
\end{equation}
where $t\in\{1,\cdots,10\}$ is a discrete time. The population size is constantly set as 300 while the number of weight vectors is set according to~\pref{tab:weights}. To investigate the performance of different algorithms under various frequencies of change, we set $\tau_t$ as 25, 50, 100, 200, respectively. \pref{tab:MIGD} and~\pref{tab:MHV} give the median and the interquartile range (IQR) of the corresponding metric values obtained by different algorithms under various circumstances. In particular, the best metric values are highlighted in the bold face with a gray background. 
In addition to the metric values, we also keep a record of the ranks of the IGD and HV values obtained by different algorithms at each time step. We assign a global rank to each algorithm by averaging the ranks obtained at all time steps. The experimental results clearly demonstrate that DTAEA is the best optimizer as it wins on all comparisons (64 out of 64 for MIGD and 64 out of 64 for MHV), and it is always top ranked. In the following paragraphs, we will explain these results in detail.

\begin{table}[htbp]
\centering
\footnotesize
\caption{Number of Weight Vectors}
\label{tab:weights}

\begin{tabular}{c|c}
\toprule
m & \# of weight vectors \\
\midrule 
2 & 300	($H$ = 299)\\
\midrule 
3 & 300 ($H$ = 23)\\
\midrule
4 & 286 ($H$ = 10)\\
\midrule 
5 & 280 ($H_1$ = 6, $H_2$ = 4)\\
\midrule 
6 & 273 ($H_1$ = 5, $H_2$ = 2)\\
\midrule 
7 & 294 ($H_1$ = 4, $H_2$ = 3)\\
\bottomrule
\end{tabular}

\begin{tablenotes}
\item[1] $H$ is the number of divisions on each coordinate. Two-layer weight vector generation method is applied for 5- to 7-objective cases. $H_1$ and $H_2$ is the number of divisions for the boundary and inside layer, respectively.
\end{tablenotes}

\end{table}

F1 is developed from DTLZ1~\cite{DTLZ} which has a multi-modal property to hinder an algorithm from converging to the $PF_t$, $t\in\{1,\cdots,10\}$. As shown in~\pref{tab:MIGD} and~\pref{tab:MHV}, the performance of DTAEA is robust as it obtains the best MIGD and MHV values under all four frequencies of change. \pref{fig:trajectory}(a) shows the trajectories of IGD values obtained by different algorithms across the whole evolution process; while~\pref{fig:rank}(a) shows the average ranks of IGD obtained by different algorithms at each time step. From these two figures, we clearly see that DTAEA shows the best performance at every time step. It is worth noting that the performance of DTAEA has some fluctuations when $\tau_t=25$. This is because, under a high frequency of change, DTAEA can hardly drive the solutions fully converge to the $PF_t$ before the environment changes. However, with the decrease of the frequency of change, i.e., the increase of $\tau_t$, the performance of DTAEA becomes stable. As for the other four algorithms, the performance of different algorithms fluctuate a lot at different time steps. It is interesting to note that the performance of the stationary algorithms, i.e., NSGA-II and MOEA/D, are comparable to their dynamic counterparts, i.e., DNSGA-II and MOEA/D-KF. In particular, under a high frequency of change, i.e, $\tau_t=25$, NSGA-II and MOEA/D have shown a better performance than DNSGA-II and MOEA/D-KF at every time step. This implies that the dynamic handling techniques of DNSGA-II and MOEA/D-KF might not be capable of handling the expansion or contraction of the objective space. Even worse, these mechanisms can have an adverse effect to the population for adapting to the changing environments. In addition, we also notice that NSGA-II and DNSGA-II show better performance than MOEA/D and MOEA/D-KF at the first several time steps; whereas their performance degenerate significantly afterwards. This is because NSGA-II can have a faster convergence than MOEA/D when the number of objectives is relatively small. As discussed in~\pref{sec:challenges}, the increase of the number of objectives does not influence the population convergence. Thus, the competitive IGD values obtained at the 2- and 3-objective cases have an accumulative effect which makes NSGA-II and DNSGA-II maintain a competitive performance at the first couple of time steps. However, the ineffectiveness of NSGA-II and DNSGA-II for handling problems with a large number of objectives~\cite{KhareYD03} leads to their poor performance at the latter time steps. Even worse, their poor performance at 6- and 7-objective scenarios also disseminate a negative accumulation which results in their poor performance when decreasing the number of objectives.

F2 is developed from DTLZ2~\cite{DTLZ} which is a relatively simple benchmark problem. From~\pref{fig:trajectory}(b) and~\pref{fig:rank}(b), we also find that DTAEA has shown the consistently best performance across all time steps. It is interesting to note that the MIGD and MHV values obtained by all five algorithms lie in the same scale. This is because F2 does not pose too much challenge to the algorithms for converging to the $PF_t$. Thus all algorithms are able to adapt their populations to the changing environments within a reasonable number of function evaluations. Due to this reason, as shown in~\pref{fig:trajectory}(b), the IGD trajectories of NSGA-II and DNSGA-II surge up to a relatively high level when the number of objectives becomes large; whereas their IGD trajectories gradually go down when decreasing the number of objectives. Moreover, similar to the observations on F1, DNSGA-II and MOEA/D-KF do not show superior performance than their corresponding stationary counterparts on F2.

F3 is developed from DTLZ3~\cite{DTLZ} which has the same $PF_t$, $t\in\{1,\cdots,10\}$, as F2 but has a multi-modal property. Due to the multi-modal property, which hinders the population from approaching the $PF_t$, the IGD trajectories of NSGA-II and DNSGA-II can hardly drop down when decreasing the number of objectives. In the meanwhile, we still notice that the dynamic handling techniques of DNGSA-II and MOEA/D-KF do not help the population adapt to the changed environments. Even worse, the injected solutions provide a false information about the newly changed PF or PS, which is harmful to the evolution process.

F4 is developed from DTLZ4~\cite{DTLZ} which also has the same $PF_t$, $t\in\{1,\cdots,10\}$, as F2 but has a parametric mapping that introduces a biased density of Pareto-optimal solutions towards some particular coordinates. In this case, F4 not only poses significant challenges for handling the changing number of objectives, but also requires that the algorithm can have a well balance between convergence and diversity. The superior performance of DTAEA can be attributed to the dedicated operations of two co-evolving populations for balancing convergence and diversity during the whole evolution process. Similar to the observations on the previous problem instances, the stationary algorithms still show competitive performance than their dynamic counterparts at most time steps.

\subsection{Results on F5 and F6}
\label{sec:empirical-F56}

Different from F1 to F4, F5 and F6 not only consider a changing number of objectives, but also have a time varying PS. Here we use~\pref{eq:mt} to define the time varying number of objectives; as for the parameters related to the time varying PS, we set its change frequency as $\tau_{\bar{t}}=5$ and change severity as $n_{\bar{t}}=10$ (these settings are widely used in the literature, e.g.,~\cite{GohT09,ShangJRWL14} and~\cite{JiangY16TEC}). From the experimental results shown in~\pref{tab:MIGD} and~\pref{tab:MHV}, we find that DTAEA has shown the best performance on almost all comparisons (30 out of 32 for MIGD and 28 out of 32 for MHV). In the following paragraphs, we will explain these results in detail.

\begin{figure*}[htbp]
\centering
    \subfloat[F1]{\includegraphics[width=.5\linewidth]{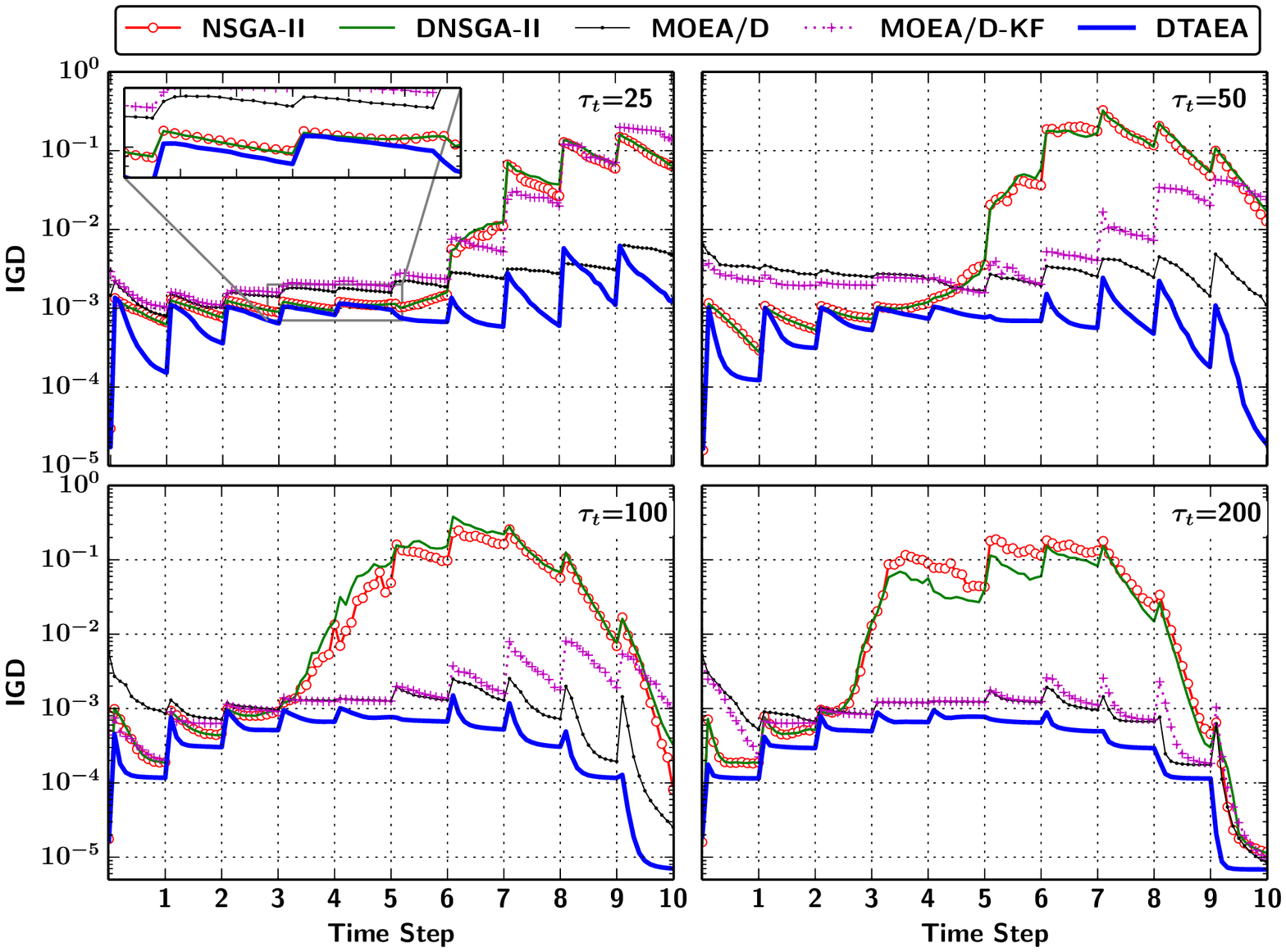}}
    \subfloat[F2]{\includegraphics[width=.5\linewidth]{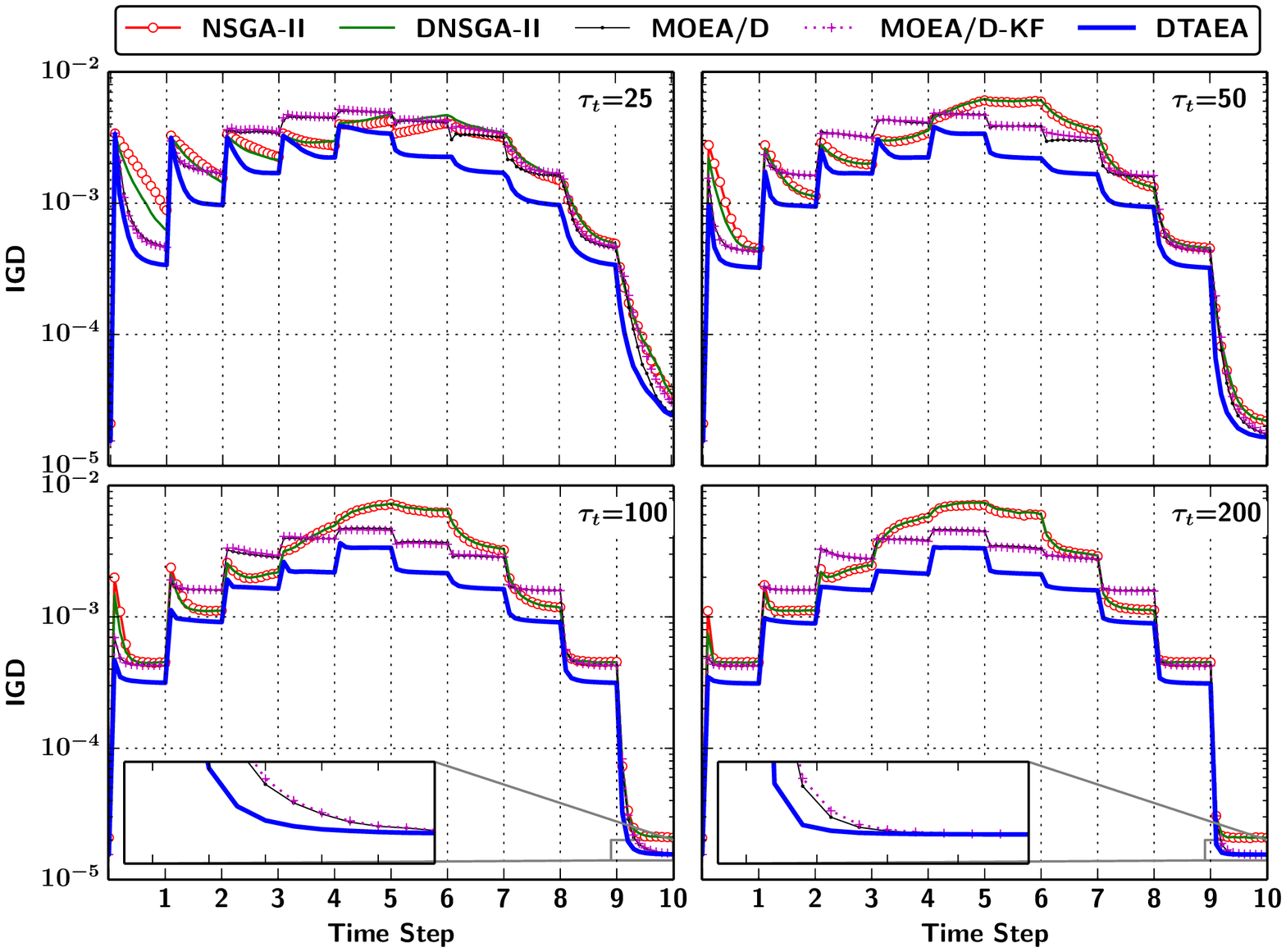}}\\
    \subfloat[F3]{\includegraphics[width=.5\linewidth]{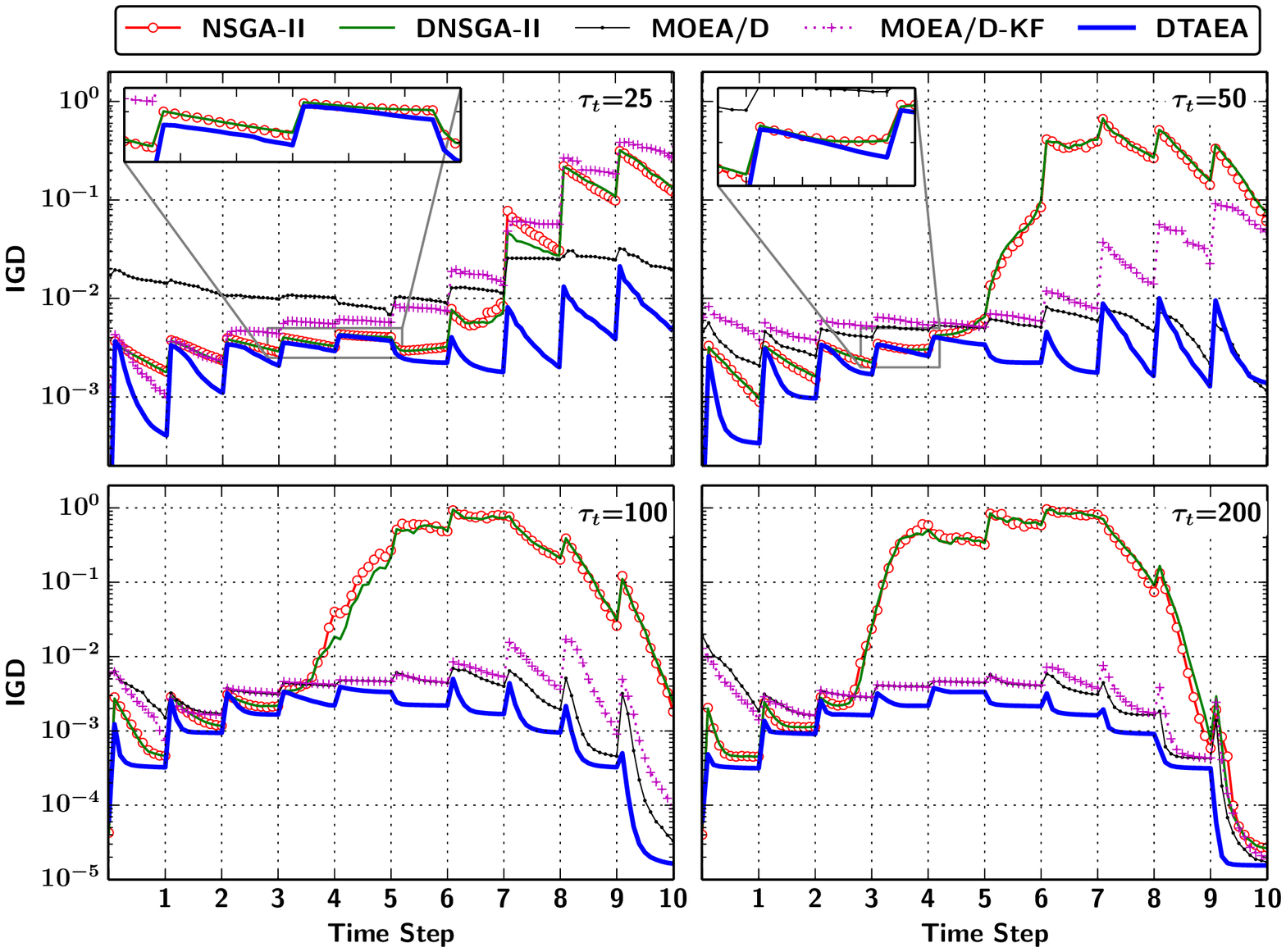}}
    \subfloat[F4]{\includegraphics[width=.5\linewidth]{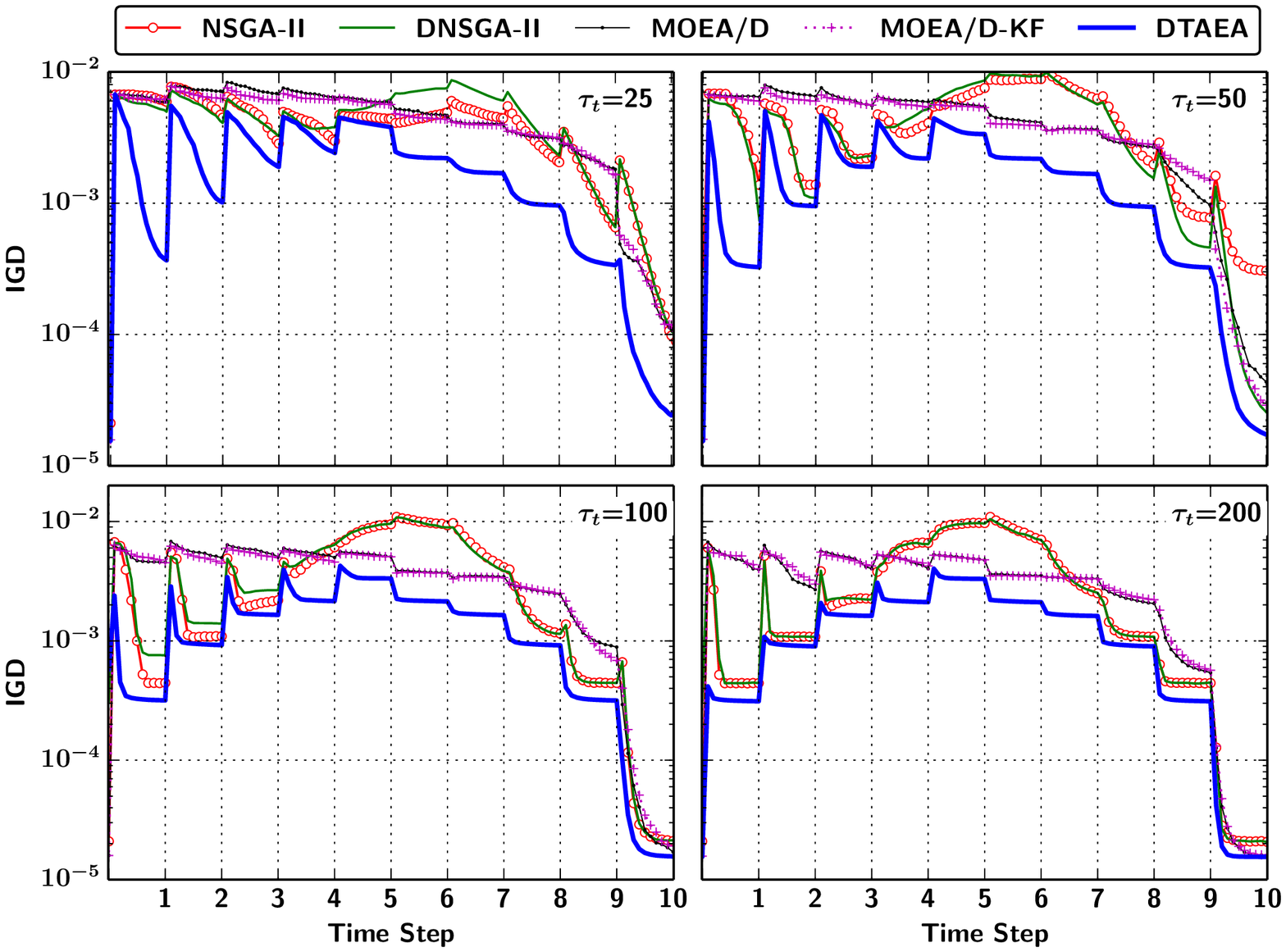}}\\
    \subfloat[F5]{\includegraphics[width=.5\linewidth]{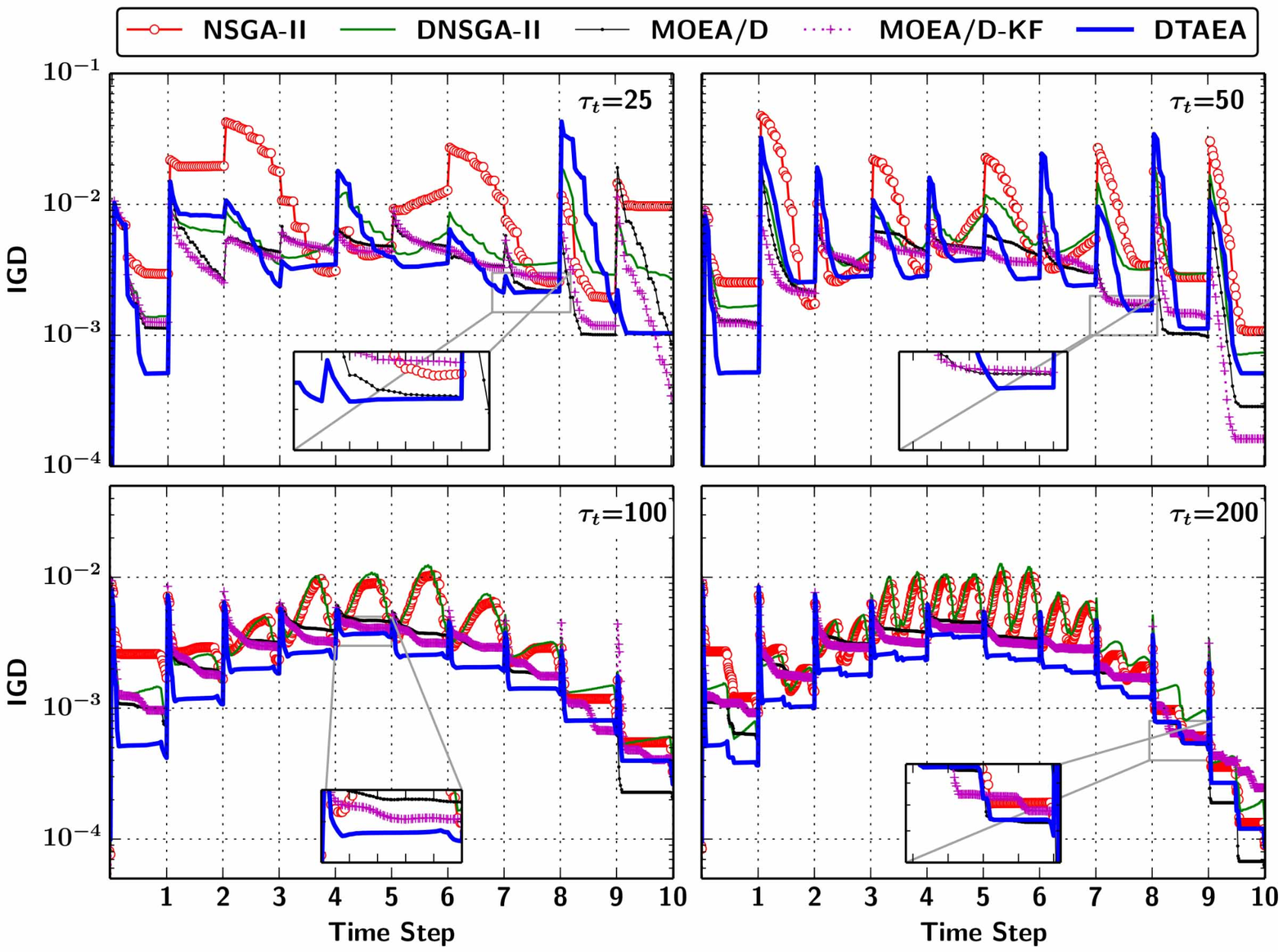}}
    \subfloat[F6]{\includegraphics[width=.5\linewidth]{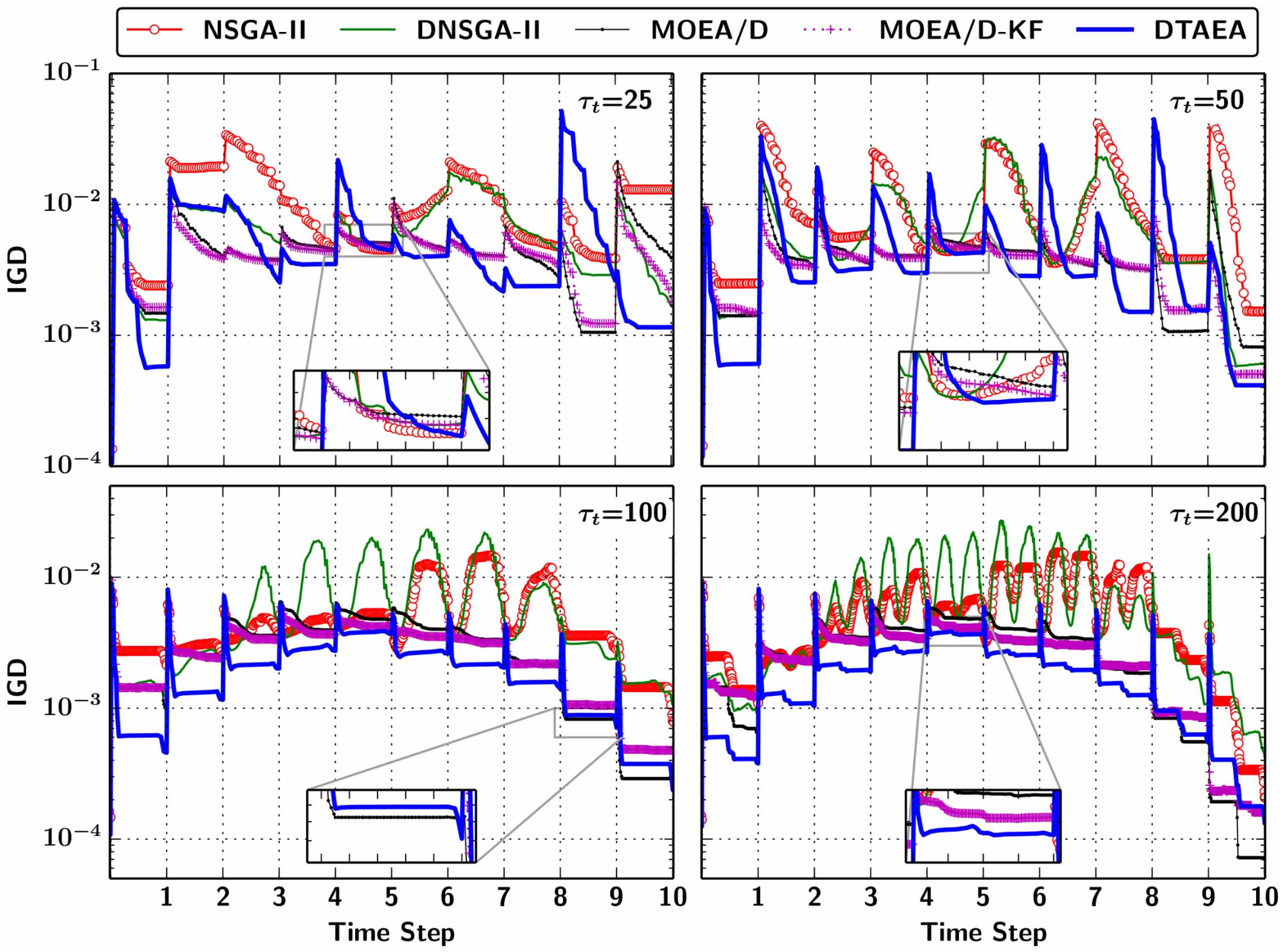}}
\caption{IGD trajectories across the whole evolution process.}
\label{fig:trajectory}
\end{figure*}

\begin{figure*}[htbp]
\centering
    \subfloat[F1]{\includegraphics[width=.5\linewidth]{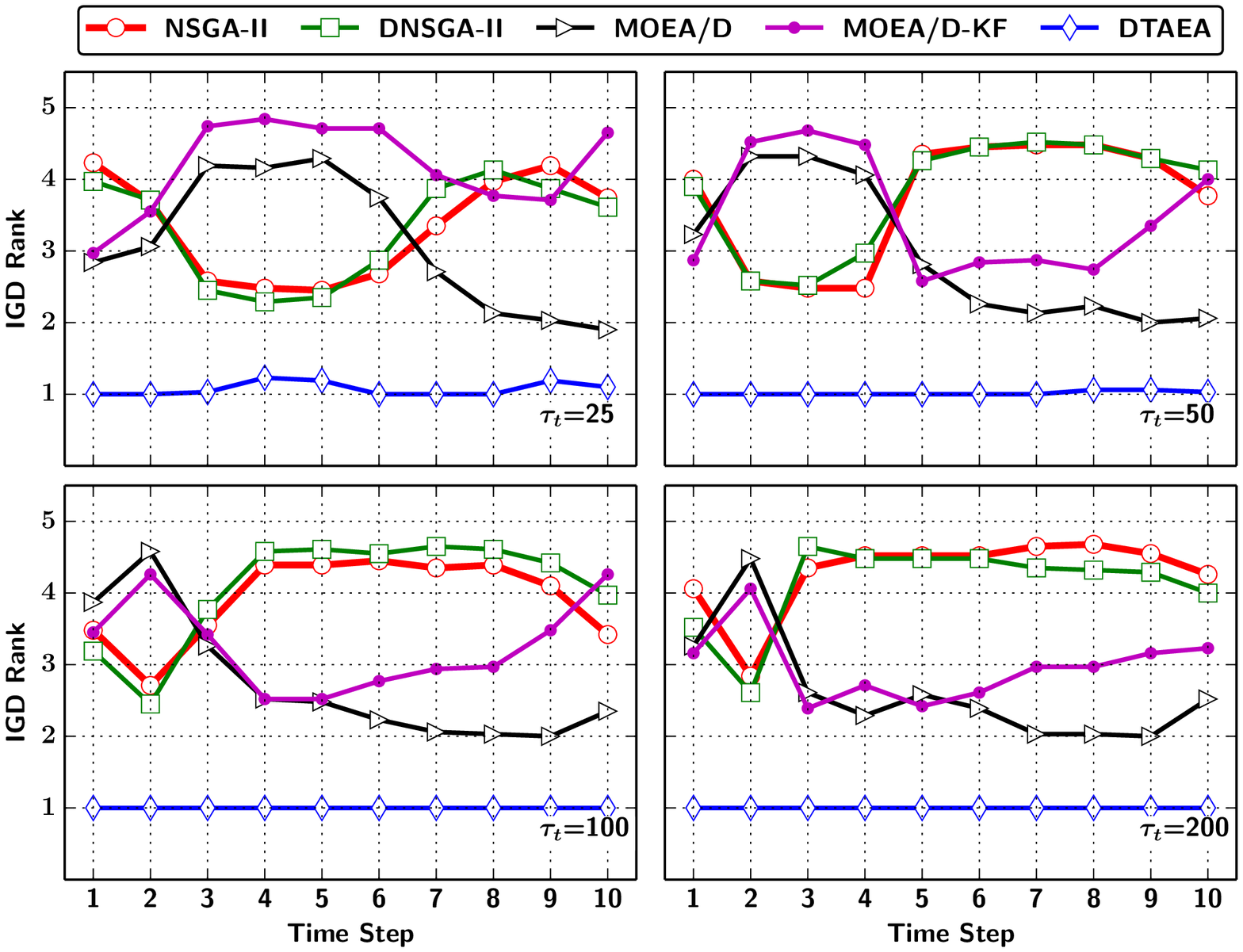}}
    \subfloat[F2]{\includegraphics[width=.5\linewidth]{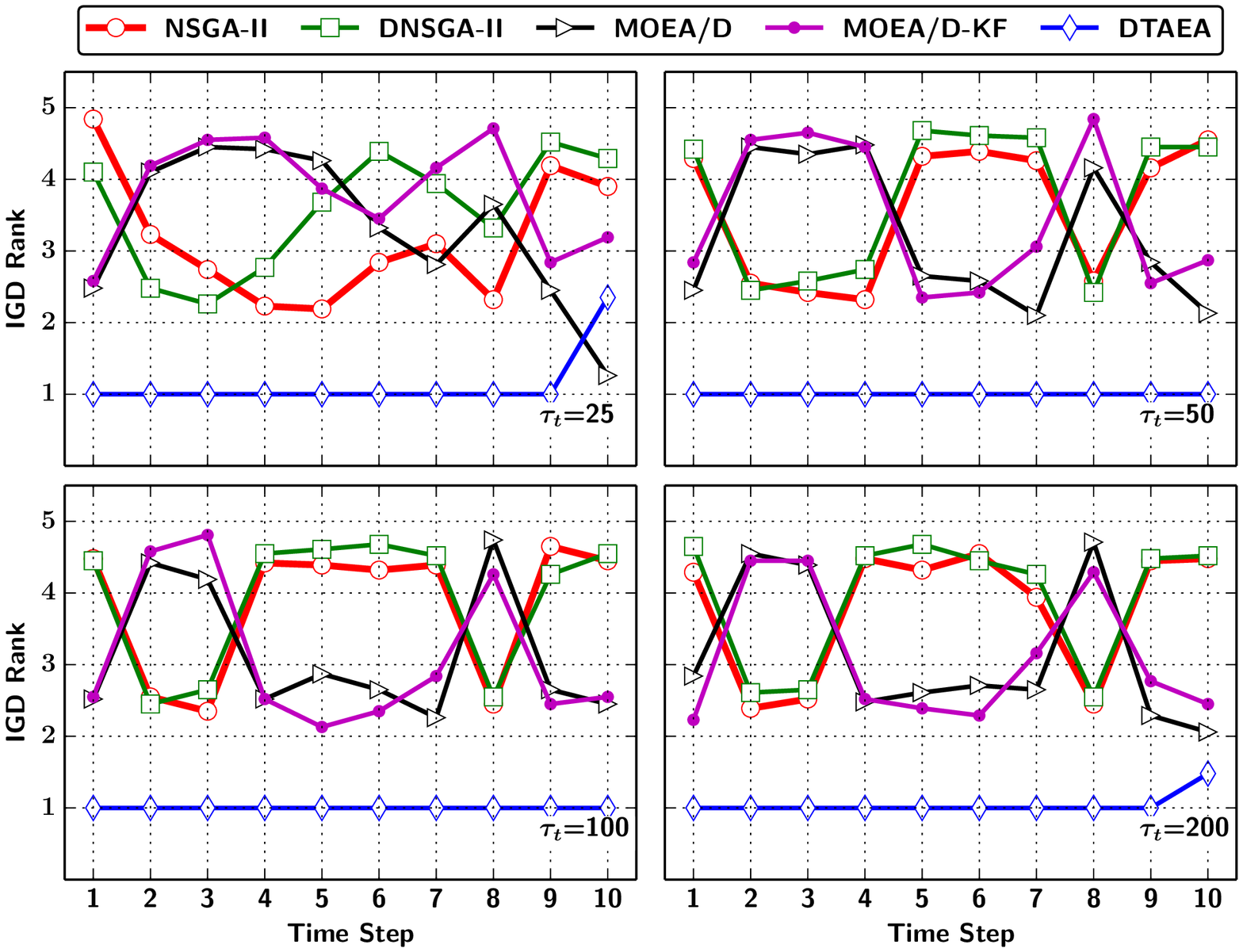}}\\
    \subfloat[F3]{\includegraphics[width=.5\linewidth]{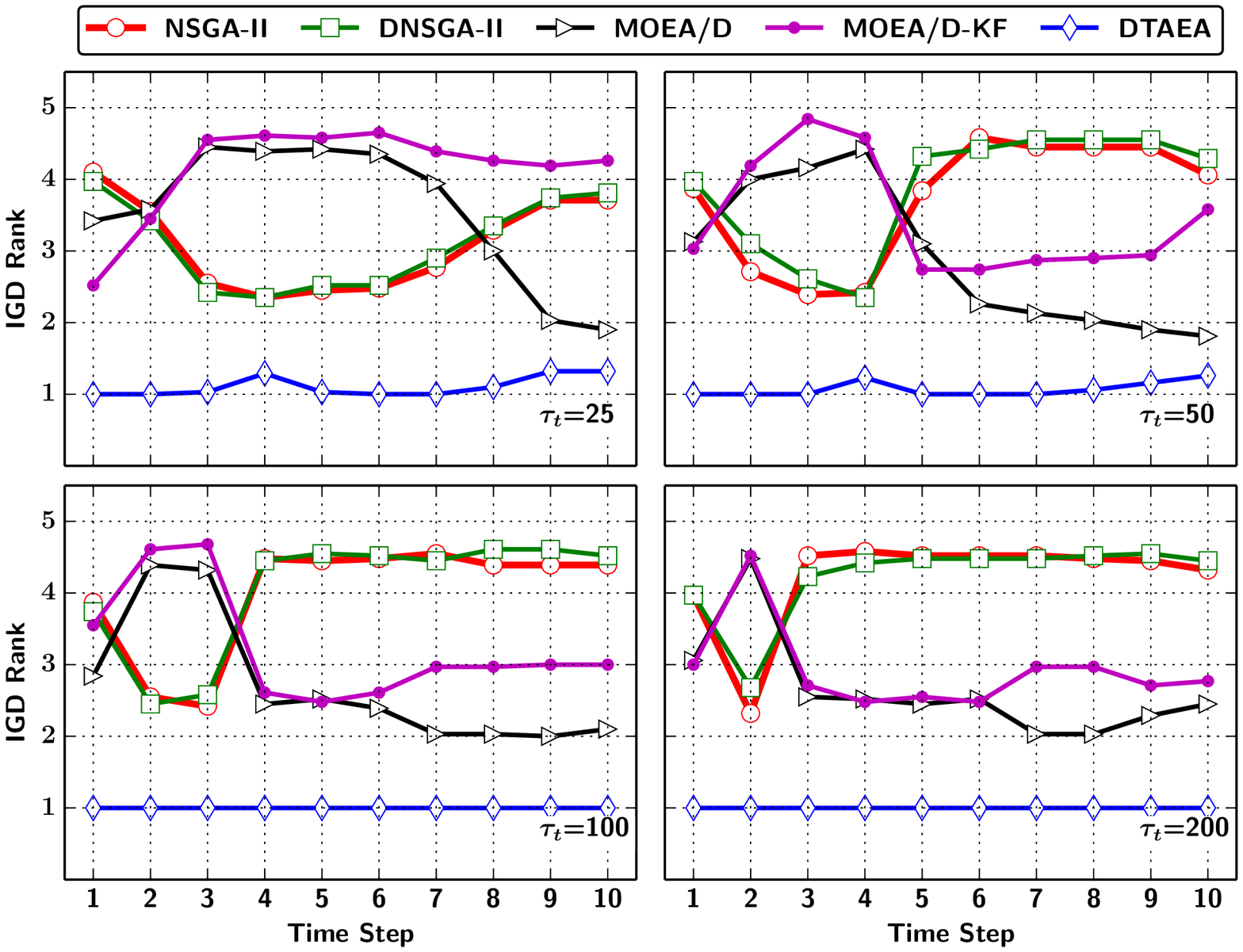}}
    \subfloat[F4]{\includegraphics[width=.5\linewidth]{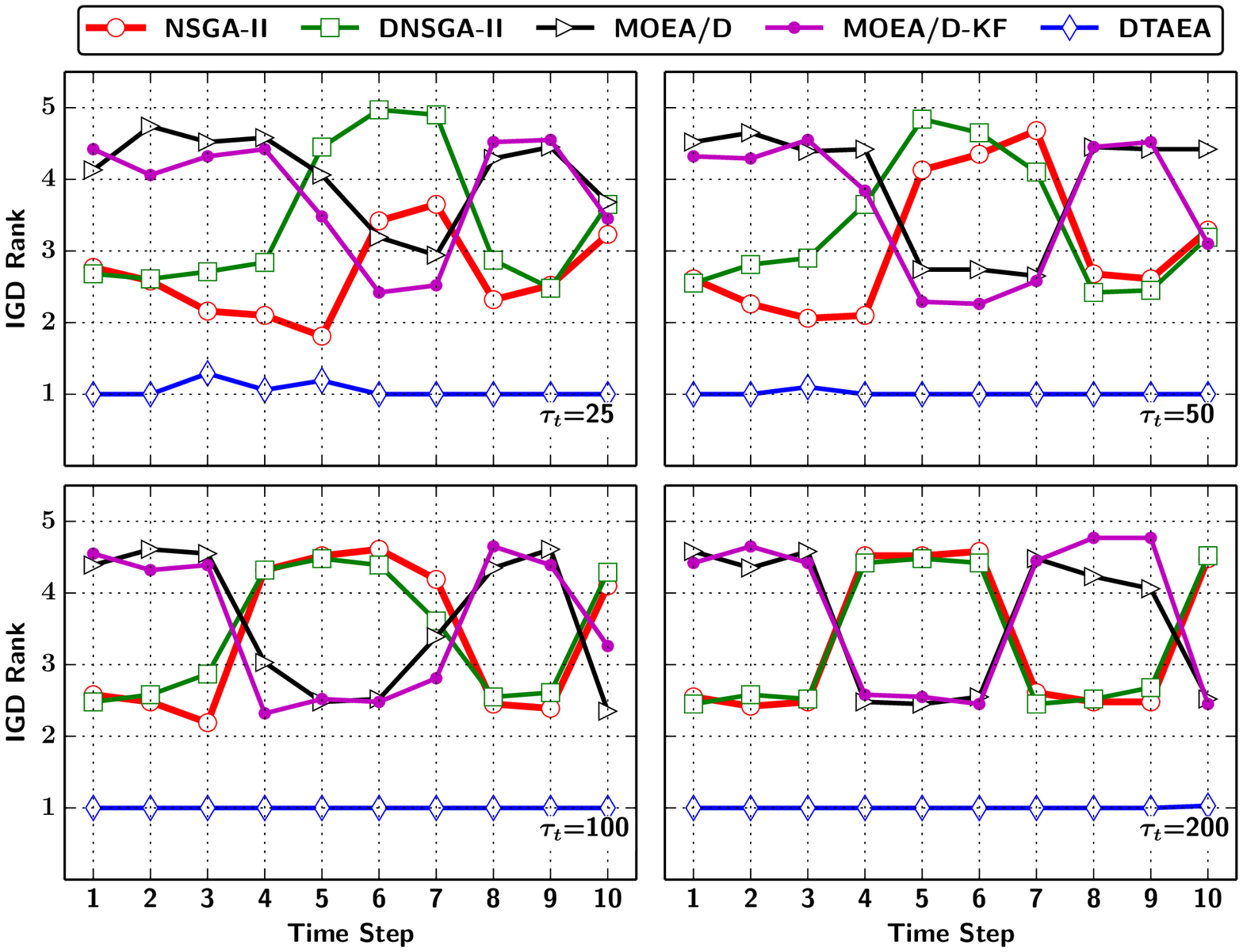}}\\
    \subfloat[F5]{\includegraphics[width=.5\linewidth]{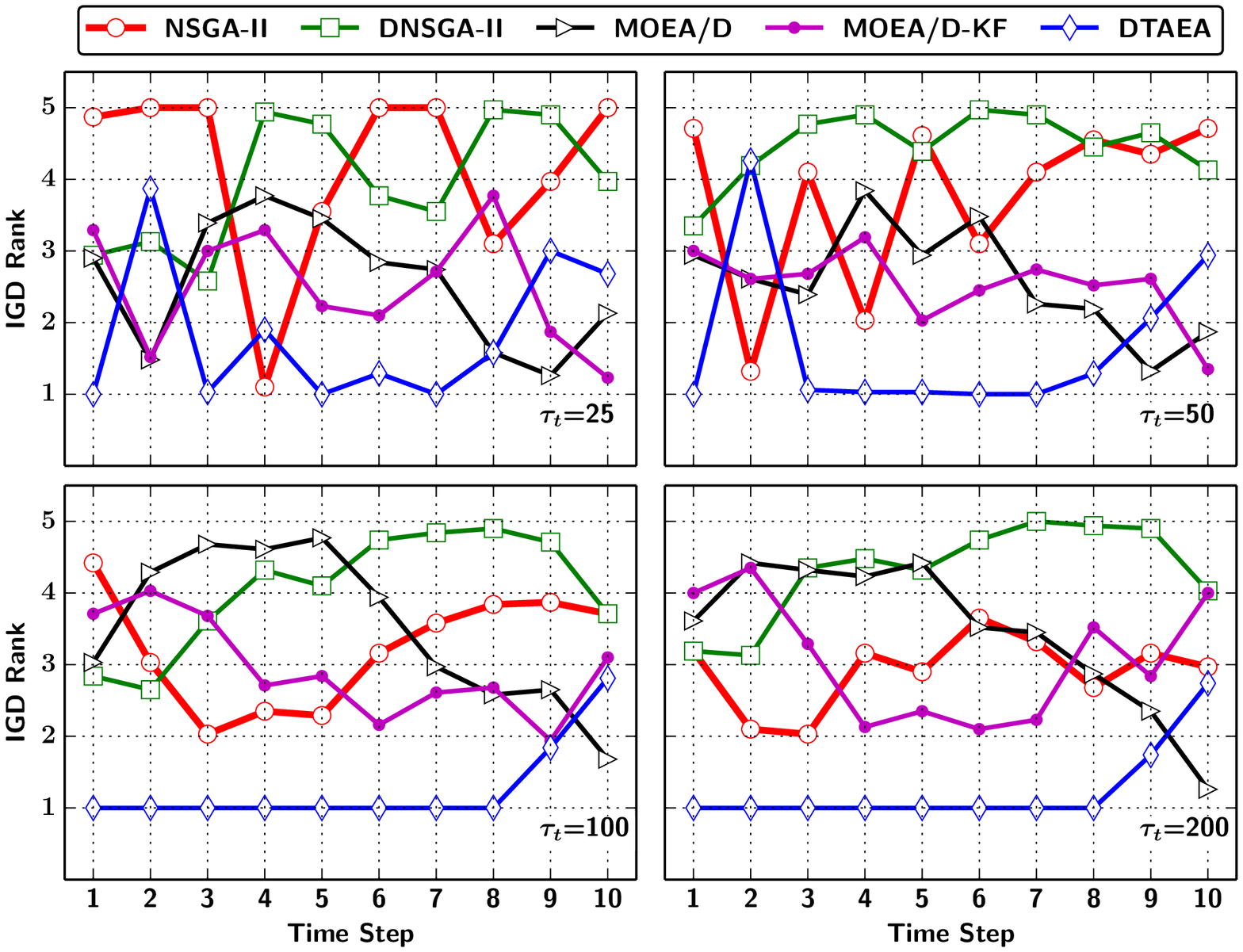}}
    \subfloat[F6]{\includegraphics[width=.5\linewidth]{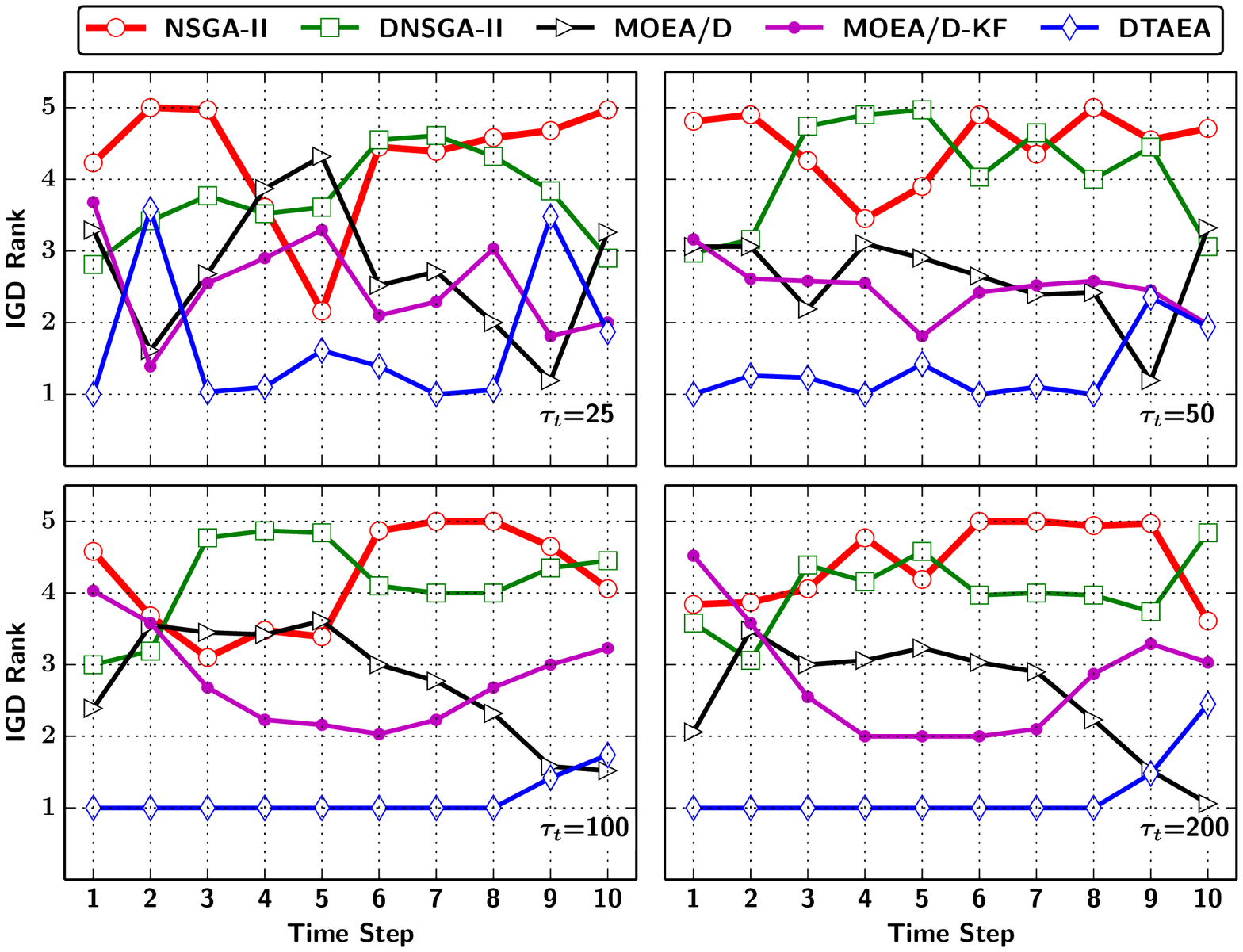}}
\caption{The rank of IGD obtained by different algorithms at each time step.}
\label{fig:rank}
\end{figure*}

F5 is developed from F2 and is with a time varying PS of which the position is changed with time. Although the overall performance of DTAEA is the best as shown in~\pref{tab:MIGD} and~\pref{tab:MHV}, its performance fluctuates significantly under a high frequency of change, i.e., $\tau_t=25$ and $\tau_t=50$, as shown in~\pref{fig:trajectory}(e) and~\pref{fig:rank}(e). It is worth noting that DTAEA does not have any dynamic detection or handling technique for the time varying PS. Thus, under a high frequency of change, DTAEA does not make a good adaption to two kinds of dynamics. However, as discussed in~\pref{sec:empirical-F14}, F2 does not pose too much difficulty to the algorithm for converging to the PF. This explains the comparable performance achieved by the other four algorithms under a high frequency of change. Nevertheless, even without any specific dynamic handling technique for the time varying PS, the complementary effects of two archives of DTAEA help the population adapt to the changing environments when decreasing the frequency of change, i.e., increasing $\tau_t$. Furthermore, NSGA-II and MOEA/D still show comparable performance than their dynamic counterparts when encountering two kinds of dynamics.

F6 is also developed from F2 but it poses more challenges to the algorithm for approaching the PF. Similar to the observations on F5, the performance of DTAEA fluctuate significantly under a high frequency of change as shown in~\pref{fig:trajectory}(f) and~\pref{fig:rank}(f). However, since the $g$ function of F6, which controls the difficulty for converging to the PF, is more difficult than that of F5, the superiority of DTAEA is more observable than F5. In addition, although the dynamic handling technique of DNSGA-II and MOEA/D-KF are designed for handling the time varying PS, they do not show better performance than their stationary counterparts. We explain these observations as that the severity of change of the time varying PS is smaller than the expansion or contraction of the PF or PS manifold when changing the number of objectives. Thus, due to the ineffectiveness of the dynamic reaction mechanisms of DNSGA-II and MOEA/D-KF for handling the changing number of objectives, it makes their performance be not much indifferent from their dynamic counterparts.
%It is interesting to note that the stationary algorithms, i.e., NSGA-II and MOEA/D, have shown superior performance to their dynamic counterparts, i.e., DNSGA-II and MOEA/D-KF, under a high frequency of change, i.e., $\tau_t=25$. Furthermore, from~\pref{fig:IGD-F1} and~\pref{fig:IGD-rank-F1}, we also notice that the stationary MOEA/D shows better performance than MOEA/D-KF after the 6-th time step, i.e., when decreasing the number of objectives. In addition, the performance of NSGA-II and DNSGA-II are relatively better than that of MOEA/D and MOEA/D-KF at the first several time steps, i.e., when increasing the number of objectives; whereas their performance become poor afterwards. As reported in many studies~\cite{KhareYD03,Hughes05}, NSGA-II cannot work well for problems for problems with more than 4 objectives. However, since NSGA-II has a decent convergence for the problems with a small number of objectives, its performance is competitive during the first several time steps.

%All these observations imply that the dynamic reaction operations of DNSGA-II and MOEA/D-KF might not be capable for handling dynamic problems with a changing number of objectives. Even worse, these dynamic reaction operations can make adverse effects to the search process when the number of objectives changes.

\begin{table*}[htbp]
\tiny
\centering
\caption{Performance Comparisons of DTAEA and the Other Algorithms on MIGD Metric}
\label{tab:MIGD}
\begin{tabular}{c|c|c|c|c|c|c|c|c|c|c|c}
\toprule
 \multirow{2}{*}{} & \multirow{3}{*}{$\tau_t$} &  \multicolumn{2}{c|}{NSGA-II} & \multicolumn{2}{c|}{DNSGA-II} & \multicolumn{2}{c|}{MOEA/D} & \multicolumn{2}{c|}{MOEA/D-KF} & \multicolumn{2}{c}{DTAEA} \\ \cmidrule{3-12}
 &  & MIGD & R & MIGD & R & MIGD & R & MIGD & R & MIGD & R \\ \midrule 
\multirow{4}{*}{F1}
 & 25 & ${\text{1.40E-2}}{(\text{9.25E-3})}$$^{\dag}$ & 3.3 & ${\text{1.32E-2}}{(\text{6.51E-3})}$$^{\dag}$ & 3.3 & ${\text{1.53E-3}}{(\text{2.18E-4})}$$^{\dag}$ & 3.1 & ${\text{2.05E-2}}{(\text{1.81E-2})}$$^{\dag}$ & 4.2 & \bb{${\text{5.50E-4}}{(\text{1.99E-4})}$} & \bb{\text{1.1}} \\
 & 50 & ${\text{3.36E-2}}{(\text{3.33E-2})}$$^{\dag}$ & 3.7 & ${\text{3.38E-2}}{(\text{2.07E-2})}$$^{\dag}$ & 3.8 & ${\text{8.53E-4}}{(\text{9.43E-4})}$$^{\dag}$ & 2.9 & ${\text{5.22E-3}}{(\text{6.23E-3})}$$^{\dag}$ & 3.5 & \bb{${\text{3.88E-4}}{(\text{1.03E-5})}$} & \bb{\text{1}} \\
 & 100 & ${\text{1.99E-2}}{(\text{5.68E-2})}$$^{\dag}$ & 3.9 & ${\text{2.84E-2}}{(\text{6.27E-2})}$$^{\dag}$ & 4.1 & ${\text{7.36E-4}}{(\text{4.29E-4})}$$^{\dag}$ & 2.7 & ${\text{9.78E-4}}{(\text{6.03E-4})}$$^{\dag}$ & 3.3 & \bb{${\text{3.64E-4}}{(\text{2.49E-6})}$} & \bb{\text{1}} \\
 & 200 & ${\text{1.71E-2}}{(\text{5.38E-2})}$$^{\dag}$ & 4.3 & ${\text{7.83E-3}}{(\text{3.45E-2})}$$^{\dag}$ & 4.1 & ${\text{6.47E-4}}{(\text{8.72E-4})}$$^{\dag}$ & 2.6 & ${\text{6.75E-4}}{(\text{4.58E-4})}$$^{\dag}$ & 3 & \bb{${\text{3.55E-4}}{(\text{1.41E-6})}$} & \bb{\text{1}} \\
\midrule

\multirow{4}{*}{F2}
 & 25 & ${\text{1.92E-3}}{(\text{1.27E-4})}$$^{\dag}$ & 3.2 & ${\text{2.02E-3}}{(\text{2.12E-4})}$$^{\dag}$ & 3.6 & ${\text{2.19E-3}}{(\text{1.17E-4})}$$^{\dag}$ & 3.3 & ${\text{2.26E-3}}{(\text{6.71E-5})}$$^{\dag}$ & 3.8 & \bb{${\text{1.26E-3}}{(\text{4.58E-6})}$} & \bb{\text{1.1}} \\
 & 50 & ${\text{2.20E-3}}{(\text{9.07E-5})}$$^{\dag}$ & 3.6 & ${\text{2.24E-3}}{(\text{1.47E-4})}$$^{\dag}$ & 3.7 & ${\text{2.08E-3}}{(\text{5.81E-5})}$$^{\dag}$ & 3.2 & ${\text{2.10E-3}}{(\text{6.03E-5})}$$^{\dag}$ & 3.5 & \bb{${\text{1.25E-3}}{(\text{3.26E-6})}$} & \bb{\text{1}} \\
 & 100 & ${\text{2.45E-3}}{(\text{1.37E-4})}$$^{\dag}$ & 3.8 & ${\text{2.47E-3}}{(\text{1.41E-4})}$$^{\dag}$ & 3.9 & ${\text{2.01E-3}}{(\text{2.89E-5})}$$^{\dag}$ & 3.1 & ${\text{1.99E-3}}{(\text{3.51E-5})}$$^{\dag}$ & 3.1 & \bb{${\text{1.22E-3}}{(\text{1.30E-6})}$} & \bb{\text{1}} \\
 & 200 & ${\text{2.46E-3}}{(\text{1.61E-4})}$$^{\dag}$ & 3.8 & ${\text{2.50E-3}}{(\text{1.48E-4})}$$^{\dag}$ & 3.9 & ${\text{1.93E-3}}{(\text{4.26E-5})}$$^{\dag}$ & 3.1 & ${\text{1.92E-3}}{(\text{3.41E-5})}$$^{\dag}$ & 3.1 & \bb{${\text{1.20E-3}}{(\text{2.20E-6})}$} & \bb{\text{1}} \\
\midrule

\multirow{4}{*}{F3}
 & 25 & ${\text{2.15E-2}}{(\text{1.78E-2})}$$^{\dag}$ & 3.1 & ${\text{2.62E-2}}{(\text{1.82E-2})}$$^{\dag}$ & 3.1 & ${\text{4.68E-3}}{(\text{1.25E-2})}$$^{\dag}$ & 3.5 & ${\text{4.90E-2}}{(\text{6.92E-2})}$$^{\dag}$ & 4.1 & \bb{${\text{2.06E-3}}{(\text{1.34E-3})}$} & \bb{\text{1.1}} \\
 & 50 & ${\text{9.09E-2}}{(\text{3.82E-2})}$$^{\dag}$ & 3.7 & ${\text{8.88E-2}}{(\text{4.13E-2})}$$^{\dag}$ & 3.9 & ${\text{2.96E-3}}{(\text{5.33E-4})}$$^{\dag}$ & 2.9 & ${\text{1.02E-2}}{(\text{5.72E-3})}$$^{\dag}$ & 3.4 & \bb{${\text{1.39E-3}}{(\text{1.06E-4})}$} & \bb{\text{1.1}} \\
 & 100 & ${\text{1.61E-1}}{(\text{1.08E-1})}$$^{\dag}$ & 4 & ${\text{1.35E-1}}{(\text{1.22E-1})}$$^{\dag}$ & 4 & ${\text{2.22E-3}}{(\text{1.31E-4})}$$^{\dag}$ & 2.7 & ${\text{2.57E-3}}{(\text{9.73E-4})}$$^{\dag}$ & 3.2 & \bb{${\text{1.24E-3}}{(\text{9.04E-6})}$} & \bb{\text{1}} \\
 & 200 & ${\text{1.82E-1}}{(\text{2.01E-1})}$$^{\dag}$ & 4.2 & ${\text{1.91E-1}}{(\text{1.47E-1})}$$^{\dag}$ & 4.2 & ${\text{2.09E-3}}{(\text{4.00E-3})}$$^{\dag}$ & 2.6 & ${\text{2.16E-3}}{(\text{9.26E-4})}$$^{\dag}$ & 2.9 & \bb{${\text{1.22E-3}}{(\text{4.17E-6})}$} & \bb{\text{1}} \\
\midrule

\multirow{4}{*}{F4}
 & 25 & ${\text{3.06E-3}}{(\text{5.08E-4})}$$^{\dag}$ & 2.7 & ${\text{3.66E-3}}{(\text{5.08E-4})}$$^{\dag}$ & 3.4 & ${\text{4.19E-3}}{(\text{3.56E-4})}$$^{\dag}$ & 4.1 & ${\text{4.02E-3}}{(\text{2.12E-4})}$$^{\dag}$ & 3.8 & \bb{${\text{1.32E-3}}{(\text{7.80E-5})}$} & \bb{\text{1.1}} \\
 & 50 & ${\text{2.93E-3}}{(\text{3.78E-4})}$$^{\dag}$ & 3.1 & ${\text{3.19E-3}}{(\text{3.71E-4})}$$^{\dag}$ & 3.4 & ${\text{3.74E-3}}{(\text{2.07E-4})}$$^{\dag}$ & 3.9 & ${\text{3.62E-3}}{(\text{3.94E-4})}$$^{\dag}$ & 3.6 & \bb{${\text{1.24E-3}}{(\text{4.05E-6})}$} & \bb{\text{1}} \\
 & 100 & ${\text{3.19E-3}}{(\text{3.12E-4})}$$^{\dag}$ & 3.4 & ${\text{3.16E-3}}{(\text{4.91E-4})}$$^{\dag}$ & 3.4 & ${\text{3.38E-3}}{(\text{6.86E-4})}$$^{\dag}$ & 3.6 & ${\text{3.15E-3}}{(\text{3.86E-4})}$$^{\dag}$ & 3.6 & \bb{${\text{1.22E-3}}{(\text{1.50E-6})}$} & \bb{\text{1}} \\
 & 200 & ${\text{2.79E-3}}{(\text{1.82E-4})}$$^{\dag}$ & 3.3 & ${\text{2.79E-3}}{(\text{3.63E-4})}$$^{\dag}$ & 3.3 & ${\text{2.73E-3}}{(\text{4.51E-4})}$$^{\dag}$ & 3.6 & ${\text{2.78E-3}}{(\text{5.72E-4})}$$^{\dag}$ & 3.8 & \bb{${\text{1.20E-3}}{(\text{1.35E-6})}$} & \bb{\text{1}} \\
\midrule

\multirow{4}{*}{F5}
 & 25 & ${\text{7.93E-3}}{(\text{3.37E-3})}$$^{\dag}$ & 4.2 & ${\text{3.69E-3}}{(\text{2.93E-4})}$$^{\dag}$ & 4 & ${\text{2.61E-3}}{(\text{2.62E-4})}$ & 2.6 & \bb{${\text{2.56E-3}}{(\text{1.78E-4})}$$^{\ddag}$} & 2.5 & ${\text{2.61E-3}}{(\text{9.71E-5})}$ & \bb{\text{1.8}} \\
 & 50 & ${\text{3.36E-3}}{(\text{2.45E-4})}$$^{\dag}$ & 3.8 & ${\text{3.69E-3}}{(\text{2.36E-4})}$$^{\dag}$ & 4.5 & ${\text{2.31E-3}}{(\text{1.25E-4})}$$^{\dag}$ & 2.6 & ${\text{2.24E-3}}{(\text{1.08E-4})}$$^{\dag}$ & 2.5 & \bb{${\text{1.91E-3}}{(\text{6.69E-5})}$} & \bb{\text{1.7}} \\
 & 100 & ${\text{2.02E-3}}{(\text{1.80E-4})}$$^{\dag}$ & 3.2 & ${\text{2.24E-3}}{(\text{1.23E-4})}$$^{\dag}$ & 4 & ${\text{2.13E-3}}{(\text{9.99E-5})}$$^{\dag}$ & 3.5 & ${\text{1.99E-3}}{(\text{4.51E-5})}$$^{\dag}$ & 2.9 & \bb{${\text{1.45E-3}}{(\text{2.98E-5})}$} & \bb{\text{1.3}} \\
 & 200 & ${\text{1.91E-3}}{(\text{1.06E-4})}$$^{\dag}$ & 2.9 & ${\text{2.20E-3}}{(\text{1.23E-4})}$$^{\dag}$ & 4.3 & ${\text{2.02E-3}}{(\text{7.70E-5})}$$^{\dag}$ & 3.4 & ${\text{1.93E-3}}{(\text{4.98E-5})}$$^{\dag}$ & 3.1 & \bb{${\text{1.36E-3}}{(\text{6.77E-6})}$} & \bb{\text{1.2}} \\
\midrule

\multirow{4}{*}{F6}
 & 25 & ${\text{8.10E-3}}{(\text{1.09E-3})}$$^{\dag}$ & 4.3 & ${\text{5.19E-3}}{(\text{7.90E-4})}$$^{\dag}$ & 3.7 & ${\text{3.12E-3}}{(\text{3.53E-4})}$$^{\dag}$ & 2.7 & \bb{${\text{2.93E-3}}{(\text{1.86E-4})}$} & 2.5 & ${\text{2.98E-3}}{(\text{1.76E-4})}$ & \bb{\text{1.7}} \\
 & 50 & ${\text{5.27E-3}}{(\text{5.91E-4})}$$^{\dag}$ & 4.5 & ${\text{5.60E-3}}{(\text{4.67E-4})}$$^{\dag}$ & 4.1 & ${\text{2.68E-3}}{(\text{7.39E-5})}$$^{\dag}$ & 2.6 & ${\text{2.71E-3}}{(\text{1.72E-4})}$$^{\dag}$ & 2.5 & \bb{${\text{2.08E-3}}{(\text{5.08E-5})}$} & \bb{\text{1.3}} \\
 & 100 & ${\text{3.56E-3}}{(\text{3.48E-4})}$$^{\dag}$ & 4.2 & ${\text{3.29E-3}}{(\text{1.11E-4})}$$^{\dag}$ & 4.2 & ${\text{2.36E-3}}{(\text{1.95E-4})}$$^{\dag}$ & 2.8 & ${\text{2.32E-3}}{(\text{1.08E-4})}$$^{\dag}$ & 2.8 & \bb{${\text{1.56E-3}}{(\text{2.25E-5})}$} & \bb{\text{1.1}} \\
 & 200 & ${\text{3.90E-3}}{(\text{3.72E-4})}$$^{\dag}$ & 4.4 & ${\text{3.10E-3}}{(\text{1.46E-4})}$$^{\dag}$ & 4 & ${\text{2.24E-3}}{(\text{9.08E-5})}$$^{\dag}$ & 2.6 & ${\text{2.16E-3}}{(\text{1.06E-4})}$$^{\dag}$ & 2.8 & \bb{${\text{1.46E-3}}{(\text{2.09E-5})}$} & \bb{\text{1.2}} \\
\bottomrule
\end{tabular}

\begin{tablenotes}
\item[1] R denotes the global rank assigned to each algorithm by averaging the ranks obtained at all time steps. Wilcoxon's rank sum test at a 0.05 significance level is performed between DTAEA and each of NSGA-II, DNSGA-II, MOEA/D and MOEA/D-KF. $^{\dag}$ and $^{\ddag}$ denote the performance of the corresponding algorithm is significantly worse than and better than that of DTAEA, respectively. The best median value is highlighted in boldface with gray background.
\end{tablenotes}

\end{table*}

\begin{table*}[htbp]
\tiny
\centering
\caption{Performance Comparisons of DTAEA and the Other Algorithms on MHV Metric}
\label{tab:MHV}
\begin{tabular}{c|c|c|c|c|c|c|c|c|c|c|c}
\toprule
 \multirow{2}{*}{} & \multirow{3}{*}{$\tau_t$} &  \multicolumn{2}{c|}{NSGA-II} & \multicolumn{2}{c|}{DNSGA-II} & \multicolumn{2}{c|}{MOEA/D} & \multicolumn{2}{c|}{MOEA/D-KF} & \multicolumn{2}{c}{DTAEA} \\ \cmidrule{3-12}
 &  & MHV & R & MHV & R & MHV & R & MHV & R & MHV & R \\ \midrule
\multirow{4}{*}{F1}
 & 25 & ${\text{74.8\%}}{(\text{9.58E-2})}$$^{\dag}$ & 3.6 & ${\text{72.3\%}}{(\text{8.08E-2})}$$^{\dag}$ & 3.7 & ${\text{97.3\%}}{(\text{1.09E-2})}$$^{\dag}$ & 2.7 & ${\text{74.6\%}}{(\text{4.33E-2})}$$^{\dag}$ & 3.9 & \bb{${\text{99.7\%}}{(\text{6.03E-3})}$} & \bb{\text{1.1}} \\
 & 50 & ${\text{60.1\%}}{(\text{1.16E-1})}$$^{\dag}$ & 4.1 & ${\text{61.2\%}}{(\text{1.02E-1})}$$^{\dag}$ & 4.1 & ${\text{99.7\%}}{(\text{3.40E-2})}$$^{\dag}$ & 2.6 & ${\text{78.0\%}}{(\text{1.22E-1})}$$^{\dag}$ & 3.2 & \bb{${\text{100.0\%}}{(\text{8.39E-5})}$} & \bb{\text{1}} \\
 & 100 & ${\text{74.9\%}}{(\text{2.10E-1})}$$^{\dag}$ & 4 & ${\text{69.3\%}}{(\text{2.43E-1})}$$^{\dag}$ & 4.1 & ${\text{99.9\%}}{(\text{1.65E-2})}$$^{\dag}$ & 2.7 & ${\text{99.0\%}}{(\text{2.07E-2})}$$^{\dag}$ & 3.2 & \bb{${\text{100.0\%}}{(\text{1.84E-5})}$} & \bb{\text{1}} \\
 & 200 & ${\text{76.0\%}}{(\text{3.43E-1})}$$^{\dag}$ & 4.1 & ${\text{83.3\%}}{(\text{3.58E-1})}$$^{\dag}$ & 3.9 & ${\text{100.0\%}}{(\text{4.65E-2})}$$^{\dag}$ & 2.8 & ${\text{100.0\%}}{(\text{2.15E-2})}$$^{\dag}$ & 3.2 & \bb{${\text{100.0\%}}{(\text{2.69E-5})}$} & \bb{\text{1}} \\
\midrule

\multirow{4}{*}{F2}
 & 25 & ${\text{92.4\%}}{(\text{3.47E-3})}$$^{\dag}$ & 4.1 & ${\text{92.4\%}}{(\text{6.01E-3})}$$^{\dag}$ & 4 & ${\text{93.5\%}}{(\text{2.91E-3})}$$^{\dag}$ & 2.6 & ${\text{93.4\%}}{(\text{1.36E-3})}$$^{\dag}$ & 3.1 & \bb{${\text{94.4\%}}{(\text{6.26E-5})}$} & \bb{\text{1}} \\
 & 50 & ${\text{91.9\%}}{(\text{4.14E-3})}$$^{\dag}$ & 4.2 & ${\text{91.7\%}}{(\text{5.28E-3})}$$^{\dag}$ & 4.1 & ${\text{93.8\%}}{(\text{1.06E-3})}$$^{\dag}$ & 2.6 & ${\text{93.7\%}}{(\text{1.36E-3})}$$^{\dag}$ & 3.1 & \bb{${\text{94.4\%}}{(\text{3.09E-5})}$} & \bb{\text{1}} \\
 & 100 & ${\text{91.0\%}}{(\text{5.33E-3})}$$^{\dag}$ & 4.3 & ${\text{90.6\%}}{(\text{8.46E-3})}$$^{\dag}$ & 4.4 & ${\text{93.9\%}}{(\text{4.55E-4})}$$^{\dag}$ & 2.5 & ${\text{93.9\%}}{(\text{6.22E-4})}$$^{\dag}$ & 2.8 & \bb{${\text{94.5\%}}{(\text{8.88E-6})}$} & \bb{\text{1}} \\
 & 200 & ${\text{90.8\%}}{(\text{6.80E-3})}$$^{\dag}$ & 4.3 & ${\text{90.4\%}}{(\text{7.10E-3})}$$^{\dag}$ & 4.4 & ${\text{94.0\%}}{(\text{5.10E-4})}$$^{\dag}$ & 2.6 & ${\text{94.0\%}}{(\text{5.09E-4})}$$^{\dag}$ & 2.7 & \bb{${\text{94.5\%}}{(\text{4.03E-6})}$} & \bb{\text{1}} \\
\midrule

\multirow{4}{*}{F3}
 & 25 & ${\text{67.1\%}}{(\text{7.75E-2})}$$^{\dag}$ & 3.4 & ${\text{69.3\%}}{(\text{7.98E-2})}$$^{\dag}$ & 3.3 & ${\text{80.6\%}}{(\text{6.04E-1})}$$^{\dag}$ & 3.4 & ${\text{59.9\%}}{(\text{5.09E-2})}$$^{\dag}$ & 3.8 & \bb{${\text{90.6\%}}{(\text{9.47E-2})}$} & \bb{\text{1.1}} \\
 & 50 & ${\text{49.9\%}}{(\text{2.21E-2})}$$^{\dag}$ & 4 & ${\text{49.3\%}}{(\text{3.46E-2})}$$^{\dag}$ & 4.1 & ${\text{89.8\%}}{(\text{2.96E-2})}$$^{\dag}$ & 2.5 & ${\text{67.2\%}}{(\text{5.19E-2})}$$^{\dag}$ & 3.3 & \bb{${\text{94.0\%}}{(\text{3.66E-3})}$} & \bb{\text{1.1}} \\
 & 100 & ${\text{45.9\%}}{(\text{6.49E-2})}$$^{\dag}$ & 4.3 & ${\text{44.8\%}}{(\text{7.63E-2})}$$^{\dag}$ & 4.3 & ${\text{93.2\%}}{(\text{8.02E-3})}$$^{\dag}$ & 2.4 & ${\text{91.3\%}}{(\text{6.33E-2})}$$^{\dag}$ & 3 & \bb{${\text{94.4\%}}{(\text{3.32E-4})}$} & \bb{\text{1}} \\
 & 200 & ${\text{44.0\%}}{(\text{7.13E-2})}$$^{\dag}$ & 4.2 & ${\text{44.6\%}}{(\text{7.75E-2})}$$^{\dag}$ & 4.2 & ${\text{93.6\%}}{(\text{7.25E-2})}$$^{\dag}$ & 2.7 & ${\text{93.3\%}}{(\text{5.40E-2})}$$^{\dag}$ & 3 & \bb{${\text{94.4\%}}{(\text{1.26E-4})}$} & \bb{\text{1}} \\
\midrule

\multirow{4}{*}{F4}
 & 25 & ${\text{91.2\%}}{(\text{4.96E-3})}$$^{\dag}$ & 3.3 & ${\text{88.8\%}}{(\text{2.09E-2})}$$^{\dag}$ & 3.9 & ${\text{90.8\%}}{(\text{1.67E-2})}$$^{\dag}$ & 3.4 & ${\text{91.7\%}}{(\text{8.38E-3})}$$^{\dag}$ & 3.3 & \bb{${\text{94.4\%}}{(\text{1.60E-4})}$} & \bb{\text{1}} \\
 & 50 & ${\text{88.8\%}}{(\text{2.04E-2})}$$^{\dag}$ & 3.6 & ${\text{87.6\%}}{(\text{2.81E-2})}$$^{\dag}$ & 3.7 & ${\text{91.9\%}}{(\text{9.97E-3})}$$^{\dag}$ & 3.4 & ${\text{92.1\%}}{(\text{1.41E-2})}$$^{\dag}$ & 3.3 & \bb{${\text{94.4\%}}{(\text{1.16E-5})}$} & \bb{\text{1}} \\
 & 100 & ${\text{86.6\%}}{(\text{2.83E-2})}$$^{\dag}$ & 3.6 & ${\text{86.7\%}}{(\text{3.45E-2})}$$^{\dag}$ & 3.5 & ${\text{92.4\%}}{(\text{1.31E-2})}$$^{\dag}$ & 3.4 & ${\text{92.6\%}}{(\text{8.25E-3})}$$^{\dag}$ & 3.5 & \bb{${\text{94.5\%}}{(\text{3.26E-6})}$} & \bb{\text{1}} \\
 & 200 & ${\text{88.8\%}}{(\text{1.32E-2})}$$^{\dag}$ & 4 & ${\text{88.5\%}}{(\text{2.65E-2})}$$^{\dag}$ & 3.9 & ${\text{92.9\%}}{(\text{1.13E-2})}$$^{\dag}$ & 3 & ${\text{92.9\%}}{(\text{1.26E-2})}$$^{\dag}$ & 3.2 & \bb{${\text{94.5\%}}{(\text{4.19E-6})}$} & \bb{\text{1}} \\
\midrule

\multirow{4}{*}{F5}
 & 25 & ${\text{58.6\%}}{(\text{1.38E-1})}$$^{\dag}$ & 4.5 & ${\text{87.0\%}}{(\text{2.36E-2})}$$^{\dag}$ & 4.1 & ${\text{93.3\%}}{(\text{6.10E-3})}$$^{\dag}$ & 2.2 & \bb{${\text{93.3\%}}{(\text{2.82E-3})}$$^{\ddag}$} & 2.3 & ${\text{89.4\%}}{(\text{6.64E-3})}$ & \bb{\text{2}} \\
 & 50 & ${\text{86.3\%}}{(\text{1.60E-2})}$$^{\dag}$ & 4.3 & ${\text{85.9\%}}{(\text{1.60E-2})}$$^{\dag}$ & 4.3 & ${\text{94.0\%}}{(\text{2.02E-3})}$$^{\dag}$ & 2.1 & \bb{${\text{94.1\%}}{(\text{3.35E-3})}$$^{\ddag}$} & \bb{\text{1.9}} & ${\text{93.7\%}}{(\text{1.25E-3})}$ & 2.4 \\
 & 100 & ${\text{93.4\%}}{(\text{5.24E-3})}$$^{\dag}$ & 3.8 & ${\text{93.5\%}}{(\text{4.20E-3})}$$^{\dag}$ & 4.3 & ${\text{94.3\%}}{(\text{1.15E-3})}$$^{\dag}$ & 2.9 & ${\text{94.4\%}}{(\text{1.95E-3})}$$^{\dag}$ & 2.3 & \bb{${\text{94.6\%}}{(\text{3.62E-4})}$} & \bb{\text{1.7}} \\
 & 200 & ${\text{94.0\%}}{(\text{3.40E-3})}$$^{\dag}$ & 3.7 & ${\text{93.3\%}}{(\text{4.36E-3})}$$^{\dag}$ & 4.5 & ${\text{94.5\%}}{(\text{6.96E-4})}$$^{\dag}$ & 2.9 & ${\text{94.5\%}}{(\text{2.20E-3})}$$^{\dag}$ & 2.3 & \bb{${\text{94.8\%}}{(\text{1.84E-4})}$} & \bb{\text{1.6}} \\
\midrule

\multirow{4}{*}{F6}
 & 25 & ${\text{60.1\%}}{(\text{6.41E-2})}$$^{\dag}$ & 4.7 & ${\text{75.8\%}}{(\text{6.36E-2})}$$^{\dag}$ & 3.8 & ${\text{93.2\%}}{(\text{1.72E-2})}$$^{\dag}$ & 2.2 & \bb{${\text{93.5\%}}{(\text{2.77E-3})}$$^{\ddag}$} & \bb{\text{2.1}} & ${\text{89.1\%}}{(\text{9.72E-3})}$ & 2.3 \\
 & 50 & ${\text{78.9\%}}{(\text{3.83E-2})}$$^{\dag}$ & 4.6 & ${\text{73.6\%}}{(\text{2.76E-2})}$$^{\dag}$ & 4.3 & ${\text{94.0\%}}{(\text{1.71E-3})}$$^{\dag}$ & 1.9 & \bb{${\text{93.9\%}}{(\text{1.73E-3})}$$^{\ddag}$} & \bb{1.8} & ${\text{93.7\%}}{(\text{1.45E-3})}$ & \text{2.4} \\
 & 100 & ${\text{88.0\%}}{(\text{1.36E-2})}$$^{\dag}$ & 4.8 & ${\text{91.8\%}}{(\text{4.13E-3})}$$^{\dag}$ & 4 & ${\text{94.3\%}}{(\text{1.54E-3})}$$^{\dag}$ & 2.3 & ${\text{94.2\%}}{(\text{2.15E-3})}$$^{\dag}$ & 2.4 & \bb{${\text{94.5\%}}{(\text{3.50E-4})}$} & \bb{\text{1.5}} \\
 & 200 & ${\text{87.9\%}}{(\text{1.91E-2})}$$^{\dag}$ & 4.7 & ${\text{92.1\%}}{(\text{5.84E-3})}$$^{\dag}$ & 4.1 & ${\text{94.5\%}}{(\text{5.71E-4})}$$^{\dag}$ & 2.5 & ${\text{94.5\%}}{(\text{1.35E-3})}$$^{\dag}$ & 2.3 & \bb{${\text{94.7\%}}{(\text{3.39E-4})}$} & \bb{\text{1.4}} \\
\bottomrule
\end{tabular}
\begin{tablenotes}
\item[1] R denotes the global rank assigned to each algorithm by averaging the ranks obtained at all time steps. Wilcoxon's rank sum test at a 0.05 significance level is performed between DTAEA and each of NSGA-II, DNSGA-II, MOEA/D and MOEA/D-KF. $^{\dag}$ and $^{\ddag}$ denote the performance of the corresponding algorithm is significantly worse than and better than that of DTAEA, respectively. The best median value is highlighted in boldface with gray background.
\end{tablenotes}
\end{table*}

\subsection{Further Analysis}
\label{sec:analysis}

From the experimental studies in~\pref{sec:empirical-F14} and~\pref{sec:empirical-F56}, it is clear to see the superiority of our proposed DTAEA for handling the DMOPs with a dynamically changing number of objectives. As introduced in~\pref{sec:proposal}, DTAEA consists of three major components. The following paragraphs analyze their effects separately.

\subsubsection{Effects of the Reconstruction Mechanism}
\label{sec:effect-reconstruction}

As discussed in~\pref{sec:challenges}, the population convergence might not be affected when increasing the number of objectives. However, the diversity enhancement strategy, e.g., the random injection of DNSGA-II, has an adverse effect to the population convergence. In the meanwhile, it cannot provide enough diversified information to remedy the loss of population diversity. As for the prediction strategy, e.g., the Kalman filter of MOEA/D-KF, it is usually developed to predict the change of the PS's position or shape, instead of the expansion or contraction of the PS or PF manifold. As a consequence, its large prediction errors are harmful to both the convergence and the diversity. Since the stationary EMO algorithms do not make any reaction to the changing environment, the population convergence is not affected; while the self-adaptive property of EA can also help the population gradually adapt to the changed environment. These explain the competitive performance of the stationary EMO algorithms comparing to their dynamic counterparts for handling the DMOPs with a dynamically changing number of objectives. In contrast, the CA of our proposed DTAEA, consisted of the elite solutions, preserves the population convergence; in the meanwhile, the DA of DTAEA, consisted of a set of uniformly sampled solutions from the decision space, provide as much diversified information as possible to help the population adapt to the changed environment. In Fig. 3 of the supplementary file\footnote{The supplementary document can be found from https://coda-group.github.io/publications/suppDTAEA.pdf}, we keep track of the population distribution for several iterations after the number of objectives increases from two to three. From this figure, we can clearly see that neither NSGA-II nor DNSGA-II is able to spread the population across the expanded PF. As for MOEA/D and MOEA/D-KF, although the population seems to have an acceptable spread along the expanded PF, the solutions are a bit away from the PF. In contrast, our proposed DTAEA successfully spread the solutions across the PF without sacrificing any convergence property.

As discussed in~\pref{sec:challenges}, the spread of the population might not be affected too much when decreasing the number of objectives, while some solutions are propelled away from the PF. In this case, the randomly injected solutions of DNGSA-II cannot provide sufficient selection pressure towards the PF. As for MOEA/D-KF, due to the large prediction errors, the effect of those predicted solutions of the Kalman filter is almost the same as the random solutions. In contrast, DTAEA uses the non-dominated solutions of the elite population to construct the CA; in the meanwhile, it uses the remaining ones, which have a well spread in the objective space, to construct the DA. By these means, the population is able to have a well balance between convergence and diversity in the changed environment. In Fig. 4 of the supplementary file, we keep track of the population distribution for several iterations after the number of objectives decreases from four to three. From this figure, we can clearly see that the population spread after decreasing the number of objectives is acceptable in the first place, but having some duplicate solutions. After several iterations, DTAEA achieves a significantly better performance on both convergence and diversity than the other comparative algorithms.

%\begin{figure*}[htbp]
%\centering
%\includegraphics[width=.9\linewidth]{figs/change_figure_1.jpg}
%\caption{Variation of the population distribution when increasing the number of objectives from 2 to 3.}
%\label{fig:population-increase}
%\end{figure*}

%\begin{figure*}[htbp]
%\centering
%\includegraphics[width=.9\linewidth]{figs/change_figure_2.jpg}
%\caption{Variation of the population distribution when decreasing the number of objectives from 4 to 3.}
%\label{fig:population-decrease}
%\end{figure*}

\subsubsection{Effects of the Restricted Mating Selection Mechanism}
\label{sec:effect-mating}

\begin{figure}[htbp]
\centering
\includegraphics[width=.5\linewidth]{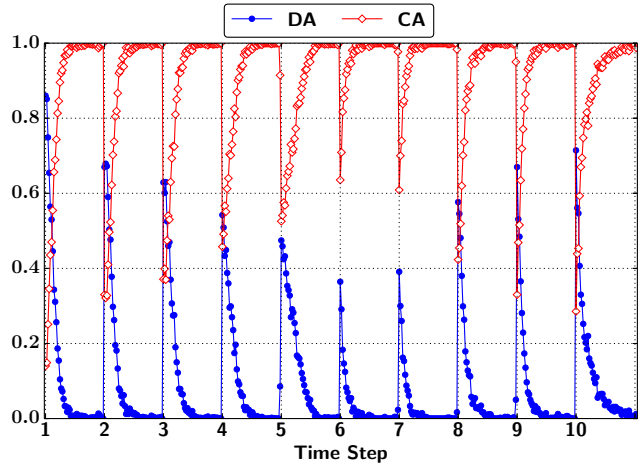}
\caption{Proportion of the second mating parent selected from the CA and DA respectively.}
\label{fig:matingrate}
\end{figure}

As discussed in~\pref{sec:reproduction}, the CA and the DA have different characteristics. In particular, the prior one concerns more about convergence whereas the latter one concerns more about the diversity. Their complementary effect, which finally contributes to the balance between convergence and diversity, is implemented by the restricted mating selection mechanism between them. In particular, one of the mating parents is constantly selected from the CA while the selection of the other one depends on the diversity information of the CA. In order to validate the effect of this restricted mating selection mechanism, we choose F2 as the instance where $\tau_t=50$ and plot the trajectories of proportion of second mating parent selected from the CA and DA respectively in~\pref{fig:matingrate}. From this figure, we notice that the proportion for selecting the second mating parent from the DA is relatively high at the beginning when the number of objectives is changed. But with the progress of evolution, both mating parents are selected from the CA when its diversity becomes better.

\subsubsection{Effects of the Update Mechanisms}
\label{sec:effect-update}

As discussed in~\pref{sec:update}, the update mechanisms are used to maintain the complementary effects between the CA and the DA. The CA keeps a continuously strong selection pressure for the population convergence; while the DA maintains a set of well diversified solutions. In order to validate the importance of the three different components of DTAEA, we developed three DTAEA variants as follows:
\begin{itemize}
    \item DTAEA-$v1$: this variant modifies DTAEA by removing the restricted mating selection mechanism proposed in~\pref{sec:reproduction}. In particular, now the mating parents are respectively selected from the CA and the DA without considering the population distribution.
    \item DTAEA-$v2$: this variant modifies DTAEA by removing the reconstruction mechanism proposed in~\pref{sec:reconstruction}. In other words, it does not make any response to the changing environment.
    \item DTAEA-$v3$: this variant merely uses the update mechanisms to maintain the CA and the DA whereas it does not make any response to the changing environment. In addition, it does not use the restricted mating selection mechanism as DTAEA-$v1$.
\end{itemize}

We conduct the experiments on F1 to F6 according to the same experimental settings introduced in~\pref{sec:settings}. \pref{tab:MIGDvariant} and \pref{tab:MHVvariant} give the median and IQR values according to the MIGD and MHV metrics. From these results we can see that the original DTAEA, consisted of all three components, has shown clearly better performance than the other three variants. More specifically, as shown in~\pref{tab:MIGDvariant} and~\pref{tab:MHVvariant}, the performance of DTAEA-$v3$ is the worst among all three variants. This observation is reasonable as DTAEA-$v3$ neither responds to the changing environment nor takes advantages of the complementary effect of the CA and the DA for offspring generation. The performance of DTAEA-$v2$ is slightly better than DTAEA-$v3$. Therefore, we can see that even without any response to the changing environment, the restricted mating selection mechanism can also provide some guidance to the search process. As for DTAEA-$v1$, we can see that the performance can be significantly improved by using the reconstruction mechanism proposed in~\pref{sec:reconstruction} to respond to the changing environment. This superiority is more obvious when the frequency of change is high. By comparing the results between DTAEA-$v1$ and the original DTAEA, we can clearly see the importance of taking advantages of the complementary effect of the CA and the DA for offspring generation. All in all, we can conclude that all three components of DTAEA are of significant importance for handling the DMOP with a changing number of objectives.

\begin{table*}[htbp]
\tiny
\centering
\caption{Performance Comparisons of DTAEA and its Three Variants on MIGD Metric}
\label{tab:MIGDvariant}

\begin{tabular}{c|c|c|c|c|c|c|c|c|c}
\toprule
 \multirow{2}{*}{} & \multirow{3}{*}{$\tau_t$} &  \multicolumn{2}{c|}{DTAEA-$v1$} & \multicolumn{2}{c|}{DTAEA-$v2$} & \multicolumn{2}{c|}{DTAEA-$v3$} & \multicolumn{2}{c}{DTAEA} \\ \cmidrule{3-10}
 &  & MIGD & R & MIGD & R & MIGD & R & MIGD & R  \\ \midrule 

\multirow{4}{*}{F1}
 & 25 & \bb{${\text{5.05E-4}}{(\text{1.85E-4})}$} & \bb{\text{1.9}} & ${\text{7.14E-4}}{(\text{1.72E-4})}$$^{\dag}$ & 2.8 & ${\text{7.29E-4}}{(\text{2.40E-4})}$$^{\dag}$ & 2.9 & ${\text{5.50E-4}}{(\text{1.99E-4})}$ & 2.3 \\
 & 50 & ${\text{3.95E-4}}{(\text{2.49E-5})}$ & 2.5 & ${\text{4.02E-4}}{(\text{1.54E-5})}$$^{\dag}$ & 2.6 & ${\text{4.04E-4}}{(\text{1.19E-5})}$$^{\dag}$ & 2.8 & \bb{${\text{3.88E-4}}{(\text{1.03E-5})}$} & \bb{\text{2.1}} \\
 & 100 & ${\text{3.72E-4}}{(\text{3.97E-6})}$$^{\dag}$ & 3.5 & ${\text{3.64E-4}}{(\text{3.09E-6})}$ & 2 & ${\text{3.65E-4}}{(\text{2.43E-6})}$ & 2.7 & \bb{${\text{3.64E-4}}{(\text{2.49E-6})}$} & \bb{\text{1.8}} \\
 & 200 & ${\text{3.59E-4}}{(\text{1.07E-6})}$$^{\dag}$ & 3.4 & ${\text{3.56E-4}}{(\text{1.20E-6})}$ & 1.8 & ${\text{3.59E-4}}{(\text{2.46E-6})}$$^{\dag}$ & 3.4 & \bb{${\text{3.55E-4}}{(\text{1.41E-6})}$} & \bb{\text{1.5}} \\
\midrule

\multirow{4}{*}{F2}
 & 25 & ${\text{1.28E-3}}{(\text{5.20E-6})}$$^{\dag}$ & 2.6 & ${\text{1.39E-3}}{(\text{1.97E-5})}$$^{\dag}$ & 2.8 & ${\text{1.40E-3}}{(\text{4.08E-5})}$$^{\dag}$ & 3.1 & \bb{${\text{1.26E-3}}{(\text{4.58E-6})}$} & \bb{\text{1.5}} \\
 & 50 & ${\text{1.26E-3}}{(\text{1.59E-6})}$$^{\dag}$ & 3.4 & ${\text{1.25E-3}}{(\text{1.76E-5})}$ & 2 & ${\text{1.25E-3}}{(\text{2.77E-5})}$$^{\dag}$ & 2.8 & \bb{${\text{1.25E-3}}{(\text{3.26E-6})}$} & \bb{\text{1.9}} \\
 & 100 & ${\text{1.23E-3}}{(\text{1.88E-6})}$$^{\dag}$ & 3.4 & ${\text{1.22E-3}}{(\text{2.24E-6})}$$^{\dag}$ & 1.9 & ${\text{1.23E-3}}{(\text{2.32E-6})}$$^{\dag}$ & 3 & \bb{${\text{1.22E-3}}{(\text{1.30E-6})}$} & \bb{\text{1.7}} \\
 & 200 & ${\text{1.21E-3}}{(\text{1.86E-6})}$$^{\dag}$ & 3.3 & ${\text{1.20E-3}}{(\text{1.31E-6})}$$^{\dag}$ & 1.8 & ${\text{1.21E-3}}{(\text{1.73E-6})}$$^{\dag}$ & 3.2 & \bb{${\text{1.20E-3}}{(\text{2.20E-6})}$} & \bb{\text{1.7}} \\
\midrule

\multirow{4}{*}{F3}
 & 25 & \bb{${\text{1.92E-3}}{(\text{9.48E-4})}$} & \bb{\text{1.9}} & ${\text{2.63E-3}}{(\text{1.29E-3})}$$^{\dag}$ & 3 & ${\text{2.32E-3}}{(\text{8.94E-4})}$ & 2.9 & ${\text{2.06E-3}}{(\text{1.34E-3})}$ & 2.2 \\
 & 50 & \bb{${\text{1.32E-3}}{(\text{4.29E-5})}$$^{\ddag}$} & \bb{\text{2.1}} & ${\text{1.42E-3}}{(\text{5.20E-5})}$ & 2.8 & ${\text{1.44E-3}}{(\text{5.95E-5})}$ & 2.9 & ${\text{1.39E-3}}{(\text{1.06E-4})}$ & 2.3 \\
 & 100 & ${\text{1.26E-3}}{(\text{5.32E-6})}$$^{\dag}$ & 3.1 & ${\text{1.25E-3}}{(\text{1.38E-5})}$ & 2.1 & ${\text{1.25E-3}}{(\text{1.87E-5})}$$^{\dag}$ & 2.7 & \bb{${\text{1.24E-3}}{(\text{9.04E-6})}$} & \bb{\text{2}} \\
 & 200 & ${\text{1.23E-3}}{(\text{3.09E-6})}$$^{\dag}$ & 3 & ${\text{1.22E-3}}{(\text{2.43E-6})}$$^{\dag}$ & 2.1 & ${\text{1.23E-3}}{(\text{5.26E-6})}$$^{\dag}$ & 3.2 & \bb{${\text{1.22E-3}}{(\text{4.17E-6})}$} & \bb{\text{1.8}} \\
\midrule

\multirow{4}{*}{F4}
 & 25 & \bb{${\text{1.29E-3}}{(\text{3.40E-5})}$$^{\ddag}$} & \bb{\text{1.7}} & ${\text{6.82E-3}}{(\text{2.18E-7})}$$^{\dag}$ & 3.3 & ${\text{6.82E-3}}{(\text{1.87E-7})}$$^{\dag}$ & 3.3 & ${\text{1.32E-3}}{(\text{7.80E-5})}$ & 1.7 \\
 & 50 & ${\text{1.25E-3}}{(\text{2.17E-6})}$$^{\dag}$ & 2 & ${\text{6.81E-3}}{(\text{8.96E-8})}$$^{\dag}$ & 3.3 & ${\text{6.81E-3}}{(\text{7.24E-8})}$$^{\dag}$ & 3.3 & \bb{${\text{1.24E-3}}{(\text{4.05E-6})}$} & \bb{\text{1.4}} \\
 & 100 & ${\text{1.22E-3}}{(\text{1.58E-6})}$$^{\dag}$ & 2 & ${\text{6.81E-3}}{(\text{1.93E-7})}$$^{\dag}$ & 3.3 & ${\text{6.81E-3}}{(\text{1.05E-7})}$$^{\dag}$ & 3.4 & \bb{${\text{1.22E-3}}{(\text{1.50E-6})}$} & \bb{\text{1.4}} \\
 & 200 & ${\text{1.21E-3}}{(\text{1.85E-6})}$$^{\dag}$ & 2.1 & ${\text{6.81E-3}}{(\text{1.58E-4})}$$^{\dag}$ & 3.3 & ${\text{6.81E-3}}{(\text{3.71E-4})}$$^{\dag}$ & 3.4 & \bb{${\text{1.20E-3}}{(\text{1.35E-6})}$} & \bb{\text{1.2}} \\
\midrule

\multirow{4}{*}{F5}
 & 25 & \bb{${\text{2.52E-3}}{(\text{9.09E-5})}$$^{\ddag}$} & 1.8 & ${\text{7.69E-3}}{(\text{5.69E-4})}$$^{\dag}$ & 3.3 & ${\text{7.66E-3}}{(\text{3.22E-4})}$$^{\dag}$ & 3.2 & ${\text{2.61E-3}}{(\text{9.71E-5})}$ & \bb{\text{1.6}} \\
 & 50 & ${\text{2.02E-3}}{(\text{6.94E-5})}$$^{\dag}$ & 1.8 & ${\text{1.35E-2}}{(\text{3.26E-4})}$$^{\dag}$ & 3.5 & ${\text{1.34E-2}}{(\text{2.17E-4})}$$^{\dag}$ & 3.5 & \bb{${\text{1.91E-3}}{(\text{6.69E-5})}$} & \bb{\text{1.2}} \\
 & 100 & ${\text{1.49E-3}}{(\text{1.72E-5})}$$^{\dag}$ & 1.7 & ${\text{1.80E-2}}{(\text{4.43E-4})}$$^{\dag}$ & 3.5 & ${\text{1.79E-2}}{(\text{9.68E-4})}$$^{\dag}$ & 3.5 & \bb{${\text{1.45E-3}}{(\text{2.98E-5})}$} & \bb{\text{1.3}} \\
 & 200 & ${\text{1.38E-3}}{(\text{1.17E-5})}$$^{\dag}$ & 1.7 & ${\text{1.95E-2}}{(\text{5.39E-4})}$$^{\dag}$ & 3.4 & ${\text{1.97E-2}}{(\text{3.73E-4})}$$^{\dag}$ & 3.6 & \bb{${\text{1.36E-3}}{(\text{6.77E-6})}$} & \bb{\text{1.3}} \\
\midrule

\multirow{4}{*}{F6}
 & 25 & \bb{${\text{2.90E-3}}{(\text{1.68E-4})}$} & 1.6 & ${\text{1.11E-2}}{(\text{7.24E-4})}$$^{\dag}$ & 3.5 & ${\text{1.12E-2}}{(\text{5.90E-4})}$$^{\dag}$ & 3.5 & ${\text{2.98E-3}}{(\text{1.76E-4})}$ & \bb{\text{1.4}} \\
 & 50 & ${\text{2.22E-3}}{(\text{7.36E-5})}$$^{\dag}$ & 1.7 & ${\text{1.61E-2}}{(\text{6.16E-4})}$$^{\dag}$ & 3.5 & ${\text{1.59E-2}}{(\text{7.97E-4})}$$^{\dag}$ & 3.5 & \bb{${\text{2.08E-3}}{(\text{5.08E-5})}$} & \bb{\text{1.3}} \\
 & 100 & ${\text{1.57E-3}}{(\text{2.31E-5})}$ & 1.6 & ${\text{2.10E-2}}{(\text{9.39E-4})}$$^{\dag}$ & 3.6 & ${\text{2.10E-2}}{(\text{9.48E-4})}$$^{\dag}$ & 3.4 & \bb{${\text{1.56E-3}}{(\text{2.25E-5})}$} & \bb{\text{1.4}} \\
 & 200 & ${\text{1.46E-3}}{(\text{1.07E-5})}$ & 1.6 & ${\text{2.23E-2}}{(\text{7.09E-4})}$$^{\dag}$ & 3.6 & ${\text{2.19E-2}}{(\text{6.50E-4})}$$^{\dag}$ & 3.4 & \bb{${\text{1.46E-3}}{(\text{2.09E-5})}$} & \bb{\text{1.4}} \\
\bottomrule
\end{tabular}
\begin{tablenotes}
\item[1] R denotes the global rank assigned to each algorithm by averaging the ranks obtained at all time steps. Wilcoxon's rank sum test at a 0.05 significance level is performed between DTAEA and each of DTAEA-$v1$, DTAEA-$v2$ and DTAEA-$v3$. $^{\dag}$ and $^{\ddag}$ denote the performance of the corresponding algorithm is significantly worse than and better than that of DTAEA, respectively. The best median value is highlighted in boldface with gray background.
\end{tablenotes}
\end{table*}

\begin{table*}[htbp]
\tiny
\centering
\caption{Performance Comparisons of DTAEA and its Three Variants on MHV Metric}
\label{tab:MHVvariant}

\begin{tabular}{c|c|c|c|c|c|c|c|c|c}
\toprule
 \multirow{2}{*}{} & \multirow{3}{*}{$\tau_t$} &  \multicolumn{2}{c|}{DTAEA-$v1$} & \multicolumn{2}{c|}{DTAEA-$v2$} & \multicolumn{2}{c|}{DTAEA-$v3$} & \multicolumn{2}{c}{DTAEA} \\ \cmidrule{3-10}
 &  & MHV & R & MHV & R & MHV & R & MHV & R  \\ \midrule 
\multirow{4}{*}{F1}
 & 25 & \bb{${\text{99.85\%}}{(\text{4.60E-3})}$} & \bb{1.7} & ${\text{99.57\%}}{(\text{3.05E-3})}$ & 3 & ${\text{99.45\%}}{(\text{6.72E-3})}$$^{\dag}$ & 3.3 & ${\text{99.7\%}}{(\text{6.03E-3})}$ & \text{2} \\
 & 50 & ${\text{100.00\%}}{(\text{1.40E-4})}$$^{\dag}$ & 2.1 & ${\text{99.98\%}}{(\text{1.86E-4})}$$^{\dag}$ & 2.7 & ${\text{99.98\%}}{(\text{2.71E-4})}$$^{\dag}$ & 3.3 & \bb{${\text{100.0\%}}{(\text{8.39E-5})}$} & \bb{\text{1.8}} \\
 & 100 & ${\text{100.00\%}}{(\text{1.82E-5})}$$^{\dag}$ & 2.5 & ${\text{100.00\%}}{(\text{2.95E-5})}$$^{\dag}$ & 2.4 & ${\text{100.00\%}}{(\text{2.75E-5})}$$^{\dag}$ & 3.4 & \bb{${\text{100.0\%}}{(\text{1.84E-5})}$} & \bb{\text{1.7}} \\
 & 200 & ${\text{100.00\%}}{(\text{1.66E-5})}$$^{\dag}$ & 2.5 & ${\text{100.00\%}}{(\text{2.19E-5})}$$^{\dag}$ & 2.2 & ${\text{100.00\%}}{(\text{2.65E-5})}$$^{\dag}$ & 3.1 & \bb{${\text{100.0\%}}{(\text{2.69E-5})}$} & \bb{\text{2.1}} \\
\midrule

\multirow{4}{*}{F2}
 & 25 & ${\text{94.37\%}}{(\text{7.52E-5})}$$^{\dag}$ & 2.6 & ${\text{93.94\%}}{(\text{1.08E-3})}$$^{\dag}$ & 2.7 & ${\text{93.86\%}}{(\text{1.94E-3})}$$^{\dag}$ & 3.5 & \bb{${\text{94.4\%}}{(\text{6.26E-5})}$} & \bb{\text{1.2}} \\
 & 50 & ${\text{94.42\%}}{(\text{2.46E-5})}$$^{\dag}$ & 2.9 & ${\text{94.42\%}}{(\text{2.21E-4})}$$^{\dag}$ & 2.4 & ${\text{94.38\%}}{(\text{2.65E-4})}$$^{\dag}$ & 3.5 & \bb{${\text{94.4\%}}{(\text{3.09E-5})}$} & \bb{\text{1.2}} \\
 & 100 & ${\text{94.45\%}}{(\text{8.64E-6})}$$^{\dag}$ & 3.3 & ${\text{94.46\%}}{(\text{7.38E-6})}$$^{\dag}$ & 1.9 & ${\text{94.45\%}}{(\text{1.34E-5})}$$^{\dag}$ & 3.4 & \bb{${\text{94.5\%}}{(\text{8.88E-6})}$} & \bb{\text{1.5}} \\
 & 200 & ${\text{94.46\%}}{(\text{4.63E-6})}$$^{\dag}$ & 3.4 & ${\text{94.46\%}}{(\text{5.29E-6})}$$^{\dag}$ & 1.8 & ${\text{94.46\%}}{(\text{8.98E-6})}$$^{\dag}$ & 3.2 & \bb{${\text{94.5\%}}{(\text{4.03E-6})}$} & \bb{\text{1.7}} \\
\midrule

\multirow{4}{*}{F3}
 & 25 & \bb{${\text{91.08\%}}{(\text{6.51E-2})}$} & \bb{1.7} & ${\text{89.86\%}}{(\text{7.13E-2})}$ & 3 & ${\text{89.84\%}}{(\text{4.89E-2})}$ & 3.3 & ${\text{90.6\%}}{(\text{9.47E-2})}$ & \text{1.9} \\
 & 50 & \bb{${\text{94.24\%}}{(\text{2.74E-3})}$} & \bb{1.9} & ${\text{93.90\%}}{(\text{1.66E-3})}$$^{\dag}$ & 2.8 & ${\text{93.83\%}}{(\text{2.62E-3})}$$^{\dag}$ & 3.2 & ${\text{94.0\%}}{(\text{3.66E-3})}$ & \text{2.1} \\
 & 100 & ${\text{94.39\%}}{(\text{3.78E-4})}$$^{\dag}$ & 2.5 & ${\text{94.40\%}}{(\text{2.96E-4})}$$^{\dag}$ & 2.4 & ${\text{94.38\%}}{(\text{5.41E-4})}$$^{\dag}$ & 3.3 & \bb{${\text{94.4\%}}{(\text{3.32E-4})}$} & \bb{\text{1.8}} \\
 & 200 & ${\text{94.43\%}}{(\text{1.60E-4})}$$^{\dag}$ & 2.9 & ${\text{94.44\%}}{(\text{1.52E-4})}$$^{\dag}$ & 2.1 & ${\text{94.44\%}}{(\text{2.72E-4})}$$^{\dag}$ & 3.2 & \bb{${\text{94.4\%}}{(\text{1.26E-4})}$} & \bb{\text{1.8}} \\
\midrule

\multirow{4}{*}{F4}
 & 25 & ${\text{94.39\%}}{(\text{1.29E-4})}$ & 1.9 & ${\text{79.48\%}}{(\text{5.82E-5})}$$^{\dag}$ & 3.3 & ${\text{79.47\%}}{(\text{6.49E-5})}$$^{\dag}$ & 3.3 & \bb{${\text{94.4\%}}{(\text{1.60E-4})}$} & \bb{\text{1.5}} \\
 & 50 & ${\text{94.44\%}}{(\text{1.76E-5})}$$^{\dag}$ & 2.1 & ${\text{79.47\%}}{(\text{1.25E-4})}$$^{\dag}$ & 3.3 & ${\text{79.47\%}}{(\text{1.24E-4})}$$^{\dag}$ & 3.3 & \bb{${\text{94.4\%}}{(\text{1.16E-5})}$} & \bb{\text{1.3}} \\
 & 100 & ${\text{94.46\%}}{(\text{8.42E-6})}$$^{\dag}$ & 2.1 & ${\text{79.46\%}}{(\text{2.01E-4})}$$^{\dag}$ & 3.3 & ${\text{79.47\%}}{(\text{1.21E-4})}$$^{\dag}$ & 3.3 & \bb{${\text{94.5\%}}{(\text{3.26E-6})}$} & \bb{\text{1.2}} \\
 & 200 & ${\text{94.46\%}}{(\text{8.77E-6})}$$^{\dag}$ & 2 & ${\text{79.47\%}}{(\text{1.22E-2})}$$^{\dag}$ & 3.3 & ${\text{79.47\%}}{(\text{3.27E-2})}$$^{\dag}$ & 3.4 & \bb{${\text{94.5\%}}{(\text{4.19E-6})}$} & \bb{\text{1.2}} \\
\midrule

\multirow{4}{*}{F5}
 & 25 & \bb{${\text{90.37\%}}{(\text{5.17E-3})}$$^{\ddag}$} & 1.8 & ${\text{65.36\%}}{(\text{2.23E-2})}$$^{\dag}$ & 3.4 & ${\text{64.82\%}}{(\text{8.67E-3})}$$^{\dag}$ & 3.4 & {${\text{89.4\%}}{(\text{6.64E-3})}$} & \bb{\text{1.5}} \\
 & 50 & ${\text{93.53\%}}{(\text{1.34E-3})}$$^{\dag}$ & 1.8 & ${\text{46.58\%}}{(\text{1.64E-2})}$$^{\dag}$ & 3.5 & ${\text{46.18\%}}{(\text{2.98E-2})}$$^{\dag}$ & 3.5 & \bb{${\text{93.7\%}}{(\text{1.25E-3})}$} & \bb{\text{1.2}} \\
 & 100 & ${\text{94.53\%}}{(\text{4.58E-4})}$$^{\dag}$ & 1.8 & ${\text{15.51\%}}{(\text{3.19E-2})}$$^{\dag}$ & 3.5 & ${\text{16.54\%}}{(\text{3.35E-2})}$$^{\dag}$ & 3.5 & \bb{${\text{94.6\%}}{(\text{3.62E-4})}$} & \bb{\text{1.2}} \\
 & 200 & ${\text{94.73\%}}{(\text{3.03E-4})}$$^{\dag}$ & 1.8 & ${\text{11.56\%}}{(\text{1.35E-2})}$$^{\dag}$ & 3.4 & ${\text{11.09\%}}{(\text{6.06E-3})}$$^{\dag}$ & 3.6 & \bb{${\text{94.8\%}}{(\text{1.84E-4})}$} & \bb{\text{1.2}} \\
\midrule

\multirow{4}{*}{F6}
 & 25 & \bb{${\text{89.31\%}}{(\text{1.18E-2})}$} & 1.7 & ${\text{51.61\%}}{(\text{4.21E-2})}$$^{\dag}$ & 3.5 & ${\text{51.54\%}}{(\text{2.37E-2})}$$^{\dag}$ & 3.5 & {${\text{89.1\%}}{(\text{9.72E-3})}$} & \bb{\text{1.3}} \\
 & 50 & ${\text{92.93\%}}{(\text{1.19E-3})}$$^{\dag}$ & 1.8 & ${\text{33.26\%}}{(\text{1.84E-2})}$$^{\dag}$ & 3.5 & ${\text{33.80\%}}{(\text{2.49E-2})}$$^{\dag}$ & 3.5 & \bb{${\text{93.7\%}}{(\text{1.45E-3})}$} & \bb{\text{1.2}} \\
 & 100 & ${\text{93.92\%}}{(\text{5.32E-4})}$$^{\dag}$ & 1.8 & ${\text{7.99\%}}{(\text{8.41E-3})}$$^{\dag}$ & 3.5 & ${\text{8.22\%}}{(\text{7.57E-3})}$$^{\dag}$ & 3.5 & \bb{${\text{94.5\%}}{(\text{3.50E-4})}$} & \bb{\text{1.2}} \\
 & 200 & ${\text{94.14\%}}{(\text{3.41E-4})}$$^{\dag}$ & 1.8 & ${\text{7.92\%}}{(\text{7.58E-3})}$$^{\dag}$ & 3.5 & ${\text{8.07\%}}{(\text{9.59E-3})}$$^{\dag}$ & 3.5 & \bb{${\text{94.7\%}}{(\text{3.39E-4})}$} & \bb{\text{1.2}} \\
\bottomrule
\end{tabular}
\begin{tablenotes}
\item[1] R denotes the global rank assigned to each algorithm by averaging the ranks obtained at all time steps. Wilcoxon's rank sum test at a 0.05 significance level is performed between DTAEA and each of DTAEA-$v1$, DTAEA-$v2$ and DTAEA-$v3$. $^{\dag}$ and $^{\ddag}$ denote the performance of the corresponding algorithm is significantly worse than and better than that of DTAEA, respectively. The best median value is highlighted in boldface with gray background.
\end{tablenotes}
\end{table*}

\subsubsection{Performance Comparisons on a Different Changing Sequence}
\label{sec:different}

In~\pref{sec:empirical-F14} and~\pref{sec:empirical-F56}, the experiments only consider the scenarios where the number of objectives increases or decreases by one at each time step. A natural question is: how is the performance of our proposed algorithm under the circumstance where the number of objectives changes in a different sequence? Without loss of generality, this further experiment considers the time varying number of objectives $m(t)$ as follows:
\begin{equation}
m(t)=
    \begin{cases}
        3, & \quad t=1\\
        m(t-1)+2, & \quad t\in[2,3]\\
        m(t-1)-2, & \quad t\in[4,5]\\
        m(t-1)-1, & \quad t=6
    \end{cases}
\label{eq:mt2}
\end{equation}
where $t\in\{1,\cdots,6\}$ is a discrete time. Here we also consider four different frequencies of change, i.e., $\tau_t$ is as 25, 50, 100 and 200, respectively. From the empirical results shown in~\pref{tab:differentMIGD} and~\pref{tab:differentMHV}, we can clearly see that our proposed DTAEA is the best optimizer on almost all comparisons (92 out of 96 for MIGD and 88 out 96 for MHV). Similar to the observations from the previous sections, the performance of DTAEA might not be stable under a high frequency of change; while its performance becomes constantly competitive with the increase of $\tau_t$.

\begin{table*}[htbp]
\tiny
\centering
\caption{Performance Comparisons on MIGD Metric with a Different Changing Sequence.}
\label{tab:differentMIGD}

\begin{tabular}{c|c|c|c|c|c|c|c|c|c|c|c}
\toprule
 \multirow{2}{*}{} & \multirow{3}{*}{$\tau_t$} &  \multicolumn{2}{c|}{NSGA-II} & \multicolumn{2}{c|}{DNSGA-II} & \multicolumn{2}{c|}{MOEA/D} & \multicolumn{2}{c|}{MOEA/D-KF} & \multicolumn{2}{c}{DTAEA} \\ \cmidrule{3-12}
 &  & MIGD & R & MIGD & R & MIGD & R & MIGD & R & MIGD & R \\ \midrule 
\multirow{4}{*}{F1}
 & 25 & ${\text{2.97E-3}}{(\text{3.40E-3})}$$^{\dag}$ & 2.7 & ${\text{2.00E-3}}{(\text{3.56E-3})}$$^{\dag}$ & 2.8 & ${\text{2.40E-3}}{(\text{8.80E-3})}$$^{\dag}$ & 3.8 & ${\text{1.67E-2}}{(\text{1.81E-2})}$$^{\dag}$ & 4.3 & \bb{${\text{1.10E-3}}{(\text{6.03E-4})}$} & \bb{\text{1.4}} \\
 & 50 & ${\text{1.31E-2}}{(\text{1.26E-2})}$$^{\dag}$ & 3.8 & ${\text{1.33E-2}}{(\text{1.21E-2})}$$^{\dag}$ & 3.8 & ${\text{7.78E-4}}{(\text{2.92E-4})}$$^{\dag}$ & 3 & ${\text{1.80E-3}}{(\text{2.10E-3})}$$^{\dag}$ & 3.5 & \bb{${\text{3.38E-4}}{(\text{3.38E-5})}$} & \bb{\text{1}} \\
 & 100 & ${\text{2.54E-2}}{(\text{3.50E-2})}$$^{\dag}$ & 4 & ${\text{3.23E-2}}{(\text{3.94E-2})}$$^{\dag}$ & 4.1 & ${\text{5.73E-4}}{(\text{4.71E-5})}$$^{\dag}$ & 2.8 & ${\text{7.27E-4}}{(\text{4.18E-4})}$$^{\dag}$ & 3.2 & \bb{${\text{2.92E-4}}{(\text{1.75E-6})}$} & \bb{\text{1}} \\
 & 200 & ${\text{3.11E-3}}{(\text{4.08E-2})}$$^{\dag}$ & 4.3 & ${\text{1.25E-2}}{(\text{3.68E-2})}$$^{\dag}$ & 4.3 & ${\text{5.02E-4}}{(\text{2.16E-4})}$$^{\dag}$ & 2.4 & ${\text{6.39E-4}}{(\text{1.89E-3})}$$^{\dag}$ & 3 & \bb{${\text{2.90E-4}}{(\text{2.10E-6})}$} & \bb{\text{1}} \\
\midrule

\multirow{4}{*}{F2}
 & 25 & ${\text{1.41E-3}}{(\text{5.72E-5})}$$^{\dag}$ & 3.2 & ${\text{1.48E-3}}{(\text{1.11E-4})}$$^{\dag}$ & 3.4 & ${\text{1.81E-3}}{(\text{8.28E-5})}$$^{\dag}$ & 3.5 & ${\text{1.82E-3}}{(\text{4.91E-5})}$$^{\dag}$ & 3.9 & \bb{${\text{1.07E-3}}{(\text{5.22E-6})}$} & \bb{\text{1}} \\
 & 50 & ${\text{1.54E-3}}{(\text{1.01E-4})}$$^{\dag}$ & 3.8 & ${\text{1.63E-3}}{(\text{1.28E-4})}$$^{\dag}$ & 4 & ${\text{1.70E-3}}{(\text{6.46E-5})}$$^{\dag}$ & 2.9 & ${\text{1.71E-3}}{(\text{6.81E-5})}$$^{\dag}$ & 3.3 & \bb{${\text{1.06E-3}}{(\text{3.42E-6})}$} & \bb{\text{1}} \\
 & 100 & ${\text{1.82E-3}}{(\text{1.32E-4})}$$^{\dag}$ & 4.1 & ${\text{1.90E-3}}{(\text{9.04E-5})}$$^{\dag}$ & 4.2 & ${\text{1.62E-3}}{(\text{2.88E-5})}$$^{\dag}$ & 2.7 & ${\text{1.64E-3}}{(\text{2.74E-5})}$$^{\dag}$ & 3.1 & \bb{${\text{1.04E-3}}{(\text{1.83E-6})}$} & \bb{\text{1}} \\
 & 200 & ${\text{1.98E-3}}{(\text{1.03E-4})}$$^{\dag}$ & 4.1 & ${\text{1.96E-3}}{(\text{1.38E-4})}$$^{\dag}$ & 4.2 & ${\text{1.58E-3}}{(\text{3.77E-5})}$$^{\dag}$ & 2.8 & ${\text{1.56E-3}}{(\text{2.12E-5})}$$^{\dag}$ & 2.8 & \bb{${\text{1.03E-3}}{(\text{1.34E-6})}$} & \bb{\text{1.1}} \\
\midrule

\multirow{4}{*}{F3}
 & 25 & ${\text{8.26E-3}}{(\text{9.99E-3})}$$^{\dag}$ & 2.8 & ${\text{8.09E-3}}{(\text{5.22E-3})}$$^{\dag}$ & 2.8 & ${\text{4.88E-3}}{(\text{4.31E-3})}$$^{\dag}$ & 3.8 & ${\text{2.11E-2}}{(\text{4.73E-2})}$$^{\dag}$ & 4.2 & \bb{${\text{2.66E-3}}{(\text{3.29E-3})}$} & \bb{\text{1.4}} \\
 & 50 & ${\text{1.33E-2}}{(\text{1.64E-2})}$$^{\dag}$ & 3.4 & ${\text{1.09E-2}}{(\text{1.84E-2})}$$^{\dag}$ & 3.4 & ${\text{2.79E-3}}{(\text{6.58E-4})}$$^{\dag}$ & 3.2 & ${\text{5.05E-3}}{(\text{6.62E-3})}$$^{\dag}$ & 3.8 & \bb{${\text{1.20E-3}}{(\text{1.04E-4})}$} & \bb{\text{1.1}} \\
 & 100 & ${\text{9.02E-2}}{(\text{5.18E-2})}$$^{\dag}$ & 3.9 & ${\text{9.16E-2}}{(\text{4.43E-2})}$$^{\dag}$ & 4 & ${\text{2.42E-3}}{(\text{2.19E-3})}$$^{\dag}$ & 2.8 & ${\text{2.41E-3}}{(\text{1.63E-3})}$$^{\dag}$ & 3.2 & \bb{${\text{1.06E-3}}{(\text{8.73E-6})}$} & \bb{\text{1}} \\
 & 200 & ${\text{1.51E-1}}{(\text{1.28E-1})}$$^{\dag}$ & 4.4 & ${\text{1.76E-1}}{(\text{9.13E-2})}$$^{\dag}$ & 4.4 & ${\text{1.69E-3}}{(\text{5.03E-5})}$$^{\dag}$ & 2.4 & ${\text{2.01E-3}}{(\text{2.33E-4})}$$^{\dag}$ & 2.9 & \bb{${\text{1.05E-3}}{(\text{7.16E-6})}$} & \bb{\text{1}} \\
\midrule

\multirow{4}{*}{F4}
 & 25 & ${\text{2.64E-3}}{(\text{4.22E-4})}$$^{\dag}$ & 2.5 & ${\text{2.79E-3}}{(\text{3.27E-4})}$$^{\dag}$ & 3 & ${\text{4.08E-3}}{(\text{7.27E-4})}$$^{\dag}$ & 4.2 & ${\text{3.91E-3}}{(\text{3.17E-4})}$$^{\dag}$ & 4.3 & \bb{${\text{1.10E-3}}{(\text{2.29E-5})}$} & \bb{\text{1}} \\
 & 50 & ${\text{1.77E-3}}{(\text{3.40E-4})}$$^{\dag}$ & 2.8 & ${\text{2.26E-3}}{(\text{3.41E-4})}$$^{\dag}$ & 3.5 & ${\text{3.44E-3}}{(\text{5.08E-4})}$$^{\dag}$ & 4 & ${\text{3.03E-3}}{(\text{3.74E-4})}$$^{\dag}$ & 3.8 & \bb{${\text{1.06E-3}}{(\text{1.24E-3})}$} & \bb{\text{1}} \\
 & 100 & ${\text{2.19E-3}}{(\text{2.30E-4})}$$^{\dag}$ & 3.2 & ${\text{2.29E-3}}{(\text{3.31E-4})}$$^{\dag}$ & 3.4 & ${\text{3.04E-3}}{(\text{3.42E-4})}$$^{\dag}$ & 3.8 & ${\text{3.14E-3}}{(\text{3.88E-4})}$$^{\dag}$ & 3.7 & \bb{${\text{1.04E-3}}{(\text{2.78E-6})}$} & \bb{\text{1}} \\
 & 200 & ${\text{2.26E-3}}{(\text{1.92E-4})}$$^{\dag}$ & 3.3 & ${\text{2.24E-3}}{(\text{1.88E-4})}$$^{\dag}$ & 3.1 & ${\text{2.71E-3}}{(\text{1.56E-4})}$$^{\dag}$ & 3.8 & ${\text{2.68E-3}}{(\text{8.39E-4})}$$^{\dag}$ & 3.8 & \bb{${\text{1.03E-3}}{(\text{2.42E-6})}$} & \bb{\text{1}} \\
\midrule

\multirow{4}{*}{F5}
 & 25 & ${\text{7.14E-3}}{(\text{3.50E-3})}$$^{\dag}$ & 3.9 & ${\text{3.87E-3}}{(\text{5.66E-4})}$$^{\dag}$ & 3.5 & ${\text{2.53E-3}}{(\text{2.74E-4})}$$^{\ddag}$ & 2.5 & \bb{${\text{2.30E-3}}{(\text{2.67E-4})}$$^{\ddag}$} & \bb{\text{2.5}} & ${\text{2.94E-3}}{(\text{3.79E-4})}$ & 2.7 \\
 & 50 & ${\text{2.83E-3}}{(\text{2.91E-4})}$$^{\dag}$ & 3.6 & ${\text{3.27E-3}}{(\text{3.15E-4})}$$^{\dag}$ & 4.4 & ${\text{1.99E-3}}{(\text{9.74E-5})}$$^{\dag}$ & 2.6 & ${\text{1.94E-3}}{(\text{1.17E-4})}$$^{\dag}$ & 2.4 & \bb{${\text{1.84E-3}}{(\text{8.27E-5})}$} & \bb{\text{2}} \\
 & 100 & ${\text{1.72E-3}}{(\text{2.23E-4})}$$^{\dag}$ & 3 & ${\text{1.84E-3}}{(\text{1.50E-4})}$$^{\dag}$ & 3.8 & ${\text{1.80E-3}}{(\text{7.46E-5})}$$^{\dag}$ & 3.3 & ${\text{1.74E-3}}{(\text{1.32E-4})}$$^{\dag}$ & 3.3 & \bb{${\text{1.29E-3}}{(\text{1.26E-5})}$} & \bb{\text{1.6}} \\
 & 200 & ${\text{1.57E-3}}{(\text{1.12E-4})}$$^{\dag}$ & 2.8 & ${\text{1.83E-3}}{(\text{1.01E-4})}$$^{\dag}$ & 4.1 & ${\text{1.68E-3}}{(\text{3.56E-5})}$$^{\dag}$ & 3.2 & ${\text{1.64E-3}}{(\text{7.30E-5})}$$^{\dag}$ & 3 & \bb{${\text{1.19E-3}}{(\text{1.29E-5})}$} & \bb{\text{1.9}} \\
\midrule

\multirow{4}{*}{F6}
 & 25 & ${\text{8.61E-3}}{(\text{1.74E-3})}$$^{\dag}$ & 4.4 & ${\text{5.03E-3}}{(\text{5.70E-4})}$$^{\dag}$ & 3.5 & ${\text{2.58E-3}}{(\text{3.54E-4})}$$^{\ddag}$ & 2.3 & \bb{${\text{2.57E-3}}{(\text{3.77E-4})}$$^{\ddag}$} & 2.6 & ${\text{3.04E-3}}{(\text{1.71E-4})}$ & \bb{\text{2.2}} \\
 & 50 & ${\text{2.92E-3}}{(\text{2.43E-4})}$$^{\dag}$ & 4 & ${\text{3.72E-3}}{(\text{6.24E-4})}$$^{\dag}$ & 3.8 & ${\text{2.24E-3}}{(\text{2.14E-4})}$$^{\dag}$ & 2.5 & ${\text{2.38E-3}}{(\text{1.57E-4})}$$^{\dag}$ & 2.7 & \bb{${\text{2.05E-3}}{(\text{1.03E-4})}$} & \bb{\text{1.9}} \\
 & 100 & ${\text{2.69E-3}}{(\text{3.24E-4})}$$^{\dag}$ & 3.9 & ${\text{2.62E-3}}{(\text{1.58E-4})}$$^{\dag}$ & 4.3 & ${\text{1.89E-3}}{(\text{9.26E-5})}$$^{\dag}$ & 2.5 & ${\text{1.98E-3}}{(\text{1.20E-4})}$$^{\dag}$ & 3 & \bb{${\text{1.35E-3}}{(\text{1.34E-5})}$} & \bb{\text{1.4}} \\
 & 200 & ${\text{2.89E-3}}{(\text{3.07E-4})}$$^{\dag}$ & 4.3 & ${\text{2.53E-3}}{(\text{1.01E-4})}$$^{\dag}$ & 4 & ${\text{1.87E-3}}{(\text{7.80E-5})}$$^{\dag}$ & 2.4 & ${\text{1.85E-3}}{(\text{7.47E-5})}$$^{\dag}$ & 2.9 & \bb{${\text{1.27E-3}}{(\text{2.42E-5})}$} & \bb{\text{1.4}} \\
\bottomrule
\end{tabular}

\begin{tablenotes}
\item[1] R denotes the global rank assigned to each algorithm by averaging the ranks obtained at all time steps. Wilcoxon's rank sum test at a 0.05 significance level is performed between DTAEA and each of NSGA-II, DNSGA-II, MOEA/D and MOEA/D-KF. $^{\dag}$ and $^{\ddag}$ denote the performance of the corresponding algorithm is significantly worse than and better than that of DTAEA, respectively. The best median value is highlighted in boldface with gray background.
\end{tablenotes}

\end{table*}

\begin{table*}[htbp]
\tiny
\centering
\caption{Performance Comparisons on MHV Metric with a Different Changing Sequence.}
\label{tab:differentMHV}

\begin{tabular}{c|c|c|c|c|c|c|c|c|c|c|c}
\toprule
 \multirow{2}{*}{} & \multirow{3}{*}{$\tau_t$} &  \multicolumn{2}{c|}{NSGA-II} & \multicolumn{2}{c|}{DNSGA-II} & \multicolumn{2}{c|}{MOEA/D} & \multicolumn{2}{c|}{MOEA/D-KF} & \multicolumn{2}{c}{DTAEA} \\ \cmidrule{3-12}
 &  & MHV & R & MHV & R & MHV & R & MHV & R & MHV & R \\ \midrule 
\multirow{4}{*}{F1}
 & 25 & ${\text{87.4\%}}{(\text{1.13E-1})}$$^{\dag}$ & 3.1 & ${\text{92.5\%}}{(\text{1.33E-1})}$$^{\dag}$ & 3.1 & ${\text{91.6\%}}{(\text{3.74E-1})}$$^{\dag}$ & 3.5 & ${\text{70.2\%}}{(\text{1.77E-1})}$$^{\dag}$ & 3.9 & \bb{${\text{98.2\%}}{(\text{2.31E-2})}$} & \bb{\text{1.3}} \\
 & 50 & ${\text{66.4\%}}{(\text{1.49E-1})}$$^{\dag}$ & 4.2 & ${\text{63.6\%}}{(\text{2.28E-1})}$$^{\dag}$ & 4.2 & ${\text{99.4\%}}{(\text{7.77E-3})}$$^{\dag}$ & 2.5 & ${\text{92.7\%}}{(\text{1.24E-1})}$$^{\dag}$ & 3 & \bb{${\text{100.0\%}}{(\text{1.26E-4})}$} & \bb{\text{1}} \\
 & 100 & ${\text{71.8\%}}{(\text{1.12E-1})}$$^{\dag}$ & 4.1 & ${\text{71.4\%}}{(\text{8.87E-2})}$$^{\dag}$ & 4 & ${\text{99.9\%}}{(\text{1.26E-3})}$$^{\dag}$ & 2.7 & ${\text{99.5\%}}{(\text{1.13E-2})}$$^{\dag}$ & 3.2 & \bb{${\text{100.0\%}}{(\text{5.61E-5})}$} & \bb{\text{1}} \\
 & 200 & ${\text{84.8\%}}{(\text{1.97E-1})}$$^{\dag}$ & 4 & ${\text{82.1\%}}{(\text{2.41E-1})}$$^{\dag}$ & 4 & ${\text{100.0\%}}{(\text{1.08E-2})}$$^{\dag}$ & 2.6 & ${\text{99.7\%}}{(\text{9.86E-2})}$$^{\dag}$ & 3.3 & \bb{${\text{100.0\%}}{(\text{1.73E-5})}$} & \bb{\text{1}} \\
\midrule

\multirow{4}{*}{F2}
 & 25 & ${\text{91.1\%}}{(\text{3.29E-3})}$$^{\dag}$ & 3.8 & ${\text{91.5\%}}{(\text{2.75E-3})}$$^{\dag}$ & 3.8 & ${\text{92.1\%}}{(\text{1.93E-3})}$$^{\dag}$ & 2.9 & ${\text{92.0\%}}{(\text{1.49E-3})}$$^{\dag}$ & 3.5 & \bb{${\text{92.9\%}}{(\text{1.06E-4})}$} & \bb{\text{1}} \\
 & 50 & ${\text{91.6\%}}{(\text{3.16E-3})}$$^{\dag}$ & 4.1 & ${\text{91.6\%}}{(\text{3.23E-3})}$$^{\dag}$ & 4 & ${\text{92.3\%}}{(\text{1.01E-3})}$$^{\dag}$ & 2.8 & ${\text{92.3\%}}{(\text{1.28E-3})}$$^{\dag}$ & 3.1 & \bb{${\text{92.9\%}}{(\text{5.67E-5})}$} & \bb{\text{1}} \\
 & 100 & ${\text{90.6\%}}{(\text{8.76E-3})}$$^{\dag}$ & 4.3 & ${\text{90.3\%}}{(\text{8.01E-3})}$$^{\dag}$ & 4.5 & ${\text{92.5\%}}{(\text{6.31E-4})}$$^{\dag}$ & 2.4 & ${\text{92.4\%}}{(\text{6.47E-4})}$$^{\dag}$ & 2.7 & \bb{${\text{92.9\%}}{(\text{1.65E-5})}$} & \bb{\text{1}} \\
 & 200 & ${\text{89.7\%}}{(\text{7.17E-3})}$$^{\dag}$ & 4.4 & ${\text{89.8\%}}{(\text{9.25E-3})}$$^{\dag}$ & 4.5 & ${\text{92.5\%}}{(\text{5.50E-4})}$$^{\dag}$ & 2.5 & ${\text{92.6\%}}{(\text{4.09E-4})}$$^{\dag}$ & 2.3 & \bb{${\text{93.0\%}}{(\text{5.51E-6})}$} & \bb{\text{1.1}} \\
\midrule

\multirow{4}{*}{F3}
 & 25 & ${\text{76.0\%}}{(\text{5.96E-2})}$ & 3.1 & ${\text{75.8\%}}{(\text{1.80E-2})}$ & 3.3 & ${\text{78.4\%}}{(\text{3.14E-1})}$$^{\dag}$ & 3.3 & ${\text{62.2\%}}{(\text{6.96E-2})}$$^{\dag}$ & 3.7 & \bb{${\text{85.5\%}}{(\text{1.92E-1})}$} & \bb{\text{1.6}} \\
 & 50 & ${\text{63.2\%}}{(\text{1.23E-1})}$$^{\dag}$ & 3.8 & ${\text{62.9\%}}{(\text{2.14E-1})}$$^{\dag}$ & 3.7 & ${\text{86.8\%}}{(\text{3.79E-2})}$$^{\dag}$ & 2.8 & ${\text{74.3\%}}{(\text{2.87E-1})}$$^{\dag}$ & 3.6 & \bb{${\text{92.5\%}}{(\text{3.81E-3})}$} & \bb{\text{1.1}} \\
 & 100 & ${\text{52.5\%}}{(\text{9.26E-2})}$$^{\dag}$ & 4.2 & ${\text{54.2\%}}{(\text{5.61E-2})}$$^{\dag}$ & 4.1 & ${\text{87.6\%}}{(\text{1.24E-1})}$$^{\dag}$ & 2.7 & ${\text{88.0\%}}{(\text{1.04E-1})}$$^{\dag}$ & 3 & \bb{${\text{92.9\%}}{(\text{4.27E-4})}$} & \bb{\text{1}} \\
 & 200 & ${\text{60.7\%}}{(\text{6.73E-2})}$$^{\dag}$ & 4.4 & ${\text{60.3\%}}{(\text{7.66E-2})}$$^{\dag}$ & 4.3 & ${\text{92.2\%}}{(\text{3.02E-3})}$$^{\dag}$ & 2.3 & ${\text{90.7\%}}{(\text{2.05E-2})}$$^{\dag}$ & 2.9 & \bb{${\text{92.9\%}}{(\text{3.01E-4})}$} & \bb{\text{1}} \\
\midrule

\multirow{4}{*}{F4}
 & 25 & ${\text{90.3\%}}{(\text{5.61E-3})}$$^{\dag}$ & 3 & ${\text{89.5\%}}{(\text{5.88E-3})}$$^{\dag}$ & 3.6 & ${\text{87.5\%}}{(\text{4.24E-2})}$$^{\dag}$ & 3.8 & ${\text{88.7\%}}{(\text{1.84E-2})}$$^{\dag}$ & 3.7 & \bb{${\text{92.9\%}}{(\text{1.65E-4})}$} & \bb{\text{1}} \\
 & 50 & ${\text{91.3\%}}{(\text{1.32E-2})}$$^{\dag}$ & 3.3 & ${\text{89.2\%}}{(\text{1.91E-2})}$$^{\dag}$ & 3.8 & ${\text{89.9\%}}{(\text{2.30E-2})}$$^{\dag}$ & 3.5 & ${\text{90.6\%}}{(\text{5.72E-3})}$$^{\dag}$ & 3.5 & \bb{${\text{92.9\%}}{(\text{4.48E-2})}$} & \bb{\text{1}} \\
 & 100 & ${\text{88.9\%}}{(\text{1.26E-2})}$$^{\dag}$ & 3.5 & ${\text{88.2\%}}{(\text{2.45E-2})}$$^{\dag}$ & 3.6 & ${\text{90.5\%}}{(\text{3.51E-3})}$$^{\dag}$ & 3.4 & ${\text{90.4\%}}{(\text{9.23E-3})}$$^{\dag}$ & 3.5 & \bb{${\text{93.0\%}}{(\text{9.81E-6})}$} & \bb{\text{1}} \\
 & 200 & ${\text{87.2\%}}{(\text{1.65E-2})}$$^{\dag}$ & 4.3 & ${\text{87.7\%}}{(\text{1.92E-2})}$$^{\dag}$ & 4 & ${\text{90.7\%}}{(\text{5.10E-4})}$$^{\dag}$ & 2.9 & ${\text{90.7\%}}{(\text{1.84E-2})}$$^{\dag}$ & 2.8 & \bb{${\text{93.0\%}}{(\text{4.10E-6})}$} & \bb{\text{1}} \\
\midrule

\multirow{4}{*}{F5}
 & 25 & ${\text{55.6\%}}{(\text{5.86E-2})}$$^{\dag}$ & 3.9 & ${\text{77.3\%}}{(\text{3.98E-2})}$$^{\dag}$ & 3.7 & ${\text{90.3\%}}{(\text{1.23E-2})}$$^{\ddag}$ & \bb{\text{2.2}}  & \bb{${\text{91.0\%}}{(\text{4.75E-3})}$$^{\ddag}$} & 2.4 & ${\text{84.7\%}}{(\text{2.59E-2})}$ & 2.8\\
 & 50 & ${\text{83.9\%}}{(\text{1.80E-2})}$$^{\dag}$ & 4.1 & ${\text{83.3\%}}{(\text{2.76E-2})}$$^{\dag}$ & 4.3 & ${\text{91.8\%}}{(\text{3.26E-3})}$ & 2.2 & \bb{${\text{91.9\%}}{(\text{4.21E-3})}$} & \bb{\text{1.8}} & ${\text{91.2\%}}{(\text{2.63E-3})}$ & 2.6 \\
 & 100 & ${\text{91.4\%}}{(\text{8.99E-3})}$$^{\dag}$ & 3.3 & ${\text{91.7\%}}{(\text{3.53E-3})}$$^{\dag}$ & 4 & ${\text{92.2\%}}{(\text{2.68E-3})}$$^{\dag}$ & 2.9 & ${\text{92.1\%}}{(\text{5.46E-3})}$$^{\dag}$ & 2.7 & \bb{${\text{92.3\%}}{(\text{4.12E-4})}$} & \bb{\text{2.1}} \\
 & 200 & ${\text{92.1\%}}{(\text{5.78E-3})}$$^{\dag}$ & 3.3 & ${\text{91.6\%}}{(\text{3.04E-3})}$$^{\dag}$ & 4.2 & ${\text{92.4\%}}{(\text{5.56E-4})}$$^{\dag}$ & 3 & ${\text{92.3\%}}{(\text{3.23E-3})}$$^{\dag}$ & 2.5 & \bb{${\text{92.7\%}}{(\text{2.17E-4})}$} & \bb{\text{2.1}} \\
\midrule

\multirow{4}{*}{F6}
 & 25 & ${\text{54.3\%}}{(\text{5.32E-2})}$$^{\dag}$ & 4.6 & ${\text{68.8\%}}{(\text{3.85E-2})}$$^{\dag}$ & 3.5 & ${\text{90.8\%}}{(\text{1.37E-2})}$$^{\ddag}$ & \bb{\text{2}} & \bb{${\text{91.1\%}}{(\text{7.15E-3})}$$^{\ddag}$} & 2.3 & ${\text{85.2\%}}{(\text{1.40E-2})}$ & 2.6 \\
 & 50 & ${\text{86.6\%}}{(\text{1.04E-2})}$$^{\dag}$ & 4.5 & ${\text{80.1\%}}{(\text{4.06E-2})}$$^{\dag}$ & 3.9 & \bb{${\text{91.8\%}}{(\text{6.93E-3})}$$^{\ddag}$} & \bb{\text{2}} & ${\text{91.7\%}}{(\text{4.34E-3})}$$^{\ddag}$ & 2 & ${\text{90.7\%}}{(\text{3.66E-3})}$ & 2.7 \\
 & 100 & ${\text{88.4\%}}{(\text{1.29E-2})}$$^{\dag}$ & 4.7 & ${\text{90.5\%}}{(\text{3.61E-3})}$$^{\dag}$ & 4 & ${\text{92.2\%}}{(\text{1.36E-3})}$$^{\dag}$ & 2.3 & ${\text{92.1\%}}{(\text{4.34E-3})}$$^{\dag}$ & 2.5 & \bb{${\text{92.3\%}}{(\text{4.21E-4})}$} & \bb{\text{1.6}} \\
 & 200 & ${\text{88.3\%}}{(\text{1.27E-2})}$$^{\dag}$ & 4.7 & ${\text{90.9\%}}{(\text{3.60E-3})}$$^{\dag}$ & 3.9 & ${\text{92.4\%}}{(\text{8.28E-4})}$$^{\dag}$ & 2.3 & ${\text{92.3\%}}{(\text{2.97E-3})}$$^{\dag}$ & 2.4 & \bb{${\text{92.6\%}}{(\text{4.68E-4})}$} & \bb{\text{1.6}} \\
\bottomrule
\end{tabular}

\begin{tablenotes}
\item[1] R denotes the global rank assigned to each algorithm by averaging the ranks obtained at all time steps. Wilcoxon's rank sum test at a 0.05 significance level is performed between DTAEA and each of NSGA-II, DNSGA-II, MOEA/D and MOEA/D-KF. $^{\dag}$ and $^{\ddag}$ denote the performance of the corresponding algorithm is significantly worse than and better than that of DTAEA, respectively. The best median value is highlighted in boldface with gray background.
\end{tablenotes}

\end{table*}

% !Tex root = main.tex

\section{Conclusions}
\label{sec:conclusions}

Different from the current studies on DMO, which mainly focus on problems with time-dependent objective functions, this paper considers the DMOP with a time varying number of objectives. This type of dynamics brings the expansion or contraction of the PF or PS manifold when increasing or decreasing the number of objectives. In this case, the existing dynamic handling techniques are not able to handle this scenario. To address this challenge, this paper has proposed a dynamic two-archive EA (denoted as DTAEA) for handling the DMOP with a changing number of objectives. More specifically, DTAEA simultaneously maintains two co-evolving populations, i.e., the CA and the DA, during the evolution process. In particular, they have complementary effects: the CA concerns more about the convergence while the DA concerns more about the diversity. The CA and the DA are separately reconstructed whenever the environment changes. In the meanwhile, they are maintained in different manners. By using a restricted mating selection mechanism, DTAEA takes advantages of the complementary effects of the CA and the DA to strike the balance between convergence and diversity all the time. Comprehensive experiments on a set of dynamic multi-objective benchmark problems fully demonstrate the superiority of DTAEA for handling the DMOP with a dynamically changing number of objectives. In particular, it is able to effectively and efficiently track the expansion or contraction of the PF or PS manifold whenever the environment changes. In the experiments, we also noticed that some state-of-the-art dynamic EMO algorithms even showed inferior performance than their stationary counterparts.

This paper is the first attempt towards a systematic investigation of the methods for handling the DMOP with a changing number of objectives. Much more attention and effort should be required on this topic. In future, it is interesting to investigate the dynamic environments where the changes are hard to detect and do not vary regularly. Constrained optimization problems, which are ubiquitous in the real-life application scenarios, have not been considered in this work. It is worth considering dynamic constraints in the dynamic environment as well and thus requires effective constraint handling techniques accordingly~\cite{LiDZK15}. Furthermore, this work focuses on designing an effective environmental selection operator to adapt to the changing environment. It is also interesting to investigate the reproduction operation which can adapt the search behavior autonomously~\cite{LiFKZ14,LiFK11,LiKWCR12,IJUFKS13,LiK14,LiKWTM13} according to the current landscape. In addition, our recently developed efficient non-domination level update method~\cite{LiDZZ16} can be considered to accelerate the process of identifying the current non-domination level structure. The benchmark problems used in this paper are developed from the classic DTLZ benchmark suite. It is interesting to develop more challenging benchmark problems with additional characteristics, e.g., from the WFG~\cite{HubandHBW06} and the CEC 2009 competition~\cite{zhang2008multiobjective} benchmark suites. Last but not the least, new performance metrics are also needed to better facilitate the analysis and comparisons of dynamic EMO algorithms.

% !tex root = main.tex

%\appendices

\section{Appendix: Proof of \pref{theorem:subset}}
\label{sec:theorem}

\begin{proof}
    Let us consider the scenario of increasing the number of objectives at first, where we prove the theorem by contradiction. At time step $t_1$, we assume that $\mathbf{x}^1$ and $\mathbf{x}^2$ are in $PS_{t_1}$. Accordingly, $\mathbf{F}(\mathbf{x}^1,t_1)$ and $\mathbf{F}(\mathbf{x}^2,t_1)$ are in $PF_{t_1}$. At time step $t_2$, we increase the number of objectives by one, i.e., $m(t_2)=m(t_1)+1$. Assume that $\mathbf{x}^1$ is still in $PS_{t_2}$ whereas $\mathbf{x}^2$ is not, thus we have $\mathbf{x}^1\preceq_{t_2}\mathbf{x}^2$. In other words, $\forall i\in\{1,\cdots,m(t_1),m(t_2)\}$ (i.e., $\forall i\in\{1,\cdots,m(t_1),m(t_1)+1\}$), $f_i(\mathbf{x}^1,t_2)\leq f_i(\mathbf{x}^2,t_2)$; and $\exists j\in\{1,\cdots,m(t_1),m(t_1)+1\}$, $f_j(\mathbf{x}^1,t_2)<f_j(\mathbf{x}^2,t_2)$. This contradicts the assumption that $\mathbf{x}^1$ and $\mathbf{x}^2$ are non-dominated from each other at time step $t_1$. Then, we conclude that $PF_{t_1}$ is a subset of $PF_{t_2}$ when increasing the number of objectives.

    Now let us consider the scenario of decreasing the number of objectives. At time step $t_1$, we assume that $\mathbf{x}^1$ and $\mathbf{x}^2$ are in $PS_{t_1}$. Accordingly, $\mathbf{F}(\mathbf{x}^1,t_1)$ and $\mathbf{F}(\mathbf{x}^2,t_1)$ are in $PF_{t_1}$. Furthermore, we assume that $\forall i\in\{1,\cdots,m(t_1)-1\}$, we have $f_i(\mathbf{x}^1,t_1)\leq f_i(\mathbf{x}^2,t_1)$ and $f_{m(t_1)}(\mathbf{x}^1,t_1)>f_{m(t_1)}(\mathbf{x}^2,t_1)$. At time step $t_2$, we decrease the number of objectives by one, i.e., $m(t_2)=m(t_1)-1$. If $f_{m(t_1)}$ is removed at time step $t_2$, we can derive that $\mathbf{x}^1\preceq_{t_2}\mathbf{x}^2$. That is to say $\mathbf{x}^2$ is not in $PF_{t_2}$. On the other hand, if $f_i$, where $i\in\{1,\cdots,m(t_1)-1\}$, is removed at time step $t_2$, we can derive that $\mathbf{x}^1$ and $\mathbf{x}^2$ are still non-dominated from each other. In other words, $\mathbf{x}^1$ and $\mathbf{x}^2$ are still in $PF_{t_2}$. All in all, we conclude that $PF_{t_1}$ is a superset of $PF_{t_2}$.
\end{proof}

\section*{Acknowledgment}
This work was supported by EPSRC (grant no. EP/K001523/1) and NSFC (grant no. 61329302). Xin Yao was also supported by a Royal Society Wolfson Research Merit Award.

\bibliographystyle{IEEEtran}
\bibliography{IEEEabrv,dmop}

\end{document}